\documentclass[10pt,twocolumn,letterpaper]{article}

\usepackage{cvpr}              
\usepackage{pifont}
\usepackage{multirow}
\usepackage{makecell}
\usepackage{afterpage}
\usepackage{caption} 
\usepackage{cuted}
\usepackage{booktabs}   
\usepackage{tabularx}   
\usepackage{siunitx}    
\usepackage{caption}    

\definecolor{cvprblue}{rgb}{0.21,0.49,0.74}
\usepackage[pagebackref,breaklinks,colorlinks,allcolors=cvprblue]{hyperref}
\def\eqref#1{(\ref{eq:#1})}
\def\eqlabel#1{\label{eq:#1}}
\def\figref#1{\ref{fig:#1}}

\def\m#1{\ensuremath{\mathtt{#1}}}

\def\v#1{\ensuremath{\mathbf{#1}}}

\def\norm#1{\left\lVert#1\right\rVert}
\def\l2#1{\norm{#1}_2}

\def\m#1{\ensuremath{\mathtt{#1}}}      

\renewcommand{\v}[1]{\ensuremath{#1}}
\renewcommand{\m}[1]{\ensuremath{#1}}

\def\norm#1{\left\lVert#1\right\rVert}

\def\l2#1{\norm{#1}_2}

\def\vb{\v b}
\def\vx{\v x}

\def\vx{\v x}

\def\mM{\m M}

\def\vc{\v c}                          
\def\veps{\v\epsilon}                  
\def\vz{\v z}                          
\def\vzT{\vz_t}                        

\def\mM{\m M}                          

\def\cLDM{\mathcal{L}_{\text{LDM}}}     
\def\cP{\mathcal{P}}                   
\def\cD{\mathcal{D}}                   
\def\cE{\mathcal{E}}                   
\def\cDec{\mathcal{D}}                 

\def\concat{\operatorname{concat}}

\def\paperID{} 
\def\confName{CVPR}
\def\confYear{2026}

\title{MAGIC: Few-Shot Mask-Guided Anomaly Inpainting with Prompt Perturbation, Spatially Adaptive Guidance, and Context Awareness}

\date{} 

\author{%
\begin{minipage}{\textwidth}
\centering
\textbf{JaeHyuck Choi}\textsuperscript{1}\quad
\textbf{Minjun Kim}\textsuperscript{2}\quad
\textbf{Je Hyeong Hong}\textsuperscript{1,2*}
\\[0.6em]
\textsuperscript{1}Department of AI Semiconductor Engineering, Hanyang University, Seoul, Korea\\
\textsuperscript{2}Department of Electronic Engineering, Hanyang University, Seoul, Korea\\
\normalsize{\texttt{\{sunhp1333, ihatelemon, jhh37\}@hanyang.ac.kr}}
\end{minipage}
}

\begin{document}

\maketitle
\footnotetext[1]{Corresponding author}

\begin{strip}
\centering
\includegraphics[width=0.99\textwidth]{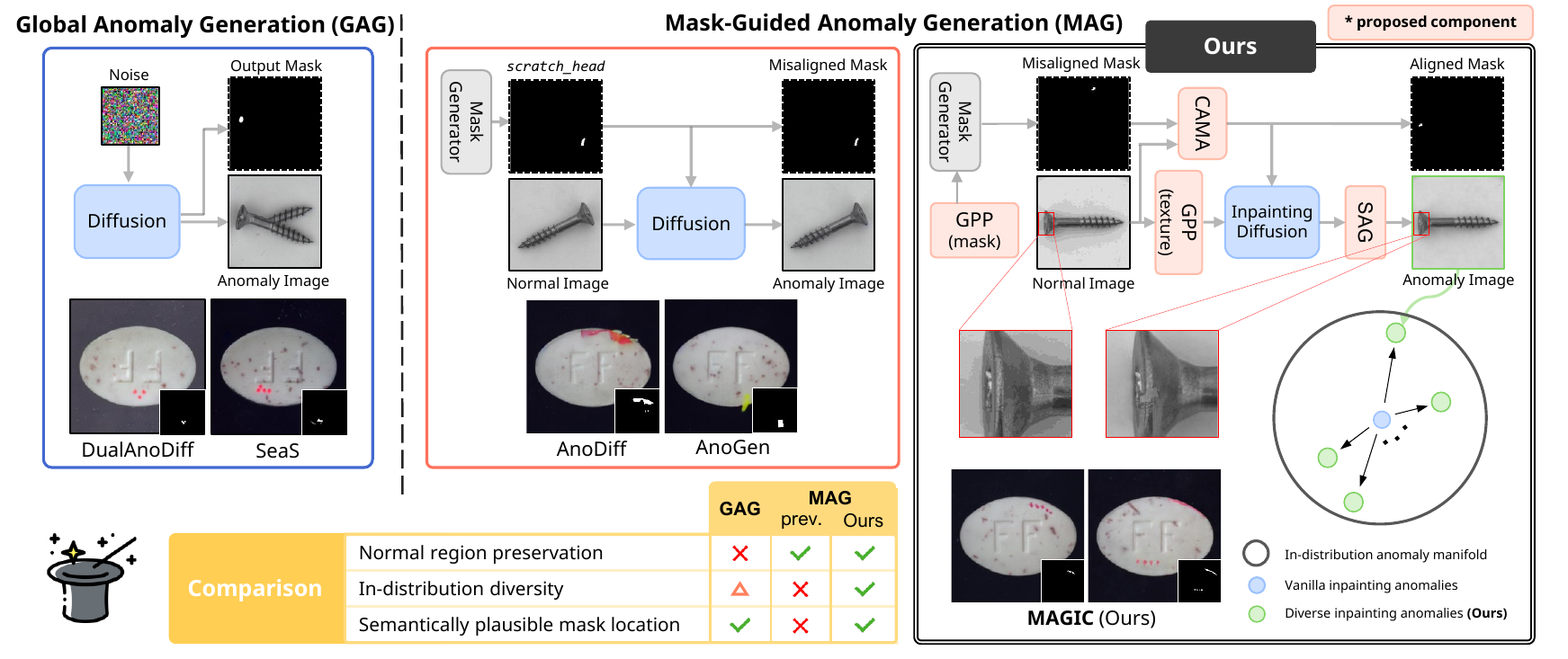}
\vspace{-3mm}
\captionof{figure}{
  Global Anomaly Generation (GAG) does not involve normal image guidance and often corrupts (normal) background textures (e.g., two screws).
  Mask-guided Anomaly Generation (MAG) keeps the background intact but (i) results in low diversity of generated images and (ii) produces low-fidelity anomalies when the mask is misplaced.
  \textbf{MAGIC} addresses both issues, promoting in-distribution diversity of anomaly mask shapes and textures and producing mask-accurate, realistic anomalies while preserving the normal background.
}


\label{fig:Motivation_figure}
\end{strip}

\begin{abstract}

Few-shot anomaly generation is a key challenge in industrial quality control. Although diffusion models are promising, existing methods struggle: global prompt-guided approaches corrupt normal regions, and existing inpainting-based methods often lack the in-distribution diversity essential for robust downstream models. We propose MAGIC, a fine-tuned inpainting framework that generates high-fidelity anomalies that strictly adhere to the mask while maximizing this diversity. MAGIC introduces three complementary components: (i) Gaussian prompt perturbation, which prevents model overfitting in the few-shot setting by learning and sampling from a smooth manifold of realistic anomalies, (ii) spatially adaptive guidance that applies distinct guidance strengths to the anomaly and background regions, and (iii) context-aware mask alignment to relocate masks for plausible placement within the host object. Under consistent identical evaluation protocol, MAGIC outperforms state-of-the-art methods on diverse anomaly datasets in downstream tasks. Code is available at \url{https://github.com/SpatialAILab/MAGIC}.

\end{abstract}

\vspace{-1mm}
\section{Introduction}

In the manufacturing industry, automatic detection, localization, and classification of anomalies are essential for quality control and improved yields~\cite{lin2024comprehensive, zhang2023dataaugmentation}.
However, practitioners face an inherent data imbalance: normal images are abundant while images depicting anomalies are scarce. Although prior studies~\cite{Roth2021TowardsTR} have shown that anomaly detection and localization can be achieved using only normal‐image training (e.g., one‐class classification or reconstruction‐based methods), accurate anomaly classification, which is crucial for identifying the source of defects, still requires labeled anomaly examples. To fill this gap, recent studies have turned to generative models that synthesize realistic anomalous images with the aim of providing training data for downstream models~\cite{Hu2023AnomalyDiffusionFA}.

Existing anomaly generation methods can be divided into two groups. The Global Anomaly Generation (GAG) methods~\cite{Jin2024DualInterrelatedDM, dai2025SeaS} generate anomaly images and masks simultaneously without requiring an input mask that specifies the anomaly region. 
While this is observed to generate diverse anomaly images, normal regions are often not preserved which can substantially degrade the realism of the generated image (see Fig.~\ref{fig:Motivation_figure}).
In contrast, Mask-guided Anomaly Generation (MAG) approaches utilize normal image and a user-provided mask to designate the anomaly region~\cite{Zavrtanik2021DRMA, Li2021CutPasteSL, wei2022mdgan, Hu2023AnomalyDiffusionFA, gui2024anogen, ali2024anomalycontrol}. While the background is now well-preserved, these models exhibit a significant new limitation: the limited in-distribution diversity of generated anomalies. Because these models are heavily conditioned on the normal image as a prior (i.e., for the non-masked region), the synthesized anomalies often lack variety, with similar textures appearing repeatedly.
Furthermore, they often suffer from provided input mask lying on semantically implausible locations within the host object, which results in unrealistic image generation. This combination of limited diversity and mask mislocation leads to poor performance on downstream tasks in the few-shot setting.


In this work, we present MAGIC to address above limitations of MAG approaches. Our solution comprises three components that work in concert to improve both the fidelity and diversity of generated anomalies. First, we introduce a method (GPP) that is applied during both training and inference to learn and sample from a smooth manifold of realistic anomalies thereby improving the in-distribution diversity. Second, we add a dedicated inference-time control (SAG) that spatially applies distinct guidance, allowing it to promote textural diversity within the anomaly region while simultaneously retaining high background consistency in the normal areas. Finally, a complementary module (CAMA) addresses the implausible mask location problem by aligning the input mask to a semantically plausible region.
These three complementary components built upon the Stable Diffusion inpainting backbone are:


\begin{itemize}[topsep=0pt,itemsep=0pt,parsep=0pt]

\item \textbf{Gaussian prompt perturbation (GPP).} GPP is applied during both training and inference, where Gaussian noise is injected into the anomaly-token embedding. During training, this process encourages the model to learn a smooth and continuous manifold of realistic, in-distribution anomalies in the few-shot setting. At test time, sampling from this manifold allows generation of a broader set of diverse anomaly textures and mask shapes.

\item \textbf{Spatially adaptive guidance (SAG).} 
This inference-time component applies spatially-varying classifier-free guidance (CFG) scales such that a lower CFG scale is applied to the masked anomaly region to promote textural diversity, while a higher CFG scale is applied to the normal region to retain high background consistency and fidelity.

\item \textbf{Context-aware mask alignment (CAMA).} Leveraging semantic correspondences, CAMA relocates an input mask to a more plausible object part, which helps prevent out-of-object artifacts and semantically invalid regions.

\end{itemize}
We validate MAGIC's effectiveness and generalization on downstream anomaly tasks with extensive experiments on MVTec-AD~\cite{bergmann2019mvtec}, VisA~\cite{zou2022visa}, MVTec 3D-AD~\cite{Bergmann2022mvtec3d}, and DAGM~\cite{Wieler2007dagm} using a standard evaluation protocol~\cite{Hu2023AnomalyDiffusionFA}.

\begin{figure*}[t]
  \centering
  \includegraphics[width=1.0\textwidth]{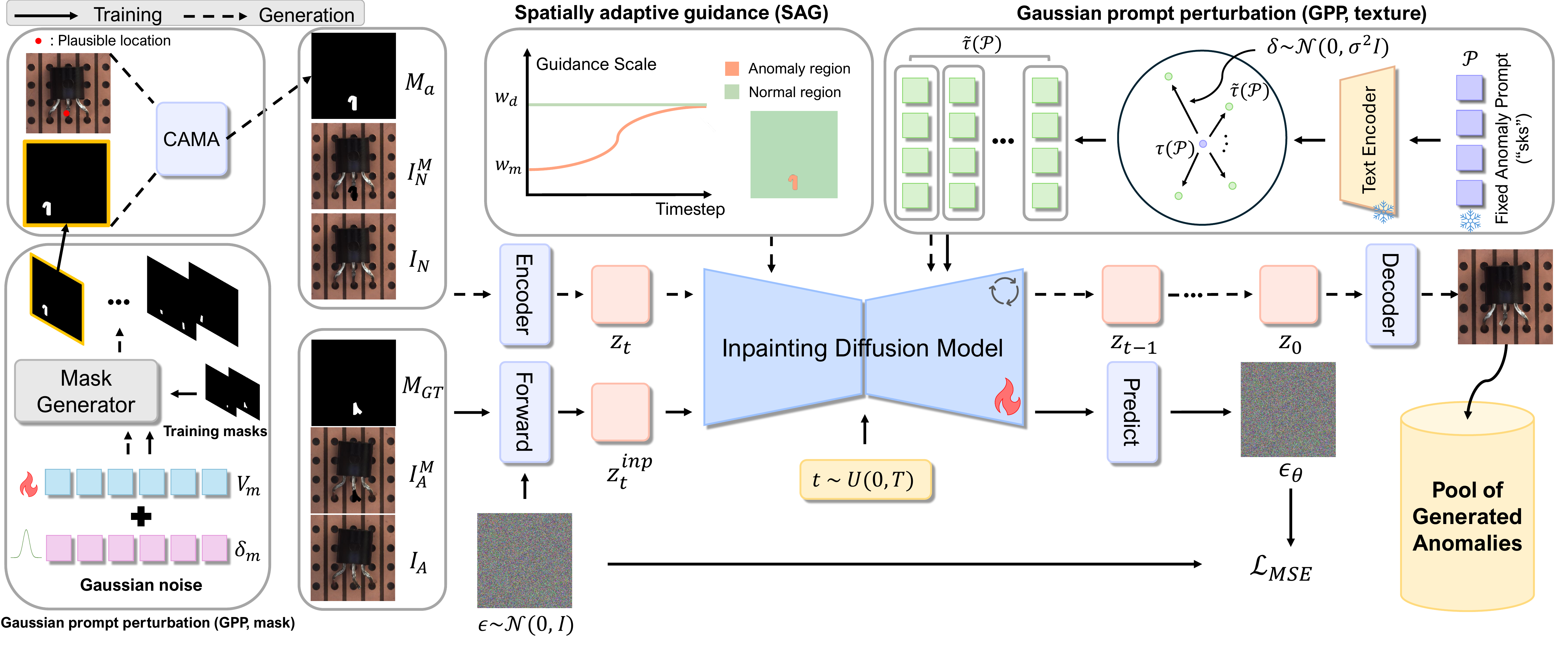}
  \vspace{-7mm}
  \caption{
    \textbf{Framework overview.}
    MAGIC presents three complementary modules to generate high-fidelity anomaly images that strictly adhere to the mask while maximizing diversity. Gaussian prompt perturbation is injected to encoded anomaly prompt $\mathcal{P}$ to learn a smooth manifold of realistic anomalies. We also apply spatially-varying CFG scales to the model to generate realistic and locally consistent anomalies. Finally, context-aware mask alignment relocates masks for plausible placement within the host object. This results in our method generating diverse but plausible anomaly images, significantly enhancing anomaly downstream tasks as will be shown in Sec.~\ref{sec:experimental_results}.
    }
\label{fig:Main_framework}
  \vspace{-5mm}
\end{figure*}

\section{Related work}
\subsection{Anomaly generation}
\paragraph{Global anomaly generation.}
Global Anomaly Generation (GAG) methods generate anomaly images and masks simultaneously without requiring an input mask that specifies the anomaly region. Early GAG approaches, often intended for data augmentation, relied on synthetic overlays. DRAEM~\cite{Zavrtanik2021DRMA} and PRNet~\cite{Zhang2022PrototypicalRN} overlay Perlin noise, with PRNet further borrowing anomaly textures from the DTD-Synthetic~\cite{Cimpoi_2014_CVPR} dataset.  
As none of these methods model the full image distribution, their anomalies can remain geometrically or photometrically inconsistent, limiting realism and diversity.
To overcome these limits, researchers adopted generative adversarial networks (GANs)~\cite{Niu2020sdgan,Zhang2021DefectGANHD,Duan2023FewShotDI}. Yet, these models still suffer from mode collapse and gradient instability~\cite{Mescheder2018WhichTM}, especially when only a few anomaly exemplars are available.
More recently, diffusion models have come to dominate high-fidelity image synthesis. DualAnoDiff~\cite{Jin2024DualInterrelatedDM} runs two attention-sharing streams and is capable of generating relatively realistic and diverse anomaly images. Similarly, SeaS~\cite{dai2025SeaS} proposes a unified U-Net model that uses specialized text prompts to simultaneously generate diverse anomalies, authentic normal products, and anomaly masks. Unfortunately, since they regenerate the entire image without mask conditioning, background textures can change undesirably, yielding unrealistic artifacts.

\vspace{-3mm}
\paragraph{Mask-guided anomaly generation.}
The mask-guided anomaly generation (MAG) approaches utilize normal image and a user-provided mask to designate the anomaly region. Same with GAG, early MAG methods were also frequently based on heuristics. Crop-Paste~\cite{Lin2020FewShotDS} transfers segmented anomalies to normal images, whereas CutPaste~\cite{Li2021CutPasteSL} creates anomalous region without utilizing real anomalies by copy–pasting patches from the same image. These hand-crafted synthesis methods often lack realism. To enhance the fidelity of the synthesized anomaly images, diffusion model has been adopted. AnoGen~\cite{gui2024anogen} and AnomalyDiffusion~\cite{Hu2023AnomalyDiffusionFA} leverage textual inversion~\cite{Gal2022AnII} to capture the detailed characteristics of anomalies. While These approaches prove spatial control, the frozen backbone restricts quality of anomaly textures and produce low-fidelity. DefectFill~\cite{Song2025DefectFillRD} fine-tunes an inpainting diffusion model and is capable of producing realistic anomaly textures, but it requires an object-specific textual prompt during training (e.g.\ “hazelnut”), which may not always be available for manufacturing components lacking descriptive labels. Furthermore, its diversity remains limited, a known issue for fine-tuned inpainting models~\cite{Wang2022DiverseII,Liu2021PDGANPD}.

\subsection{Personalized diffusion models}

State-of-the-art text-to-image diffusion backbones (Stable Diffusion, Imagen, GLIDE, etc.)~\cite{Rombach2021HighResolutionIS, Saharia2022PhotorealisticTD, Nichol2021GLIDETP, Zhang2023AddingCC} deliver high fidelity but struggle to reproduce a new subject from only a few reference images. Embedding-based methods leave the backbone frozen: Textual inversion~\cite{Gal2022AnII} learns a single pseudo-token, and DreamDistribution~\cite{zhao2025dreamdistribution} extends this to a distribution of soft-prompt embeddings, gaining diversity but often lacking  subject-specific details for images from narrow specific domains.

DreamBooth~\cite{ruiz2023dreambooth} instead fine-tunes the entire model such that a rare text prompt maps to the target concept, achieving better fidelity but more prone to overfitting (reduced diversity). This motivates us to adopt a DreamBooth-tuned inpainting model for high fidelity then  present modules to restore in-distribution diversity of generated images.

\section{Preliminaries}

\vspace{-1mm}
\paragraph{Text-to-image diffusion models.}

Diffusion models are generative models that transform samples drawn from a Gaussian noise distribution into realistic data through an iterative denoising process.
In text-to-image generation, a text encoder $\tau$ maps a prompt $\cP$ to a conditioning vector $\vc = \tau(\cP)$, which guides the image generation process.
LDMs~\cite{Rombach2021HighResolutionIS} extend this framework by operating in a lower-dimensional latent space for efficiency.
A pretrained encoder $\cE$ maps an image $\vx$ to a latent representation $\vz_0 = \cE(\vx)$, and a decoder $\cDec$ reconstructs the image via $\vx \approx \cDec(\vz_0)$.
Given a dataset $\cD = \{(\vx,\vc)\}$ of image–text pairs, the model learns the conditional distribution $p(\vx | \vc)$.
At each diffusion step $t$, Gaussian noise $\veps\sim\mathcal{N}(0,I)$ is added to the latent $\vz_0$ to obtain $\vzT$.
Then, training involves finding the diffusion model weights $\theta$ which minimizes
\begin{align}
  \cLDM
    &= \mathbb{E}_{\vzT,\veps,t}
      \Bigl[
        \bigl\lVert
          \veps - \veps_{\theta}(\vzT,\, t,\, \vc)
        \bigr\rVert_2^{2}
      \Bigr],
  \eqlabel{ldm_loss}
\end{align}
where $\veps_\theta$ is the noise predicted by the latent diffusion model at time step $t$ provided $z_t$ and $c$.


\vspace{-3mm}
\paragraph{Inpainting diffusion models.}

Inpainting with LDMs involves synthesizing plausible content in missing regions of a given image. 
For this purpose, the denoising process is now additionally conditioned on the  background-only image $B$ (with the anomaly region masked out) with its latent representation $\vb = \cE(B)$ and a binary mask $\mM$.
Training the model with this background conditioning can simply be achieved by forming a concatenated latent representation $\vzT^{\text{inp}} := \concat(\vzT,\, \vb,\, \mM)$ then minimize the same loss as in Eq.~\eqref{ldm_loss} with the exception of replacing $\vzT$ by $\vzT^{\text{inp}}$.
Our method employs the stable diffusion-based inpainting backbone~\cite{Rombach2021HighResolutionIS} with Dreambooth fine-tuning~\cite{ruiz2023dreambooth}.

\vspace{-3mm}
\paragraph{Classifier-free guidance.}
Classifier-free guidance (CFG) \cite{ho2022classifierfreediffusionguidance} is a widely used technique in recent text-to-image diffusion models to balance fidelity and diversity.
The denoising model produces two noise estimates: a conditional noise $\epsilon_{\theta}(x_t, t, c)$ that depends on input condition (e.g., text prompt) and an unconditional noise $\epsilon_{\theta}(x_t, t)$ that ignores it.
The final noise prediction $\tilde{\epsilon}$ is a linear combination of both:
\begin{equation}
\tilde{\epsilon}_{\theta}(x_t, t, c) = \epsilon_{\theta}(x_t, t) + w \cdot \big(\epsilon_{\theta}(x_t, t, c) - \epsilon_{\theta}(x_t, t)\big),
\end{equation}
where $w$ is the guidance scale controlling the influence of the conditional signal.
Adjusting $w$ allows a trade-off between image diversity and quality: larger values produce outputs more faithful to the condition but less diverse.



\begin{figure}[t]
  \centering
  \includegraphics[width=\linewidth]{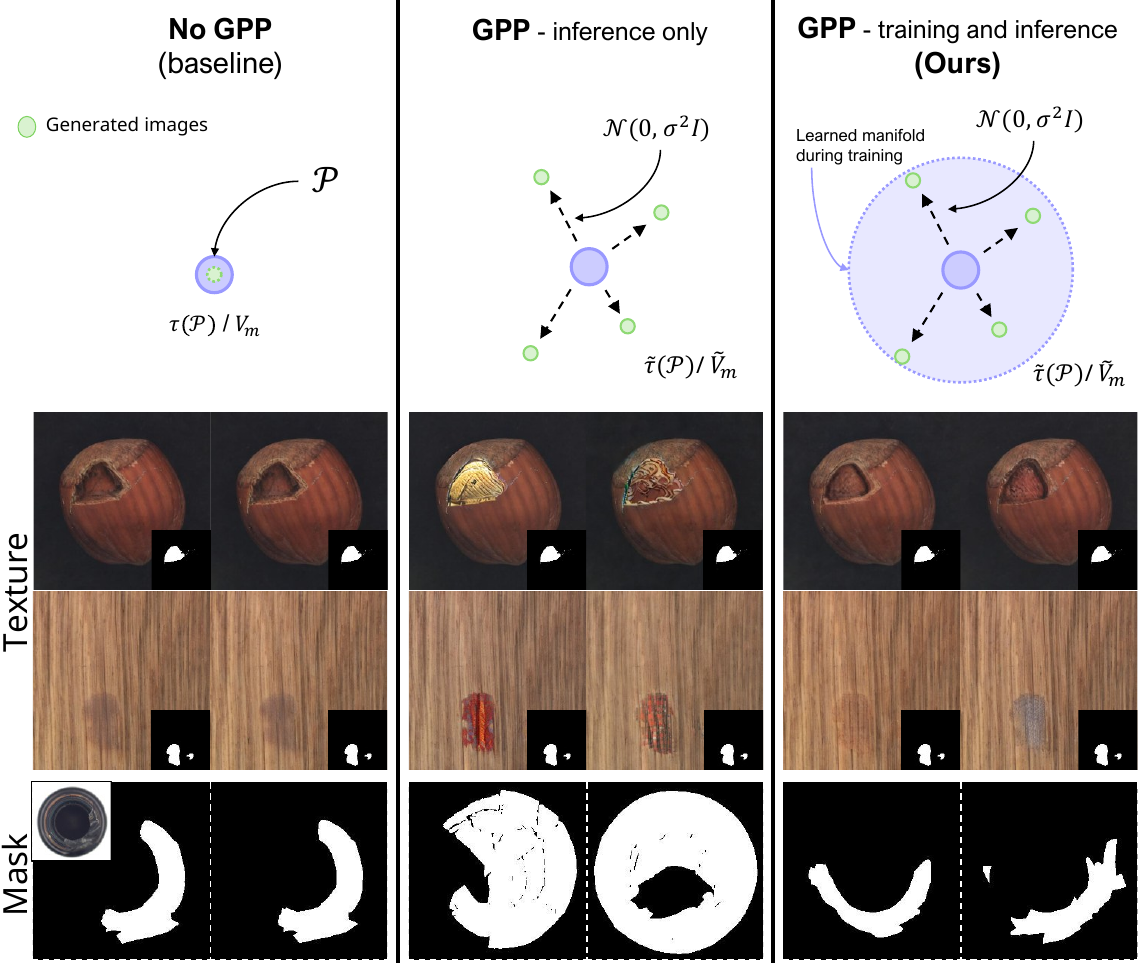}
  \vspace{-6mm}
  \caption{Effect of Gaussian Prompt Perturbation (GPP) shown on two different noise seeds. Applying GPP at both training and test times broadens the global appearance of both anomaly textures and mask shapes and avoids unrealistic out-of-distribution results.}
  \label{fig:GPP}
  \vspace{-4mm}
\end{figure}

\section{Proposed method}
\label{sec:method}
Our approach, MAGIC, introduces three complementary components to a fine-tuned inpainting diffusion model. We aim to enhance the in-distribution diversity of generated anomalies while ensuring they remain realistic and are placed in semantically plausible locations.

\subsection{Gaussian prompt perturbation}
\label{sec:GPP}

\begin{figure}[t]
\centering
\includegraphics[width=\linewidth]{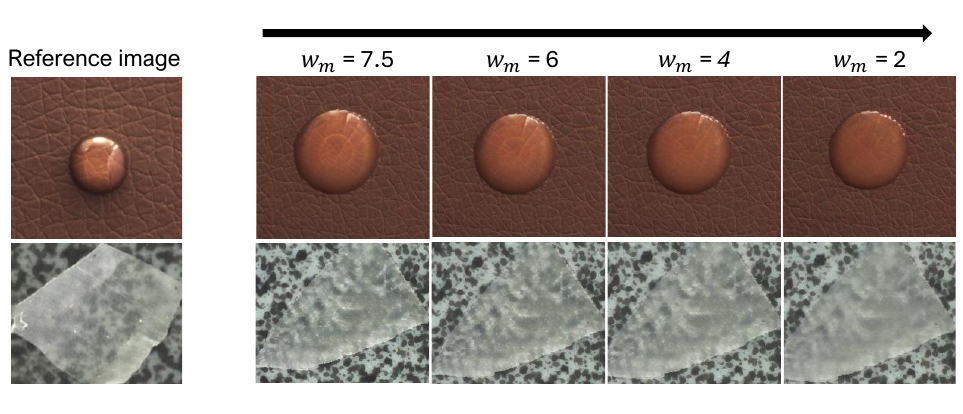}
\vspace{-7mm}
\caption{Effect of our Spatially Adaptive Guidance (SAG). Decreasing $w_m$ increases textural diversity within the anomaly region compared to the reference images.
Meanwhile, we observe the texture of the normal background is largely preserved.}
\vspace{-5mm}
\label{fig:SAG}
\end{figure}
A primary challenge in few-shot anomaly generation is the propensity for model overfitting. With only a few training examples, the fine-tuned model tends to memorize the training set, leading to a pronounced decrease in in-distribution diversity. To address this, we introduce Gaussian Prompt Perturbation (GPP), a strategy centered on a key finding: perturbation must be applied in both training and inference.


Some prior work~\cite{lingenberg2024diagen} has explored applying noise to embeddings at inference time to increase the diversity. However, in our experiments, we found this approach can be problematic. When the model encounters perturbed prompt embeddings at inference that it has never encountered during training, it can result in a distribution shift, leading to generation of unrealistic, out-of-distribution anomalies as shown in Fig.~\figref{GPP} and Table~\ref{tab:mask_evaluation_table}.

Instead, GPP integrates the perturbation directly into the training process. For a fixed anomaly prompt $\mathcal{P}$, we first map it to its embedding $\tau(\mathcal{P})$. Then, during both training and inference, we sample Gaussian noise $\delta$ and use the perturbed embedding:
\begin{align}
\tilde{\tau}(\mathcal{P}) = \tau(\mathcal{P}) + \delta, \quad \delta \sim \mathcal{N}(\v 0, \sigma^2 I)
\end{align}
where \(\sigma\) controls the perturbation scale.
By exposing the model to this ``ball'' of noisy embeddings during training, GPP functions as an effective regularizer. It deters the model from converging its representation to a few discrete points (the training examples) and instead guides it to learn a smoother, more continuous manifold. This mechanism is fundamental to generating novel samples that remain within the target distribution.

At test time, sampling from this same manifold allows generating a wider variety of anomaly textures and mask shapes that are both diverse and remain in-distribution (i.e. high-fidelity) (see Fig.~\figref{GPP}).
We apply this GPP strategy to both texture inpainting (via Dreambooth fine-tuning) and mask generation (via textual inversion) tasks.



\begin{figure*}[t]
  \centering
  \includegraphics[width=1.0\textwidth]{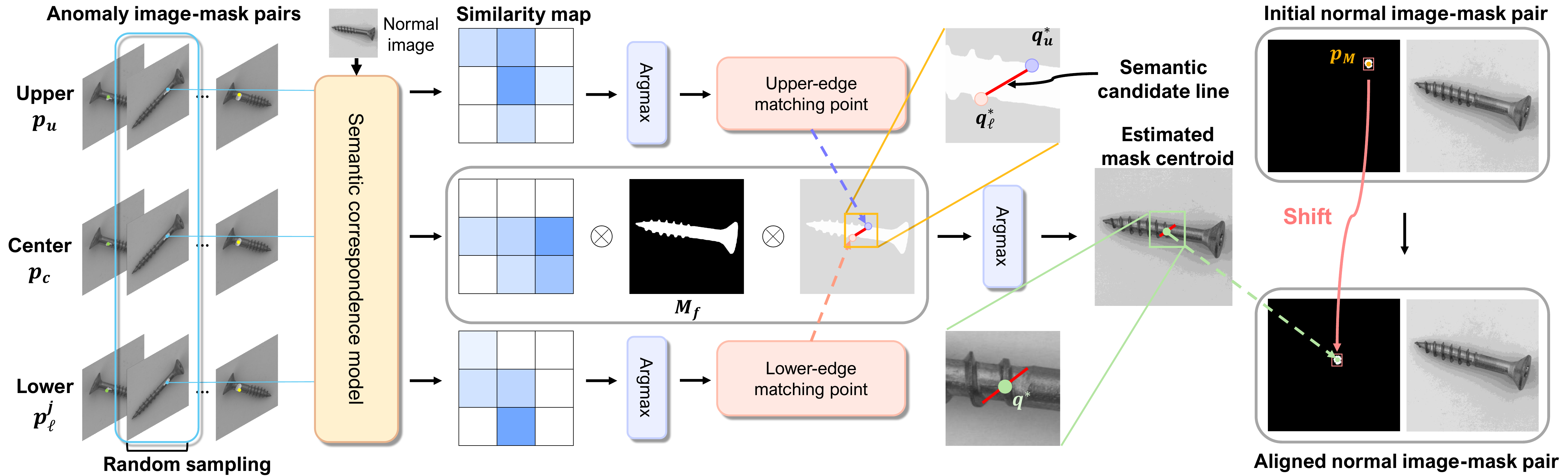}
  \vspace{-7mm}
  \caption{%
    An overview of our CAMA module.
    After sampling an anomaly image-mask pair from the anomaly training set, we automatically select three keypoints, upper \(p_u\), center \(p_c\) and lower \(p_\ell\), which are on the same vertical line crossing the mask centroid.
    These points are  matched to the normal image to create three similarity maps, $S_u$, $S_c$ and $S_\ell$. 
    The most likely locations of upper and lower points in the normal image ($q_u^*$ and $q_\ell^*$) yield a candidate line $\mathcal L$, providing a geometric cue for the mask center.
    This constraint, along with the foreground mask $M_f$ and similarity map $S_c$, is used to estimate optimal translation for shifting the mask to a semantically plausible region.
    }
    \label{fig:CAMA}
  \vspace{-5mm}
\end{figure*}

\vspace{-3mm}
\paragraph{Texture inpainting.} 
 To train the anomaly texture inpainting, we use the ground-truth anomaly image $I_A$ and its mask $M_{GT}$ to create the masked (normal background) image $I_A^M := (1 - M_{GT}) \odot I_A$, from which we obtain its latent encoding $b_A = \mathcal{E}(I_A^M)$. The input latent at timestep $t$ is $z_t^{inp} = \text{concat}(z_t, b_A, M_{GT})$. The network is then conditioned on our Gaussian-perturbed prompt $\tilde{\tau}(\mathcal{P})$, and its weights $\theta$ are optimized via:
\begin{align}
\theta^* = \underset{\theta}{\arg\min} \, \mathbb{E}_{z_t^{\text{inp}}, t, \epsilon, \delta} [ \| \epsilon - \epsilon_\theta(z_t^{\text{inp}}, t, \tilde{\tau}(\mathcal{P})) \|_2^2 ].
\end{align}
As shown in Fig.~\figref{GPP}, this approach diversifies the textural appearance without compromising the realism when applied at both training and test phases.

\vspace{-3mm}
\paragraph{Mask generation.}
\label{sec:mask_generator}
We found this same principle of diversity applies to the anomaly masks themselves. To avoid generating repetitive spatial patterns, we also apply GPP to the mask generation process. Following~\cite{Hu2023AnomalyDiffusionFA}, we learn a trainable mask embedding $V_m$ via textual inversion~\cite{Gal2022AnII}. During training, we add Gaussian noise $\delta_m \sim \mathcal{N}(0, \sigma_m^2 I)$ to get a perturbed version $\tilde{V}_m = V_m + \delta_m$. 
Formally, the mask embedding is optimized via:
\begin{align}
V^*_m = \underset{V_m}{\arg\min} \, 
\mathbb{E}_{z_t, t, \epsilon, \delta_m} 
\left[ \| \epsilon - \epsilon_\theta(z_t, t, \tilde{V}_m) \|_2^2 \right].
\end{align}
At inference, we use the optimized $V_m^*$ with Gaussian noise $\delta_m$ added to generate geometrically diverse yet semantically consistent masks as shown in Fig.~\figref{GPP} and Table~\ref{tab:mask_evaluation_table}.

\subsection{Spatially adaptive guidance}
\label{sec:SAG}

While GPP encourages diversity at the prompt embedding level, we hypothesize that diversity could be further promoted within the generative process itself. 
Standard classifier-free guidance (CFG) applies a uniform guidance scale $w$, which can force the model to adhere too strictly to the prompt and reduce textural variety.
To counteract this, we propose an inference-time mechanism called \emph{Spatially Adaptive Guidance (SAG)} in which we adopt spatially-varying CFG scales instead of fixed $w$. Specifically, we maintain a high fixed CFG scale $w_d =7.5$ for the normal region ($M_a=0$) to preserve background fidelity while we dynamically adjust the scale for the anomaly region ($M_a=1$).

During the early noisy diffusion steps (when the latent $z_t$ is dominated by strong noise), we apply a lower guidance scale $w_m$ to the anomaly region. This encourages the model to explore more diverse textural variations. As denoising progresses, we gradually increase this scale back to the default $w_d$ using a cosine schedule, allowing the model to refine the details and ensure the final anomaly texture is realistic. The adaptive scale $w_a(t)$ for the anomaly region at timestep $t$ is:
\begin{equation}
w_a(t) = w_m + (w_d - w_m) \cdot \tfrac{1}{2}\left(1 - \cos\left(\tfrac{\pi t}{T}\right)\right),
\end{equation}
where $T$ denotes the total number of diffusion steps. 





SAG allows the model to transition smoothly from low to high guidance within anomaly areas, promoting diversity during the early noisy phases and restoring realism in later denoising stages, while preserving the background fidelity.

\subsection{Context-aware mask alignment}
\label{sec:cama}

In object-centric categories (e.g., screw, transistor), where the object's position and orientation can vary from image to image, anomalies are often not random; they typically appear only in specific, semantically meaningful parts of the object (see Fig.~\ref{fig:CAMA}). Therefore, we need to ensure that any generated anomaly mask is placed in a semantically plausible region (e.g., screw head). To address this, we propose \emph{Context-Aware Mask Alignment (CAMA)}, a module that aligns a given anomaly mask to an appropriate spatial location on the target normal image by leveraging semantic correspondences. This enables consistent and realistic anomaly placement, even as object positions vary.


Our approach works by transferring the location of a real anomaly from our training set onto the target normal image. The process has two key inputs: the target normal image $I_N$ and a newly generated mask $M$ (which we aim to place). To find the correct placement for $M$, we first sample a reference anomaly pair ($I_A, M_{GT}$) from our few-shot training set. We use this reference pair only to find a semantically plausible location. Instead of relying on unstable single-point alignment~\cite{oquab2023dinov2}, CAMA selects three keypoints from the reference mask $M_{GT}$, namely the centroid and two boundary points, for a more robust mask transfer.
\begin{figure}[t]
    \centering
    \includegraphics[width=0.5\textwidth]{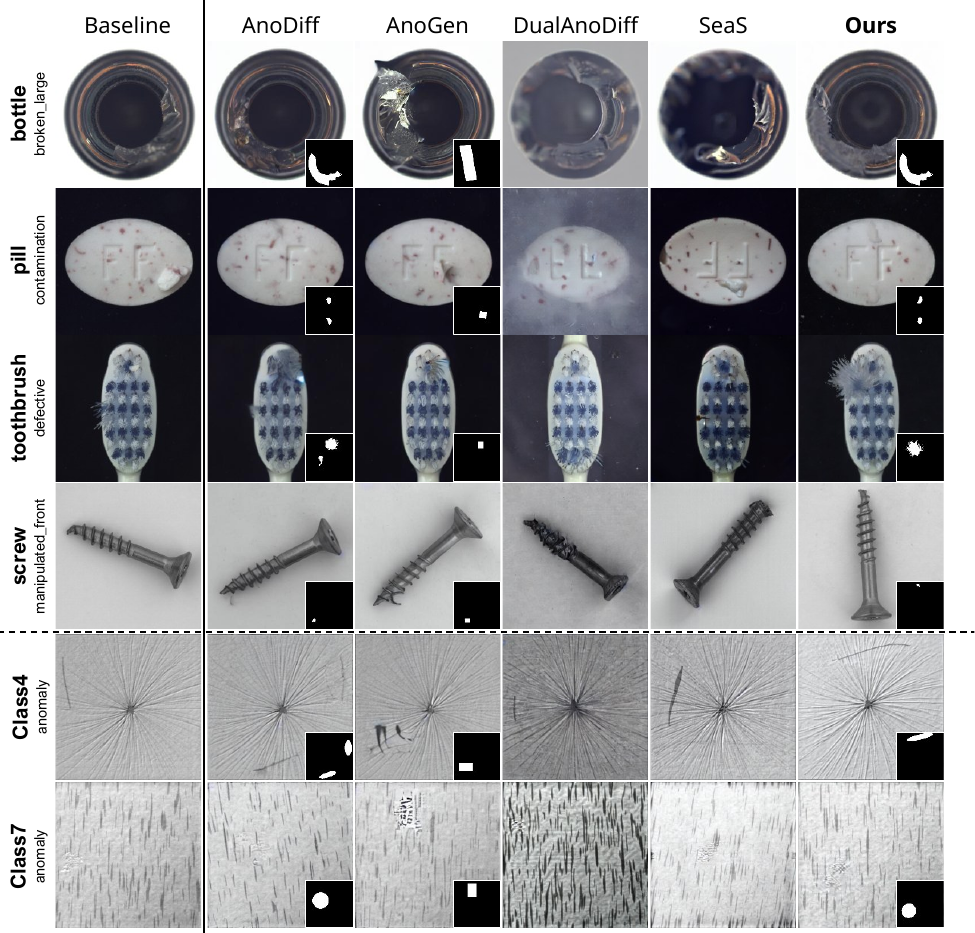}
    \vspace{-6mm}
    \caption{Qualitative comparison of generated anomaly images across different methods on the MVTec-AD and DAGM datasets. Anomaly masks are also provided for mask-guided approaches.}
    \label{fig:generated_image_2}
    \vspace{-5mm}
\end{figure}

The alignment process begins by extracting the three keypoints, namely center $p_c$, upper $p_u$, and lower $p_\ell$ keypoints, from our reference mask $M_{GT}$. These points are sampled from the edge of the mask region and share the same $x$-coordinate such that they lie on the same vertical line. Using a pretrained semantic correspondence model~\cite{zhang2024telling}, we then match these three keypoints from the reference image $I_A$ to the target normal image $I_N$. This yields three similarity maps $S_u, S_c$, and $S_\ell$. The most likely upper and lower correspondence points in the normal image are then found as $q_u^* = \arg\max_{(x,y)} {S}_u(x,y)$ and $q_\ell^* = \arg\max_{(x,y)} {S}_\ell(x,y)$ respectively, which together define a ``semantic candidate line'' $\mathcal L$ on the target image. 

With this line defined, we estimate the optimal new center point of the mask $q_c^*$, which should lie on the candidate line $\mathcal L$. We constrain our search to $\mathcal L$ and select the point that both lies within the object's foreground mask $M_f$ (extracted using U$^2$-Net~\cite{Qin2020U2NetGD}) and maximizes the score in the center similarity map $S_c$. This is formalized as:
\begin{align}
q_c^* = {\arg\max}_{(x,y),\in,\mathcal{L},,M_f(x,y)=1}, S_c(x,y).
\end{align}
Finally, we take the newly generated mask $M$ (from our mask generator) and translate it so its centroid aligns with this optimal point $q_c^\star$. This gives the aligned mask $M_a$. To ensure the anomaly remains perfectly within the object's bounds, we perform a logical AND operation with the foreground mask to yield the final mask $M_a \leftarrow M_a \cap M_f$.




\begin{figure}[t]
    \centering
    \includegraphics[width=0.5\textwidth]{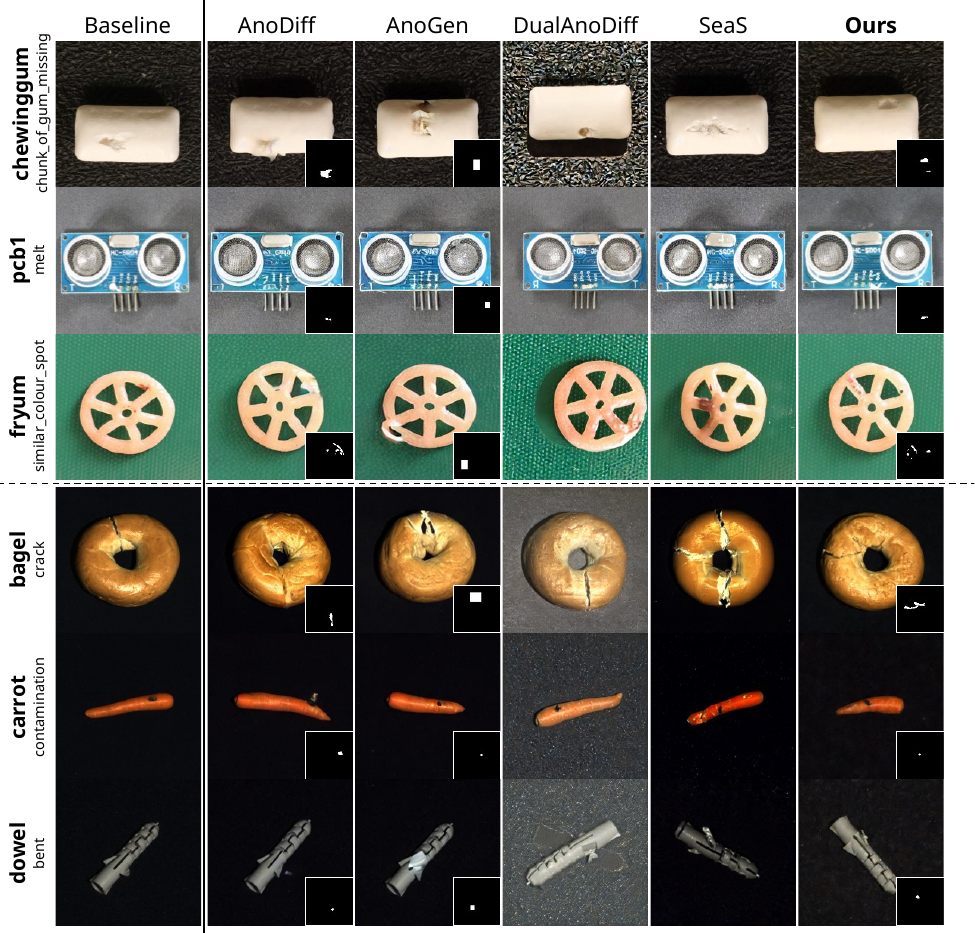}
    \vspace{-6mm}
    \caption{Qualitative comparison of generated anomaly images across different methods on the VisA and MVTec 3D-AD datasets. Anomaly masks are also provided for mask-guided approaches.}
    \label{fig:generated_image_1}
    \vspace{-4mm}
\end{figure}


\begin{table*}[t]
    \centering
    \footnotesize

    \setlength{\tabcolsep}{8pt} 
    \begin{tabular}{l|cc|cc|cc|cc}
        \toprule
        \multirow{2}{*}{\textbf{Method}}
        & \multicolumn{2}{c|}{\textbf{MVTec-AD}} 
        & \multicolumn{2}{c|}{\textbf{VisA}} 
        & \multicolumn{2}{c|}{\textbf{DAGM}} 
        & \multicolumn{2}{c}{\textbf{MVTec 3D-AD}} \\
        & KID $\downarrow$ & IC-L $\uparrow$
        & KID $\downarrow$ & IC-L $\uparrow$
        & KID $\downarrow$ & IC-L $\uparrow$
        & KID $\downarrow$ & IC-L $\uparrow$ \\
        \midrule
        AnoDiff~\cite{Hu2023AnomalyDiffusionFA}
        & 104.01 & 0.30
        & 110.76 & 0.30
        & \underline{64.76} & 0.44
        & \underline{70.54} & 0.26 \\
        AnoGen~\cite{gui2024anogen}
        & 105.39 & \underline{0.31}
        & 106.41 & 0.30
        & 100.70 & \underline{0.46}
        & 73.10 & 0.26 \\
        DualAnoDiff~\cite{Jin2024DualInterrelatedDM}
        & \underline{96.82} & \textbf{0.36}
        & 152.64 & \textbf{0.43}
        & 172.83 & 0.45
        & 143.20 & \textbf{0.37} \\
        SeaS~\cite{dai2025SeaS}
        & 126.59 & 0.35
        & \underline{97.00} & 0.23
        & 134.48 & \textbf{0.48}
        & 101.42 & 0.33 \\
        \midrule
        MAGIC (Ours)
        & \textbf{40.27} & 0.30
        & \textbf{79.81} & \underline{0.32}
        & \textbf{30.87} & 0.43
        & \textbf{51.28} & \underline{0.34} \\
        \bottomrule
    \end{tabular}
    \vspace{-2mm}
    \captionof{table}{
    Comparison of average KID and IC-LPIPS scores.
    \textbf{Bold} and \underline{underlined} indicate the best and second-best results, respectively.
    }
    \label{tab:kid_icl}

    \vspace{0.8em}

    \setlength{\tabcolsep}{5pt}
    \begin{tabular}{l|ccc|ccc|ccc|ccc}
        \toprule
        \multirow{2}{*}{\textbf{Method}}
        & \multicolumn{3}{c|}{\textbf{MVTec-AD}}
        & \multicolumn{3}{c|}{\textbf{VisA}}
        & \multicolumn{3}{c|}{\textbf{DAGM}}
        & \multicolumn{3}{c}{\textbf{MVTec 3D-AD}} \\
        & AUROC & AP & $F_1$-max
        & AUROC & AP & $F_1$-max
        & AUROC & AP & $F_1$-max
        & AUROC & AP & $F_1$-max \\
        \midrule
        AnoDiff~\cite{Hu2023AnomalyDiffusionFA}
        & 98.47 & \underline{99.40} & 97.25
        & 85.03 & 81.24 & 78.05
        & 96.95 & 94.68 & 93.48
        & 86.07 & 94.20 & 90.96 \\
        AnoGen~\cite{gui2024anogen}
        & 98.27 & 99.20 & 96.89
        & \underline{89.64} & 85.64 & 81.28
        & 96.84 & 95.02 & 92.07
        & 88.10 & 95.08 & 90.90 \\
        DualAnoDiff~\cite{Jin2024DualInterrelatedDM}
        & 97.09 & 98.64 & 95.25
        & 86.45 & 82.05 & 75.47
        & \underline{98.83} & \underline{98.32} & \underline{97.25}
        & 80.76 & 91.57 & 88.88 \\
        SeaS~\cite{dai2025SeaS}
        & \underline{98.73} & 99.29 & \underline{97.32}
        & 88.60 & \underline{87.44} & \underline{81.51}
        & 95.31 & 92.56 & 93.67
        & \underline{88.32} & \underline{95.48} & \underline{91.05} \\
        \midrule
        MAGIC (Ours)
        & \textbf{99.36} & \textbf{99.73} & \textbf{98.49}
        & \textbf{94.28} & \textbf{92.80} & \textbf{88.12}
        & \textbf{99.86} & \textbf{99.68} & \textbf{99.09}
        & \textbf{93.37} & \textbf{97.52} & \textbf{92.21} \\
        \bottomrule
    \end{tabular}
    \vspace{-2mm}
    \captionof{table}{Comparison of image-level anomaly detection and localization results across four datasets.}
    \label{tab:detection_performance_4datasets_final_v3}

    \vspace{0.8em}

    \setlength{\tabcolsep}{2pt}
    \begin{tabular}{l|cccc|cccc|cccc|cccc}
        \toprule
        \multirow{2}{*}{\textbf{Method}}
        & \multicolumn{4}{c|}{\textbf{MVTec-AD}}
        & \multicolumn{4}{c|}{\textbf{VisA}}
        & \multicolumn{4}{c|}{\textbf{DAGM}}
        & \multicolumn{4}{c}{\textbf{MVTec 3D-AD}} \\
        & AUROC & AP & $F_1$-max & PRO
        & AUROC & AP & $F_1$-max & PRO
        & AUROC & AP & $F_1$-max & PRO
        & AUROC & AP & $F_1$-max & PRO \\
        \midrule
        AnoDiff~\cite{Hu2023AnomalyDiffusionFA}
        & \underline{98.39} & 74.02 & 70.14 & 92.57
        & 96.01 & 44.48 & 44.90 & 84.13
        & \underline{97.97} & 72.33 & 67.67 & 93.83
        & 97.31 & 15.16 & 19.85 & 91.32 \\
        AnoGen~\cite{gui2024anogen}
        & 96.25 & 64.20 & 61.60 & 90.65
        & 97.25 & 45.41 & 46.85 & \underline{87.44}
        & 97.24 & 69.96 & 66.44 & 92.98
        & 98.30 & 19.96 & 26.46 & 91.33 \\
        DualAnoDiff~\cite{Jin2024DualInterrelatedDM}
        & 97.41 & 76.79 & 72.90 & 91.32
        & 96.98 & 49.46 & 51.03 & 86.73
        & 97.77 & \underline{80.29} & \underline{75.31} & \underline{94.97}
        & 97.34 & 16.79 & 24.64 & 89.68 \\
        SeaS~\cite{dai2025SeaS}
        & 98.34 & \underline{78.33} & \underline{73.71} & \underline{94.42}
        & \underline{97.88} & \textbf{69.60} & \textbf{67.56} & 85.90
        & 94.27 & 74.88 & 69.41 & 89.28
        & \underline{99.04} & \underline{25.86} & \underline{30.67} & \underline{92.84} \\
        \midrule
        MAGIC (Ours)
        & \textbf{99.15} & \textbf{81.99} & \textbf{77.33} & \textbf{96.22}
        & \textbf{98.38} & \underline{59.08} & \underline{60.30} & \textbf{92.06}
        & \textbf{99.15} & \textbf{81.94} & \textbf{75.84} & \textbf{97.38}
        & \textbf{99.18} & \textbf{26.01} & \textbf{30.72} & \textbf{96.38} \\
        \bottomrule
    \end{tabular}
    \vspace{-2mm}
    \captionof{table}{Comparison of pixel-level anomaly detection and localization results across four datasets.}
    \label{tab:detection_performance_4datasets_final}
    \vspace{-4mm}
\end{table*}

\section{Experimental results}
\label{sec:experimental_results}

\subsection{Experimental settings}
\label{sec:experimental_settings}

\paragraph{Datasets.}
We conduct experiments on four datasets: MVTec-AD~\cite{bergmann2019mvtec}, VisA~\cite{zou2022visa}, DAGM~\cite{Wieler2007dagm}, and MVTec 3D-AD~\cite{Bergmann2022mvtec3d} (RGB images). 
MVTec-AD includes 15 categories with up to 8 anomaly types; VisA covers 12 objects in 3 domains; MVTec 3D-AD contains 10 categories with up to 4 anomaly types; DAGM provides 10 texture-like categories with defect labels.

\vspace{-3mm}
\paragraph{Implementation details.}
We adopt the protocol of AnomalyDiffusion~\cite{Hu2023AnomalyDiffusionFA}, using one-third (rounded down) of the anomaly images for training and the remainder for testing. We avoid the DualAnoDiff split~\cite{Jin2024DualInterrelatedDM} because it may introduce train--test overlap (see supplementary material). For DAGM, since anomaly images are abundant, we use five anomaly images per category to embrace a few-shot training setting.
For model implementation, our framework is built on the Stable Diffusion 2 inpainting model with a DDIM scheduler and 50 inference steps. For GPP, we set $\sigma = 1.0$, $\sigma_m = 0.1$, and use a threshold $w_m = 4$. Additional details are provided in the supplementary material.

\vspace{-3mm}
\paragraph{Evaluation metrics.}
To assess the quality and diversity of the generated images, we synthesized 500 anomaly image–mask pairs per anomaly type in each category. We measured two key aspects: in-distribution quality using KID~\cite{Binkowski2018DemystifyingMG}, and raw diversity using IC-LPIPS~\cite{Ojha2021FewshotIG}. While a high IC-LPIPS score is desirable, this metric can be artificially inflated by unrealistic, out-of-distribution samples. We therefore consider KID to be the more critical measure of useful, in-distribution generative quality. Since our training setting involves limited data, we adopted KID instead of Fréchet Inception Distance (FID)~\cite{Heusel2017GANsTB} as it provides a more reliable estimate in low-data regimes~\cite{karras2020training}. To validate downstream effectiveness, we trained a ResNet-34~\cite{he16resnet} for anomaly classification. For image-level anomaly detection, we measure AUROC, AP, and F1-max score. For pixel-level localization, we report these same metrics and additionally evaluate PRO (per-region-overlap).

\begin{table}[t]
    \centering
    \small
    \setlength{\tabcolsep}{3pt}
    \begin{tabular}{l|ccc|c}
        \toprule
        \textbf{Method} & \textbf{MVTec-AD} & \textbf{VisA} & \textbf{MVTec3D} & \textbf{Average}\\
        \midrule
        AnoDiff~\cite{Hu2023AnomalyDiffusionFA}
        & 64.90 & 42.86 & \underline{46.07} & 51.28 \\
        AnoGen~\cite{gui2024anogen}
        & 56.92 & 46.75 & 42.68 & 48.78 \\
        DualAnoDiff~\cite{Jin2024DualInterrelatedDM}
        & \underline{68.50} & \underline{52.14} & 37.41 & \underline{52.82}\\
        SeaS~\cite{dai2025SeaS}
        & 52.73 & 36.55 & 39.01 & 42.76\\
        \midrule
        \textbf{MAGIC (ours)} & \textbf{78.06} & \textbf{68.51} & \textbf{53.33} & \textbf{66.63} \\
        \bottomrule
    \end{tabular}
    \vspace{-2mm}
    \caption{Comparison of average anomaly classification accuracies (\%) across three datasets. ResNet-34 is used for classification.}
    \vspace{-4mm}
    \label{tab:classification_accuracy}
\end{table}

\vspace{-4mm}
\paragraph{Compared methods.}
We compare our method with AnomalyDiffusion~\cite{Hu2023AnomalyDiffusionFA}, AnoGen~\cite{gui2024anogen}, DualAnoDiff~\cite{Jin2024DualInterrelatedDM}, and SeaS~\cite{dai2025SeaS}, reproducing all baselines with their official implementations. Each method uses fixed hyperparameters across datasets except for SeaS, which uses different hyperparameters per dataset as outlined in their paper. DefectFill is not included in the main experiment due to its reliance on test-set masks and its limited handling of objects with varying geometric transformations (e.g., rotation or translation) but partly compared in the supplementary material. For all methods including ours, generated anomaly images are evaluated without any filtering or post-processing to reflect their raw generation performance. For anomaly type classification, we train a ResNet-34 using the synthetic samples from each methods. 
Further implementation details are provided in the supplementary material.

\subsection{Comparison of anomaly generation results}
Figs.~\ref{fig:generated_image_2} and~\ref{fig:generated_image_1} illustrate representative anomaly images generated by each method, while Table~\ref{tab:kid_icl} provides quantitative evaluations of fidelity and diversity. Our method achieves the lowest KID scores among all baselines, suggesting a closer alignment with the real anomaly distribution. Although consistently low (i.e., favorable) KID scores are observed across categories and datasets, MAGIC yields smaller (worse) IC-LPIPS values compared to other approaches. As noted in Sec.~\ref{sec:experimental_settings}, we anticipate that this may stem from inherent limitations of the metric: IC-LPIPS can be inflated when background preservation is insufficient (as observed in GAG), and it may also be sensitive to out-of-distribution behavior in generation (as seen in MAG).

\subsection{Comparison on anomaly downstream tasks}

\paragraph{Anomaly classification.}
We evaluate the downstream utility of our model by measuring anomaly classification accuracy. A ResNet-34~\cite{he16resnet} is trained on the generated samples and tested on~\cite{bergmann2019mvtec, Bergmann2022mvtec3d, zou2022visa}. DAGM~\cite{Wieler2007dagm} is excluded as the dataset does not comprise anomaly-type labels. As shown in Table~\ref{tab:classification_accuracy}, our approach consistently achieves higher accuracy than existing baselines, with an average gain of 13.81 pp over the previous state of the art~\cite{Jin2024DualInterrelatedDM}.

\vspace{-3mm}
\paragraph{Anomaly detection and localization.}
We evaluate the effectiveness of our generated data for anomaly detection and localization on~\cite{bergmann2019mvtec, Bergmann2022mvtec3d, zou2022visa, Wieler2007dagm}, comparing results with AnomalyDiffusion, AnoGen, DualAnoDiff, and SeaS. Each method generates 500 anomaly image–mask pairs, used to train a U-Net~\cite{ronneberger15unet}.
We report image-level evaluation in Table~\ref{tab:detection_performance_4datasets_final_v3} and the pixel-level evaluation in Table~\ref{tab:detection_performance_4datasets_final}. 
Our approach achieves new state-of-the-art performance on almost all metrics across four  datasets, demonstrating significant and consistent improvements over existing methods.

\begin{table}
    \centering
    \footnotesize
    \setlength{\tabcolsep}{2.5pt}
    \begin{tabular}{cccc|ccc}
    \toprule
    \makecell[c]{\textbf{GPP}\\\textbf{(mask)}} &
    \makecell[c]{\textbf{GPP}\\\textbf{(texture)}} & \textbf{SAG} & \textbf{CAMA} & \textbf{KID} & \textbf{IC-LPIPS} &     \makecell[c]{\textbf{Cls acc.}\\\textbf{(\%)}} \\
    \midrule
    
       \ding{55}     &     \ding{55}     &     \ding{55}      &    \ding{55}      &
    45.98 & 0.299 & 71.53 \\
    
    \ding{51}      &    \ding{55}      &     \ding{55}      &   \ding{55}       &
    43.19 & 0.301 & 73.14 \\
    
    \ding{51} &    \ding{51}      &      \ding{55}    &      \ding{55}    & 
    42.59 & \underline{0.303} & 74.36 \\

    \ding{51} &    \ding{51}      & \ding{51} &     \ding{55}      & 
    \underline{41.66} & \textbf{0.306} & \underline{76.50} \\
    \midrule
    \ding{51} &     \ding{51}     & \ding{51} & \ding{51} & 
    \textbf{40.27} & 0.302 & \textbf{78.06}\\
    
    \bottomrule
    \end{tabular}
    \vspace{-2mm}
    \caption{Ablation study of our proposed components (GPP, SAG, and CAMA) on the MVTec-AD~\cite{bergmann2019mvtec} dataset.
    }
    \label{tab:ablation_main}
    \vspace{-4mm}
\end{table}

\subsection{Ablation study}

We conduct an ablation study to evaluate the individual contributions of GPP, SAG, and CAMA (Table~\ref{tab:ablation_main}).
Applying GPP to both mask and texture generation reduces KID by 3.39, slightly improves IC-LPIPS, and increases classification accuracy by 2.83 pp. Adding SAG further lowers KID and yields an additional IC-LPIPS improvement, leading to a 2.14 pp accuracy gain. Together, these results indicate that GPP and SAG enhance in-distribution diversity while preserving realism.
Finally, applying CAMA further reduces KID and improves accuracy by 1.56 pp. This demonstrates that the module effectively enforces structural consistency between synthesized anomalies and the surrounding context, benefiting classification performance.

\begin{table}[t]
    \centering
    \small
    \setlength{\tabcolsep}{3pt}
    \footnotesize

    \begin{tabular}{l|cc|ccc}
        \toprule
        & \multicolumn{2}{c|}{\textbf{Mask}} 
        & \multicolumn{3}{c}{\textbf{Texture}} \\
        \textbf{Method} & KID $\downarrow$ & IC-L $\uparrow$ & KID $\downarrow$ & IC-L $\uparrow$ & Cls acc.(\%)\\
        \midrule
        No GPP
        & 71.95 & 0.094 & \underline{43.19} & 0.301 & \underline{73.14} \\ 
        GPP (inference only)~\cite{lingenberg2024diagen}
        & \underline{57.33} & \textbf{0.101} & 58.95 & \textbf{0.341} & 64.87\\ 
        \midrule
        GPP (training + inference)
        & \textbf{46.30} & \underline{0.098} & \textbf{42.59} & \underline{0.303} & \textbf{74.36} \\
        \bottomrule
    \end{tabular}
    \vspace{-2mm}
    \caption{Detailed ablation study of GPP. We compare the baseline (No GPP), inference-only GPP, and our full training + inference GPP approach on mask/texture generation quality and downstream classification accuracy on the MVTec-AD dataset.}
    \label{tab:mask_evaluation_table}
    \vspace{-4mm}
\end{table}

We also conducted an ablation study specifically on the GPP strategy (Table~\ref{tab:mask_evaluation_table}). To evaluate its impact, we measured quality metrics for both generated masks and anomaly images (textures): generated masks were compared against the training masks in MVTec-AD, and generated textures were compared against the training anomaly images.
We compared on three configurations: (1) the baseline without perturbation, (2) applying noise only at inference time~\cite{lingenberg2024diagen}, and (3) our full approach (training + inference).
Applying GPP only at inference introduces a significant train-test mismatch, shifting the model to an unseen latent space. This mismatch results in unrealistic artifacts (KID +15.76) and a noticeable drop in downstream classification performance (–8.27 pp). When GPP is applied during both training and inference, this mismatch is mitigated.
We conjecture that, by learning to operate on the ``smooth manifold'' during training, the model can  sample diverse yet realistic anomalies at the inference phase, achieving the best in-distribution quality (KID) and highest anomaly classification accuracy.

\section{Conclusion}

We presented MAGIC, a framework for few-shot mask-guided anomaly inpainting that addresses key limitations of existing methods, namely limited in-distribution diversity and semantically implausible mask placement. Our solution's first component, Gaussian prompt perturbation (GPP), is applied during both training and inference, allowing the model to learn and sample from a smooth manifold of realistic, in-distribution anomalies. This serves as the foundation for generating diverse examples and is complemented by spatially adaptive guidance (SAG), an inference-time mechanism that applies distinct CFG scales: a lower scale is applied to the anomaly region to promote textural diversity, while a higher scale is applied to the normal region to retain background consistency. Finally, context-aware mask alignment (CAMA) ensures semantic plausibility by relocating input masks to more appropriate object regions. 
Through extensive experiments and ablations on diverse anomaly datasets~\cite{bergmann2019mvtec, zou2022visa, Bergmann2022mvtec3d, Wieler2007dagm}, we showed that our components work in concert to generate a high-fidelity and diverse set of anomalies. This improved generative quality leads to consistent and significant gains in downstream anomaly detection, localization, and classification tasks, establishing a new state-of-the-art in the few-shot setting.


\vspace{-4mm}
\paragraph{Limitations.}

CAMA's accuracy depends on a plausible input mask; significant deviations can reduce alignment precision. Additionally, MAGIC's performance relies on its pre-trained components (U$^2$-Net, GeoAware-SC), so their inherent limitations in visually ambiguous or unseen domains may lead to less precise anomaly placement.


\section{Acknowledgement}


This work was supported by Hyundai Motor Company, by the Korea Evaluation Institute of Industrial Technology (KEIT, RS-2026-25525507, 50\%) and the Korea Institute for Advancement of Technology~(KIAT, Corporate Demand-driven Challenge and Innovative R\&D Program for Next Generation Researchers, RS-2026-25539646, 25\%) both funded by the Ministry of Trade
Industry \& Energy (MOTIE, Korea), and by the Institute of Information \& Communications Technology Planning \& Evaluation~(IITP) under the Artificial Intelligence Semiconductor Support Program to Nurture the Best Talents~(IITP-(2025)-RS-2023-00253914, 25\%) grant funded by the Korea government~(MSIT).

{
    \small
    \bibliographystyle{ieeenat_fullname}
    \bibliography{cvpr_2026/main}
}

\maketitlesupplementary

\section{Additional details on datasets and experimental settings}
\label{sec:dataset_usage}

\subsection{Details on experimental datasets.}
\label{subsec:dataset_usage}

\paragraph{MVTec-AD, VisA, and MVTec 3D-AD datasets.}
For these datasets, we followed a unified experimental setup based on the protocol proposed in AnomalyDiffusion~\cite{Hu2023AnomalyDiffusionFA}. 
All three datasets were trained under the same configuration to ensure consistency across benchmarks. 
For the VisA dataset, we used only the single-anomaly subset.

\vspace{-3mm}
\paragraph{DAGM dataset.}
To follow the few-shot setting, we used five anomaly images per class as training images and treated the remaining images as test samples. Since DAGM primarily consists of texture-type defects, all categories were trained under identical settings without CAMA.

\subsection{Implementation details}
\label{subsec:implementation_details}

Our framework is based on the Stable Diffusion 2 inpainting model, using a DDIM scheduler with 50 denoising steps during inference. 
Training is performed for 5000 steps using AdamW (learning rate $5\times10^{-6}$) with a batch size of 4 on a single NVIDIA RTX A6000 GPU, requiring approximately 1.5 hours per anomaly class. 
Inference takes about 1 second per image without CAMA and roughly 5 seconds with CAMA.
For semantic correspondence extraction in CAMA, we use the pretrained GeoAware-SC model~\cite{zhang2024telling}.

\section{Additional results}
\label{sec:additional_results}

\begin{table*}[t]
\centering
\small
\begin{subtable}[t]{0.48\linewidth}
\centering
\begin{tabular}{ccccc}
\toprule
\(\sigma\) & KID $\downarrow$ & Cls.\ Acc.\ $\uparrow$ & AUC-P $\uparrow$ & AP-P $\uparrow$ \\
\midrule
0.3 & 41.65& 76.16 & 98.5 & 80.0 \\
0.5 & 41.08& 76.32 & \underline{99.1} & 80.7 \\
0.7 & 41.08& 77.54 & \underline{99.1} & 82.1 \\
0.9 & \underline{40.89} & \underline{77.73} & \textbf{99.2} & \underline{83.0} \\
1.0 & \textbf{40.27} & \textbf{78.06} & \textbf{99.2} & 82.0 \\
1.2 & 41.32& 76.40 & \textbf{99.2} & \textbf{83.6} \\
\bottomrule
\end{tabular}
\end{subtable}
\hfill
\begin{subtable}[t]{0.48\linewidth}
\centering
\begin{tabular}{ccccc}
\toprule
\(\sigma_m\) & KID $\downarrow$ & IC-LPIPS $\uparrow$ \\
\midrule
0.5 & 55.13 & \textbf{0.111} \\
0.1 & \textbf{46.30} & \underline{0.098} \\
0.05 & \underline{49.82} &  0.097 \\

\bottomrule
\end{tabular}
\end{subtable}
\caption{
Classification and pixel-level detection performance under varying hyperparameters.
The left subtable presents the effect of the texture-related perturbation scale \(\sigma\) used for texture anomaly generation, 
while the right subtable shows the effect of the mask-related perturbation scale \(\sigma_m\) used for mask generation.
}
\label{tab:classification_tau2}
\vspace{-2mm}
\end{table*}

\begin{table*}[t]
\centering
\small
\begin{minipage}{0.48\textwidth}
\centering
\begin{tabular}{ccccc}
\toprule
\(w_m\) & KID $\downarrow$ & Cls.\ Acc.\ $\uparrow$ & AUC-P $\uparrow$ & AP-P $\uparrow$ \\
\midrule
2.5 & 40.28 & \underline{77.34} & \underline{99.2} & 83.1 \\
3.0 & 42.04 & 77.15 & 99.1 & 82.8 \\
3.5 & \textbf{39.39} & 76.32 & 99.1 & 82.6 \\
4.0 & 40.72 & \textbf{78.06} & 99.1 & 82.0 \\
4.5 & \underline{40.32} & 76.89 & \underline{99.2} & \underline{83.2} \\
5.0 & 40.39 & 76.62 & \textbf{99.3} & \textbf{83.9} \\
\bottomrule
\end{tabular}
\caption{
Effect of the minimum guidance scale \(w_m\) used in Spatially Adaptive Guidance (SAG).
}
\label{tab:classification_wm}
\end{minipage}
\hfill
\begin{minipage}{0.48\textwidth}
\centering
\begin{tabular}{l|cc}
\toprule
\multirow{2}{*}{\textbf{Dataset}} & \multicolumn{2}{c}{Alignment Score (\%)} \\
&  w/o CAMA  & w/ CAMA \\
\midrule
MVTec-AD & 69.71 & \textbf{91.06} (+ 21.35 pp) \\
MVTec 3D-AD & 79.71 & \textbf{86.86} (+ 7.15 pp) \\
VisA & 50.11 & \textbf{73.06} (+ 22.95 pp) \\
\bottomrule
\end{tabular}
\caption{
Evaluation on CAMA. It measures how much of the input mask lies on the object's foreground mask.
}
\label{tab:alignment_cama}
\end{minipage}
\end{table*}

\begin{figure*}[!t]
  \centering
  \includegraphics[width=1.0\textwidth]{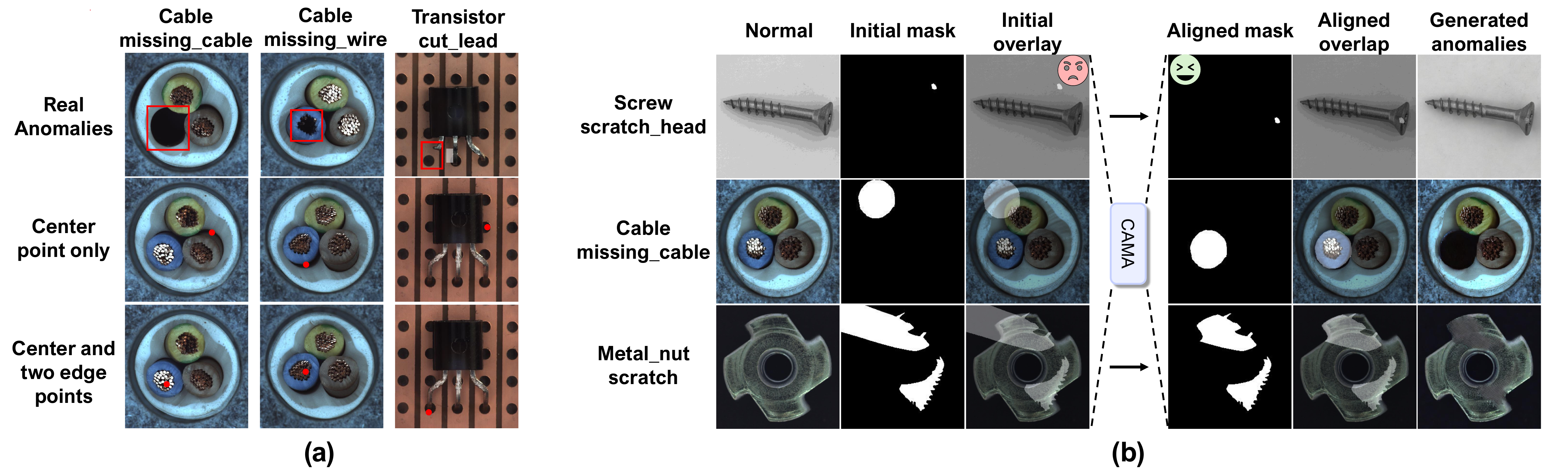}
    \caption{
    Qualitative results of keypoint-based anomaly localization. 
    (a) Comparison between using the center point only ($p_c$) and using the center point with edge points ($p_u$, $p_\ell$) in CAMA. Red dots denote the matched positions $q_c^*$ on the normal image. When edge points are included, $q_c^*$ is more accurately localized to semantically meaningful regions.
    (b) With CAMA, even misaligned masks are refined to align with the normal context, enabling precise anomaly generation.
    }
  \label{fig:CAMA_supple}
  \vspace{-5mm}
\end{figure*}

\begin{table*}[t]
  \centering
  \small
  \setlength{\tabcolsep}{5pt}
  \renewcommand{\arraystretch}{1.05}
  \resizebox{\textwidth}{!}{%
    \begin{tabular}{c|cccc|cccc|cccc|cccc}
      \toprule
      \multirow{2}{*}{Category}
      & \multicolumn{4}{c|}{DualAnoDiff (official code)}
      & \multicolumn{4}{c|}{DualAnoDiff$^{*}$ (pretrained weights)}
      & \multicolumn{4}{c|}{DualAnoDiff$^\dagger$(paper values)}
      & \multicolumn{4}{c}{MAGIC (ours)}\\
      & AUC-P & AP-P & F1-P & AP-I
      & AUC-P & AP-P & F1-P & AP-I
      & AUC-P & AP-P & F1-P & AP-I
      & AUC-P & AP-P & F1-P & AP-I\\
      \midrule
bottle & \textbf{99.7} & \textbf{98.7} & \underline{88.4} & 81.4 & 99.3 & 92.7 & 84.1 & \underline{99.9} & \underline{99.5} & 93.4 & 85.7 & \textbf{100.0} & \textbf{99.7} & \underline{95.4} & \textbf{88.5} & \textbf{100} \\

cable & \textbf{98.6} & \textbf{91.3} & 72.9 & 69.9 & 91.2 & 54.4 & 52.6 & 89.4 & \underline{97.5} & \underline{82.6} & \textbf{76.9} & \underline{98.3} & 96.5 & 81.5 & \underline{75.4} & \textbf{99.3} \\

capsule & \underline{98.8} & 47.8 & 52.8 & 99.0 & 97.8 & 45.7 & 46.0 & 96.2 & \textbf{99.5} & \textbf{73.2} & \textbf{67.0} & \underline{99.2} & 98.3 & \underline{61.7} & \underline{60.8} & \textbf{99.5} \\

carpet & 99.1 & 87.1 & 78.5 & 99.6 & \textbf{99.6} & \underline{87.8} & \underline{79.3} & \underline{99.7} & \underline{99.4} & \textbf{89.1} & \textbf{80.2} & \textbf{99.9} & 99.1 & 86.5 & 78.4 & 98.7 \\

grid & 95.4 & 54.9 & \underline{55.0} & 99.3 & \underline{98.6} & 53.3 & 52.1 & \textbf{100.0} & 98.5 & \underline{57.2} & 54.9 & 99.7 & \textbf{99.5} & \textbf{59.3} & \textbf{59.0} & \underline{99.9} \\

hazelnut & 99.5 & 88.7 & 81.0 & \underline{99.9} & \underline{99.7} & \underline{96.3} & \underline{91.1} & \underline{99.9} & \textbf{99.8} & \textbf{97.7} & \textbf{92.8} & \textbf{100.0} & \textbf{99.8} & 95.9 & 90.2 & \textbf{100.0} \\

leather & \textbf{100.0} & \textbf{99.8} & \textbf{82.8} & \underline{73.6} & \underline{99.9} & 86.5 & 77.3 & \textbf{100.0} & \underline{99.9} & \underline{88.8} & \underline{78.8} & \underline{73.6} & 99.6 & 82.2 & 74.4 & \textbf{100.0} \\

metal\_nut & 99.5 & 96.7 & 91.8 & 99.8 & 99.5 & 97.5 & 91.0 & 99.7 & \underline{99.6} & \underline{98.0} & \underline{93.0} & \underline{99.9} & \textbf{99.8} & \textbf{99.0} & \textbf{95.3} & \textbf{100.0} \\

pill & \underline{97.1} & 83.0 & 77.9 & 97.4 & 96.3 & 80.0 & 75.3 & 98.3 & \textbf{99.6} & \textbf{95.8} & \textbf{89.2} & \underline{99.0} & \textbf{99.6} & \underline{90.2} & \underline{82.4} & \textbf{99.6} \\

screw & \underline{98.2} & 50.4 & 51.2 & \underline{97.6} & 95.2 & 9.4 & 14.0 & 95.0 & 98.1 & \textbf{57.1} & \textbf{56.1} & 95.0 & \textbf{99.2} & \underline{51.4} & \underline{52.1} & 99.4 \\

tile & 99.6 & 95.4 & 89.2 & \textbf{100.0} & 99.6 & 95.3 & 88.8 & \textbf{100.0} & \underline{99.7} & \underline{97.1} & \underline{91.0} & \textbf{100.0} & \textbf{99.8} & \textbf{97.9} & \textbf{92.5} & \textbf{100.0} \\

toothbrush & 93.8 & 56.2 & 60.9 & 96.3 & \underline{98.3} & 53.4 & 59.1 & 99.0 & 98.2 & \underline{68.3} & \underline{68.6} & \underline{99.7} & \textbf{99.2} & \textbf{71.2} & \textbf{74.5} & \textbf{100.0} \\

transistor & 92.5 & 73.9 & 71.1 & \underline{98.1} & \underline{98.0} & 77.4 & 71.0 & 93.6 & \underline{98.0} & \textbf{86.7} & \textbf{79.6} & 93.7 & \textbf{98.9} & \underline{84.9} & \underline{78.3} & \textbf{100.0} \\

wood & 98.3 & 84.6 & 77.2 & \underline{99.9} & \underline{99.0} & \underline{87.8} & \underline{80.0} & \textbf{100.0} & \textbf{99.4} & \textbf{91.6} & \textbf{83.8} & \underline{99.9} & 98.8 & 85.1 & 77.4 & 99.6 \\

zipper & \underline{99.5} & \underline{89.0} & \underline{82.0} & \textbf{100.0} & \underline{99.5} & 86.0 & 78.6 & \textbf{100.0} & \textbf{99.6} & \textbf{90.7} & \textbf{82.7} & \textbf{100.0} & \underline{99.5} & 87.6 & 80.8 & \textbf{100.0} \\

\midrule
Average 
& 97.4 & 76.8 & 72.9 & 98.6
& 98.1 & 73.6 & 69.4 & 98.0 
& \underline{99.1} & \textbf{84.5} & \textbf{78.7} & \underline{99.0} 
& \textbf{99.2} & \underline{82.0} & \underline{77.3} & \textbf{99.7} \\
      \bottomrule
    \end{tabular}%
  }
  \vspace{1mm}
  \caption{Detection \& localization results (AUC-P, AP-P, F1-P, AP-I). 
    DualAnoDiff denotes our reproduced results using the official implementation; 
    DualAnoDiff$^{*}$ indicates results obtained using the author's GitHub pretrained weights; 
    DualAnoDiff$^{\dagger}$ shows the values reported in the original paper. \textbf{Note that we have not been able to reproduce the paper values to this date.} For MAGIC (ours), AUC-P, AP-P, F1-P, and AP-I correspond to AUROC-P, AP-P, f1\_max-P, and PRO-P from Table~2. 
    Bold = best, underline = second-best.}
  \label{tab:dualanodiff_detection_performance}
\end{table*}

\begin{table*}[t]
  \centering
  \small 
  \setlength{\tabcolsep}{4pt}
  \renewcommand{\arraystretch}{1.05}
  \begin{tabular}{lcccc}
    \toprule
    \textbf{Category} & DualAnoDiff (official code) & DualAnoDiff$^{*}$ (pretrained weights) & DualAnoDiff$^\dagger$(paper values) & MAGIC (ours) \\
    \midrule
bottle      & 72.09 & 58.14 & \textbf{79.07} & \underline{76.74} \\
cable       & 56.25 & 56.25 & \textbf{78.12} & \underline{68.75} \\
capsule     & 48.00 & 53.33 & \textbf{70.67} & \underline{58.67} \\
carpet      & \underline{70.97} & 67.74 & \textbf{79.03} & 62.90 \\
grid        & 60.00 & \underline{70.00} & \textbf{80.00} & 60.00 \\
hazelnut    & 85.42 & 83.33 & \underline{89.58} & \textbf{97.92} \\
leather     & 84.13 & \textbf{95.24} & \underline{90.48} & 85.71 \\
metal\_nut  & 76.56 & 70.31 & \underline{89.06} & \textbf{90.62} \\
pill        & 33.33 & 50.00 & \underline{56.25} & \textbf{67.71} \\
screw       & 58.02 & 48.15 & \underline{70.37} & \textbf{82.72} \\
tile        & \underline{98.25} & \textbf{100.00} & \textbf{100.00} & \textbf{100.00} \\
transistor  & \textbf{71.43} & \underline{67.86} & \textbf{71.43} & \textbf{89.29} \\
wood        & 71.43 & \textbf{92.86} & \underline{85.71} & 73.81 \\
zipper      & \underline{73.17} & 26.83 & \textbf{75.61} & \textbf{78.05} \\
    \midrule
\textbf{Average} & 68.50 & 67.15 & \textbf{79.67} & \underline{78.06} \\
    \bottomrule
  \end{tabular}
  \vspace{1mm}
  \caption{Quantitative comparison of anomaly classification accuracy (\%) across different generation methods using ResNet-34~\cite{he16resnet}. DualAnoDiff denotes our reproduced results using the official implementation; 
    DualAnoDiff$^{*}$ indicates results obtained using the author's GitHub pretrained weights; 
    DualAnoDiff$^{\dagger}$ shows the values reported in the original paper.
    \textbf{Note that we have not been able to reproduce the paper values to this date.} Bold = best, underline = second-best.}
  \label{tab:classification_dualanodiff}
  \vspace{-3mm}
\end{table*}


\subsection{Ablation on hyperparameters}

We investigate the impact of three key hyperparameters: the texture perturbation scale~\(\sigma\) used in Gaussian Prompt Perturbation, the mask perturbation scale~\(\sigma_m\), and the minimum CFG scale~\(w_m\) used in our Spatially Adaptive Guidance (SAG) mechanism. A summary of these hyperparameters is provided in Table~\ref{tab:classification_tau2}.
The texture perturbation scale~\(\sigma\), applied during both training and inference, controls the degree of global variation in the generated anomalies. We observe that \(\sigma = 1.0\) strikes the most effective balance between diversity and fidelity, resulting in the best downstream anomaly classification performance. Smaller values reduce anomaly diversity, whereas larger values introduce excessive variation that can slightly degrade reconstruction fidelity and classification accuracy. Similarly, for the mask perturbation scale~\(\sigma_m\), we find that \(\sigma_m = 0.1\) achieves the optimal downstream anomaly classification performance, offering a stable balance between perturbation strength and feature consistency.
As shown in Table~\ref{tab:classification_wm}, We also examine the effect of the minimum CFG scale \(w_m\) that governs the initial guidance strength within anomaly regions in SAG. Lower values encourage diversity in early diffusion steps, while higher values restrict variability. Among the tested configurations, we observe that \(w_m = 4.0\) provides the best overall performance.
Although the model remains relatively robust to moderate changes in these hyperparameters, we report these ablation results in the main paper~\cite{choi26cvprf} to support reproducibility and to highlight empirically effective settings.

\subsection{CAMA ablation study}

\paragraph{IoU evaluation.}
To directly evaluate how much of the generated anomaly mask $M_A$ falls onto the object's foreground mask $M_O$, we compute an alignment score defined as $|M_A \cap M_O| / |M_A|$. A higher score indicates better alignment. We evaluated this metric on the object-centric classes from MVTec-AD, MVTec 3D-AD, and VisA. 

As illustrated in Table~\ref{tab:alignment_cama}, CAMA provides a consistent improvement in mask alignment across all three tested datasets, suggesting it as an effective module.
\vspace{-3mm}
\paragraph{Qualitative results.}
As illustrated in Fig.~\ref{fig:CAMA_supple}, incorporating multiple spatial keypoints can improve the accuracy of anomaly placement in Context-Aware Mask Alignment (CAMA). In particular, relying only on the central keypoint \( p_c \) may lead to suboptimal alignment, especially for anomaly classes that involve missing or removed structural parts, such as \textit{missing\_cable} in the \textit{cable} category or \textit{cut\_lead} in \textit{transistor}. In such cases, the central region often lacks distinctive features, making correspondence based solely on \( p_c \) less reliable. To address this, we supplement the alignment process with additional keypoints—namely, the upper and lower keypoints \( p_u \) and \( p_\ell \)—to provide better spatial guidance. Fig.~\ref{fig:CAMA_supple} qualitatively shows that using all three keypoints together consistently helps to guide the anomaly to more semantically appropriate regions.

\begin{table*}[t!]
  \centering
  \small
  \renewcommand{\arraystretch}{1.05}

  \begin{minipage}[t]{0.52\textwidth}
    \centering
    \scriptsize
    \setlength{\tabcolsep}{12pt}
    \begin{tabular}{ccc|cc}
      \toprule
      \multirow{2}{*}{Category} &
        \multicolumn{2}{c|}{DefectFill (paper values)} &
        \multicolumn{2}{c}{MAGIC (ours)} \\
      & KID $\downarrow$ & IC-LPIPS $\uparrow$
      & KID $\downarrow$ & IC-LPIPS $\uparrow$ \\
      \midrule
      bottle      & \textbf{30.99} & 0.12 & 47.16 & \textbf{0.13} \\
      capsule     & \textbf{5.60}  & 0.18 & 22.61 & \textbf{0.20} \\
      carpet      & 50.37          & \textbf{0.22} & \textbf{33.85} & \textbf{0.22} \\
      hazelnut    & \textbf{1.13}  & 0.31 &  5.75 & \textbf{0.32} \\
      leather     & \textbf{74.66} & 0.30 & 113.50 & \textbf{0.31} \\
      pill        & \textbf{8.76}  & 0.23 & 59.52 & \textbf{0.26} \\
      tile        & \textbf{45.14} & \textbf{0.44} & 51.47 & 0.43 \\
      toothbrush  & \textbf{3.19}  & 0.15 & 30.62 & \textbf{0.25} \\
      wood        & \textbf{4.72}  & \textbf{0.35} & 28.53 & 0.34 \\
      zipper      & \textbf{34.91} & 0.20 & 58.61 & \textbf{0.20} \\
      \midrule
      \textbf{Average} & \textbf{25.95} & 0.25 & 45.16 & \textbf{0.27} \\
      \bottomrule
    \end{tabular}
  \end{minipage}
  \hfill
  \begin{minipage}[t]{0.46\textwidth}
    \centering
    \scriptsize
    \setlength{\tabcolsep}{20pt}
    \begin{tabular}{c c c}
      \toprule
      \textbf{Category} &
      \shortstack[c]{DefectFill\\(paper values)} &
      MAGIC (ours) \\
      \midrule
      bottle     & 97.56 & \textbf{97.67} \\
      capsule    & 87.50 & \textbf{98.67} \\
      carpet     & 87.72 & \textbf{90.32} \\
      hazelnut   & \textbf{100} & \textbf{100} \\
      leather    & 93.22 & \textbf{100} \\
      pill       & \textbf{97.53} & 95.83 \\
      tile       & \textbf{100} & \textbf{100} \\
      wood       & \textbf{100} & 92.86 \\
      zipper     & 90.91 & \textbf{98.78} \\
      \midrule
      \textbf{Average} & 94.94 & \textbf{97.13} \\
      \bottomrule
    \end{tabular}
  \end{minipage}
  \vspace{1mm}
    \caption{
    Quantitative results of DefectFill and MAGIC. The left table reports KID and IC-LPIPS scores, and the right table shows anomaly classification accuracy using ResNet-34~\cite{he16resnet}. 
    All results are obtained following the same evaluation protocol described in the DefectFill paper~\cite{Song2025DefectFillRD}, where test-set anomaly masks are used during generation.
    Since this setting relies on test annotations and does not explicitly account for variation in object poses (e.g., rotation or translation), we respectfully present these results in the supplementary material rather than the main paper~\cite{choi26cvprf}, in order to maintain alignment with our assumption of training-only supervision.
    }
  \label{tab:defectfill_comparison}
  \vspace{-5mm}
\end{table*}

\begin{table}[t]
  \centering
  \small
  \setlength{\tabcolsep}{2pt}
  \renewcommand{\arraystretch}{1.05}
  \begin{tabular}{l|cccc|cccc}
    \toprule
    \multirow{2}{*}{Category} &
      \multicolumn{4}{c|}{DefectFill (paper values)} &
      \multicolumn{4}{c}{MAGIC (ours)} \\
    & AUC-P & AP-P & F1-P & PRO
    & AUC-P & AP-P & F1-P & PRO \\
    \midrule
    bottle     & \textbf{1.00} & 0.96 & 0.90 & 0.97 & \textbf{1.00} & \textbf{0.97} & \textbf{0.92} & \textbf{0.98} \\
    capsule    & \textbf{1.00} & \textbf{0.75} & 0.69 & 0.96 & 0.99 & 0.71 & \textbf{0.70} & \textbf{0.96} \\
    carpet     & \textbf{0.99} & \textbf{0.92} & \textbf{0.86} & \textbf{0.96} & \textbf{0.99} & 0.89 & 0.82 & 0.94 \\
    hazelnut   & \textbf{1.00} & \textbf{0.99} & 0.94 & \textbf{0.99} & \textbf{1.00} & \textbf{0.99} & \textbf{0.95} & 0.98 \\
    leather    & \textbf{1.00} & \textbf{0.91} & \textbf{0.83} & \textbf{0.98} & \textbf{1.00} & 0.89 & 0.81 & \textbf{0.98} \\
    pill       & \textbf{1.00} & \textbf{0.98} & \textbf{0.93} & \textbf{0.98} & \textbf{1.00} & 0.93 & 0.84 & 0.97 \\
    tile       & \textbf{1.00} & 0.97 & 0.90 & 0.98 & \textbf{1.00} & \textbf{0.99} & \textbf{0.94} & \textbf{0.99} \\
    toothbrush & 0.99 & 0.89 & 0.82 & 0.94 & \textbf{1.00} & \textbf{0.98} & \textbf{0.93} & \textbf{0.98} \\
    wood       & 0.99 & 0.89 & 0.82 & 0.94 & \textbf{1.00} & \textbf{0.92} & \textbf{0.88} & \textbf{0.98} \\
    zipper     & \textbf{1.00} & 0.93 & 0.86 & 0.98 & \textbf{1.00} & \textbf{0.95} & \textbf{0.88} & \textbf{0.98} \\
    \midrule
    \textbf{Average} &
      \textbf{1.00} & \textbf{0.92} & \textbf{0.86} 
  \centering& \textbf{0.97} &
      \textbf{1.00} & \textbf{0.92} & \textbf{0.86} & \textbf{0.97} \\
    \bottomrule
  \end{tabular}
  \vspace{-2mm}
  \caption{Comparison of pixel-level anomaly detection and localization performance between DefectFill (from paper) and MAGIC.}
  \label{tab:detection_defectfill_magic}
  \vspace{-7mm}
\end{table}

\section{Implementation details and reproduction settings for baselines}
\label{sec:baseline_comparison}

\paragraph{DualAnoDiff.} We reproduced the DualAnoDiff baseline using the official implementation provided by the authors. Nonetheless, we found that our reproduced results did not fully match the quantitative values reported in the original paper. For the MVTecAD dataset, we applied the Background Compensation Module (BCM) to the same categories reported in their paper: \textit{bottle}, \textit{toothbrush}, and \textit{pill}. For the VisA and MVTec 3D-AD datasets, BCM was applied to all categories following our unified evaluation protocol. In contrast, for the DAGM dataset—which primarily consists of texture-type anomalies—we did not apply BCM, as foreground–background separation is not meaningful in this setting. Finally, for all datasets (MVTecAD, VisA, and MVTec 3D-AD), foreground masks were extracted using U\textsuperscript{2}-Net with the default threshold of 0.5, as specified in the original codebase.

Table~\ref{tab:dualanodiff_detection_performance} in this document summarizes the quantitative results in the following order: our reproduced model based on the official implementation, the results obtained from the official pre-trained checkpoint released by the authors, and the original performance reported in the paper. Additionally, we report classification performance and our proposed method's results under the same evaluation setup.

Furthermore, according to the authors of the original DualAnoDiff paper~\cite{Jin2024DualInterrelatedDM}, the number of defective samples was divided by three and, in some cases, selectively rounded up when a remainder was present. However, it is not clearly specified in the paper which categories this rounding was applied to, nor which specific image indices or labels were used for training. To avoid reproduction variation due to such ambiguity, we adopted a deterministic protocol that is identical to the setting used in \emph{AnomalyDiffusion}~\cite{Hu2023AnomalyDiffusionFA} by sorting test anomaly images within each anomaly class and using the first third as the training set without rounding up. Accordingly, all experiments in our main paper~\cite{choi26cvprf} were conducted using this consistent setup.

In Table~\ref{tab:dualanodiff_detection_performance} and Table~\ref{tab:classification_dualanodiff}, both our method and the reproduced DualAnoDiff baseline (denoted as DualAnoDiff) generate anomaly images following the same data selection protocol as AnomalyDiffusion~\cite{Hu2023AnomalyDiffusionFA} to ensure consistency in comparison. This reproduction was performed using the official codebase released by the authors. DualAnoDiff$^{*}$ refers to the results obtained using the official checkpoint provided by the original authors, while DualAnoDiff$^\dagger$ denotes the original performance reported in their paper. While our reproduced results are slightly lower than the original DualAnoDiff$^\dagger$ scores, they generally achieve higher performance than those obtained using the publicly available DualAnoDiff$^{*}$ model across most categories in both detection and classification tasks.

\vspace{-3mm}
\paragraph{DefectFill.}

To assess MAGIC under the same evaluation conditions as DefectFill~\cite{Song2025DefectFillRD}, we additionally conduct experiments following the protocol described in their paper. Specifically, we adopt the same data split, where the generator is trained using one-third of the available anomaly images and evaluated on the remaining two-thirds. We follow the experimental procedure exactly as outlined in the DefectFill paper to ensure comparability under identical conditions. While this setup enables a fair comparison under identical conditions, it is worth noting that DefectFill makes use of \textbf{ground-truth anomaly masks} from the test set, which may not be available in practical deployment scenarios. Moreover, the paper does not report results for categories that exhibit significant variation in object poses (e.g., \textit{screw} and \textit{metal\_nut}), which are included in our evaluation. To ensure consistency, we follow the same category selection as described in their work and disable CAMA during this evaluation.

In addition, the reported performance of DefectFill is based on images selected using a Low-Fidelity Selection (LFS) strategy, which filters generated samples based on their LPIPS similarity to real anomalies. In contrast, we evaluate MAGIC on all generated images without applying any filtering. As shown in Table~\ref{tab:defectfill_comparison}, DefectFill yields lower KID scores, suggesting slightly higher fidelity. Meanwhile, MAGIC consistently achieves higher IC-LPIPS scores—approximately 0.02 higher—indicating greater diversity. This diversity may contribute to improved downstream performance: Table~\ref{tab:defectfill_comparison} shows a 2.19\% improvement in classification accuracy. In contrast, the detection results in Table~\ref{tab:detection_defectfill_magic} show that DefectFill and MAGIC achieve identical scores across AUP-P, AP-P, and all other reported metrics. It is important to note that the evaluation protocol used in DefectFill directly leverages ground-truth anomaly masks from the test set during generation. As a result, the generated anomalies are inherently well-aligned with real defects, which can naturally lead to lower KID scores and improved classification and detection metrics. In contrast, the setting adopted in our main paper avoids reliance on test annotations and instead uses only training data to guide generation, which we believe provides a fairer and more practical framework for evaluating generalization to unseen anomalies.
\vspace{-3mm}
\paragraph{SeaS.}
Unlike other baselines and our method, SeaS applies different manually tuned CFG scales per dataset for anomaly image generation. For SeaS, we reproduced the baseline using the official implementation released by the authors and followed the recommended guidance scale (CFG\_scale) values exactly as described in the main paper~\cite{choi26cvprf}. Specifically, we used a CFG\_scale of 8 for MVTecAD, 2 for VisA, and 5 for MVTec 3D-AD. Since the DAGM dataset was not addressed in the original SeaS paper, we applied the same CFG\_scale used for MVTecAD (i.e., 8). In contrast, all other baselines, including MAGIC, were evaluated using the \textbf{same hyperparameters} to ensure a fair and consistent comparison.

\section{Analysis on artifact boundary}
\label{sec:artifact boundary}

We examined the visual continuity of synthesized defects by measuring the mean gradient magnitude ($|\nabla I|$) along mask boundaries. As shown in Tab.~\ref{tab:boundary_grad_mean}, the average gradient of our generated images (5.59) remains lower than that of real images (10.41). These results suggest that our framework does not introduce systematic boundary artifacts, maintaining a natural transition.

\begin{table}[!h]
\centering
\vspace{-3.5mm}
\begin{minipage}[b]{0.4\columnwidth}
\centering
\captionsetup{font=scriptsize}
\scriptsize
\setlength{\tabcolsep}{3pt}
\renewcommand{\arraystretch}{1.0}
\begin{tabular}{lc}
\hline
 & Boundary $|\nabla I|$ \\
\hline
Real & 10.41 \\
Generated & 5.59 \\
\hline
\end{tabular}
\vspace{-1.0mm}
\caption{Mean image gradient magnitude ($|\nabla I|$) in mask boundary pixels on MVTec-AD.}
\label{tab:boundary_grad_mean}
\end{minipage}
\hfill
\begin{minipage}[b]{0.5\columnwidth}
\centering
\captionsetup{font=scriptsize}
\tiny
\setlength{\tabcolsep}{2pt} 
\renewcommand{\arraystretch}{0.95}
\begin{tabular}{lcc}
\hline
Method & KID$\downarrow$ & Cls. acc.$\uparrow$ \\
\hline
CADS(s=0.1)   & 74.41  & 55.52 \\
CADS(s=0.05)   & 74.76  & 53.65 \\
CADS(s=0.025)   & 73.98  & 54.67 \\
DIAGEN & 58.95 & 64.87 \\
\textbf{Ours (GPP)} & \textbf{42.59}  & \textbf{74.36} \\
\hline
\end{tabular}
\vspace{-1.0mm}
\caption{Comparison vs other prompt augmentation baselines on MVTec-AD.}
\label{tab:prompt_aug_baselines}
\end{minipage}
\end{table}

\vspace{-3mm}
\section{Comparison with other prompt augmentation baselines}
While natural-language prompt diversification (noun/adjective swaps) is less applicable in our setting because the anomaly concept is represented by a fixed learned token (``sks''), we nevertheless compare against prompt augmentation baselines CADS~\cite{Sadat2023CADSUT} and DIAGEN~\cite{lingenberg2024diagen} and show GPP yields substantially better KID and downstream accuracy (Tab.~\ref{tab:prompt_aug_baselines}). 

\section{Ablation study on SAG}
We evaluated various \textbf{SAG} scheduling functions—\textit{Linear}, \textit{Exp}, \textit{Poly}, and \textit{Cosine}. As shown in Tab.~\ref{tab:scheduler_ablation}, the \textbf{Cosine} scheduler yields the best classification accuracy (78.06\%), suggesting that a gradual decay in guidance strength is beneficial for our framework.

\begin{table}[!h]
\vspace{-3.0mm}
\centering
\small 
\setlength{\tabcolsep}{9pt}
\captionsetup{font=small} 
\renewcommand{\arraystretch}{1.0} 
\begin{tabular}{lcccc}
\hline
 & Linear & Exp & Poly & Cosine (ours) \\
\hline
\textbf{Cls acc.} & \underline{76.73} & 75.71 & 75.51 & \textbf{78.06} \\
\hline
\end{tabular}
\vspace{-3.0mm}
\caption{Ablation of schedulers in SAG.}
\label{tab:scheduler_ablation}
\vspace{-5mm}
\end{table}

\section{Ablation of CAMA design choices.} 
CAMA uses two boundary points because single-point correspondence is unstable when the center region is weakly distinctive (e.g., ``missing part'').
The upper and lower keypoints are obtained by drawing a vertical line crossing the mask centroid.
The ``vertical-line'' rule is just for selecting two separated boundary keypoints.
Horizontal vs vertical is comparable (Tab.~\ref{tab:cand_line_horizontal}), but center-only is worse (Tab.~\ref{tab:edge_points_ablation}), validating the need for edge points.

\begin{table}[!h]
\vspace{-1mm}
\centering
\scriptsize
\begin{minipage}[t]{0.54\columnwidth}
\centering
\captionsetup{font=scriptsize}
\begin{tabular}{lcc}
\hline
 & Horizontal & Vertical (ours) \\
\hline
Cls acc. & 78.02 & \textbf{78.06} \\
\hline
\end{tabular}
\vspace{-1mm}
\caption{Ablation on candidate line direction.}
\label{tab:cand_line_horizontal}
\end{minipage}
\hfill
\begin{minipage}[t]{0.45\columnwidth}
\centering
\scriptsize
\captionsetup{font=scriptsize}
\begin{tabular}{lcc}
\hline
& Center-only & Ours \\
\hline
Cls acc. & 75.68 & \textbf{78.06} \\
\hline
\end{tabular}
\vspace{-1mm}
\caption{Ablation on edge points.}
\label{tab:edge_points_ablation}
\end{minipage}
\vspace{-5mm}
\end{table}

\section{Ablation study on the VisA dataset} 
We add a module ablation on VisA (Tab.~\ref{tab:ablation_mvtec_compact}), showing consistent gains.

\begin{table}[h!]
\vspace{-2mm} 
\centering
\small 
\captionsetup{font=small}
\setlength{\tabcolsep}{6pt} 
\renewcommand{\arraystretch}{1.0} 
\begin{tabular}{cccccc}
\hline
GPP & SAG & CAMA & KID$\downarrow$ & IC-L$\uparrow$ & Acc$\uparrow$ \\
\hline
$\times$ & $\times$ & $\times$ & 81.76 & 0.326 & 61.34 \\
$\checkmark$ & $\times$ & $\times$ & 81.11 & \underline{0.333} & 63.37 \\
$\checkmark$ & $\checkmark$ & $\times$ & \underline{80.11} & \textbf{0.336} & \underline{65.34} \\
\hline
$\checkmark$ & $\checkmark$ & $\checkmark$ & \textbf{79.81} & 0.323 & \textbf{68.51} \\
\hline
\end{tabular}
\vspace{-1mm}
\caption{Ablation of MAGIC components on VisA.}
\label{tab:ablation_mvtec_compact}
\vspace{-3mm}
\end{table}

\section{Ablation of downstream task model}
We evaluated the generalization of \textbf{MAGIC} using ResNet-18 as a downstream classifier. As shown in Tab.~\ref{tab:comparison}, the performance gains observed with ResNet-34 persist even with this smaller architecture. This consistent improvement suggests that our method captures robust anomaly representations regardless of the specific backbone.


\begin{table}[!h]
\vspace{-4mm}
\centering
\footnotesize 
\captionsetup{font=footnotesize}
\setlength{\tabcolsep}{4pt} 
\begin{tabular}{lccccc}
\hline
 & AnoDiff & AnoGen & DualAnoDiff & SeaS & \textbf{MAGIC} \\
\hline
Cls acc. (\%) & 66.14 & 57.98 & \underline{69.20} & 58.68 & \textbf{77.31} \\
\hline
\end{tabular}
\vspace{-2mm} 
\caption{Mean anomaly classification accuracies using ResNet-18 on MVTec-AD.}
\label{tab:comparison}
\vspace{-2mm}
\end{table}

\section{Failure cases}
\label{sec:failure cases}

\begin{figure}[t]
    \centering
    \includegraphics[width=\linewidth]{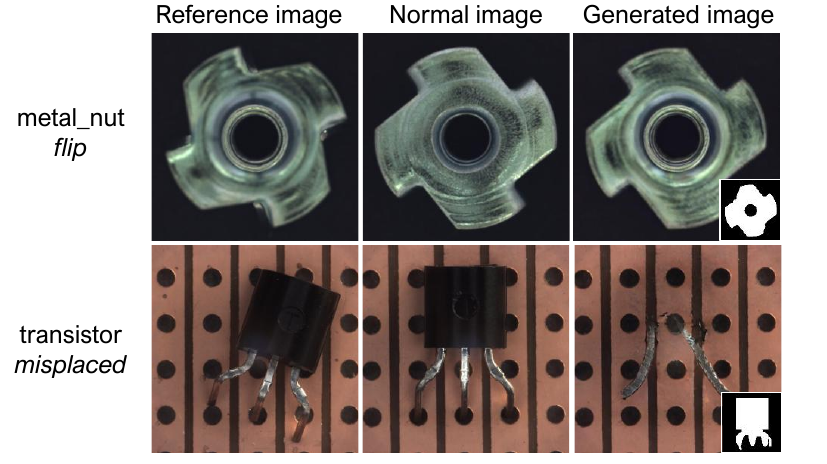}
    \caption{
        Failure cases of our method. 
        The examples illustrate situations involving large structural changes or global object displacements, where MAG-based generation fails to produce plausible anomalies.
    }
    \label{fig:failure_cases}
\end{figure}

As illustrated in Fig.~\ref{fig:failure_cases}, MAGIC is a mask-guided anomaly generation (MAG) method that synthesizes anomalies by inpainting defect regions on top of normal images. Due to this design choice, MAG inherently struggles with defects that involve large structural changes or global object displacements. For instance, categories such as \textit{metal\_nut (flip)} and \textit{transistor (misplaced)} exhibit object-level transformations that cannot be captured as localized inpainting regions, making them particularly challenging for MAG-based approaches to generate.

\section{Category-wise detection results}
\label{sec:detection_results}

We provide category-wise detection results for all benchmark datasets, including MVTecAD, VisA, DAGM, and MVTec 3D-AD. Detailed per-category performance is summarized in Tab.~\ref{tab:image_detection_mvtec},~\ref{tab:pixel_detection_mvtec} (MVTec-AD),~\ref{tab:image_detection_visa},~\ref{tab:pixel_detection_visa} (VisA), \ref{tab:image_detection_dagm},~\ref{tab:pixel_detection_dagm} (DAGM), \ref{tab:image_detection_3d},~\ref{tab:pixel_detection_3d} (MVTec 3D-AD), offering a comprehensive view of the model’s detection capability across diverse anomaly types and object categories.

\section{Qualitative results of generated anomalies}
\label{sec:generated_image}

\begin{figure*}[t]
    \centering
    \includegraphics[width=1.0\textwidth]{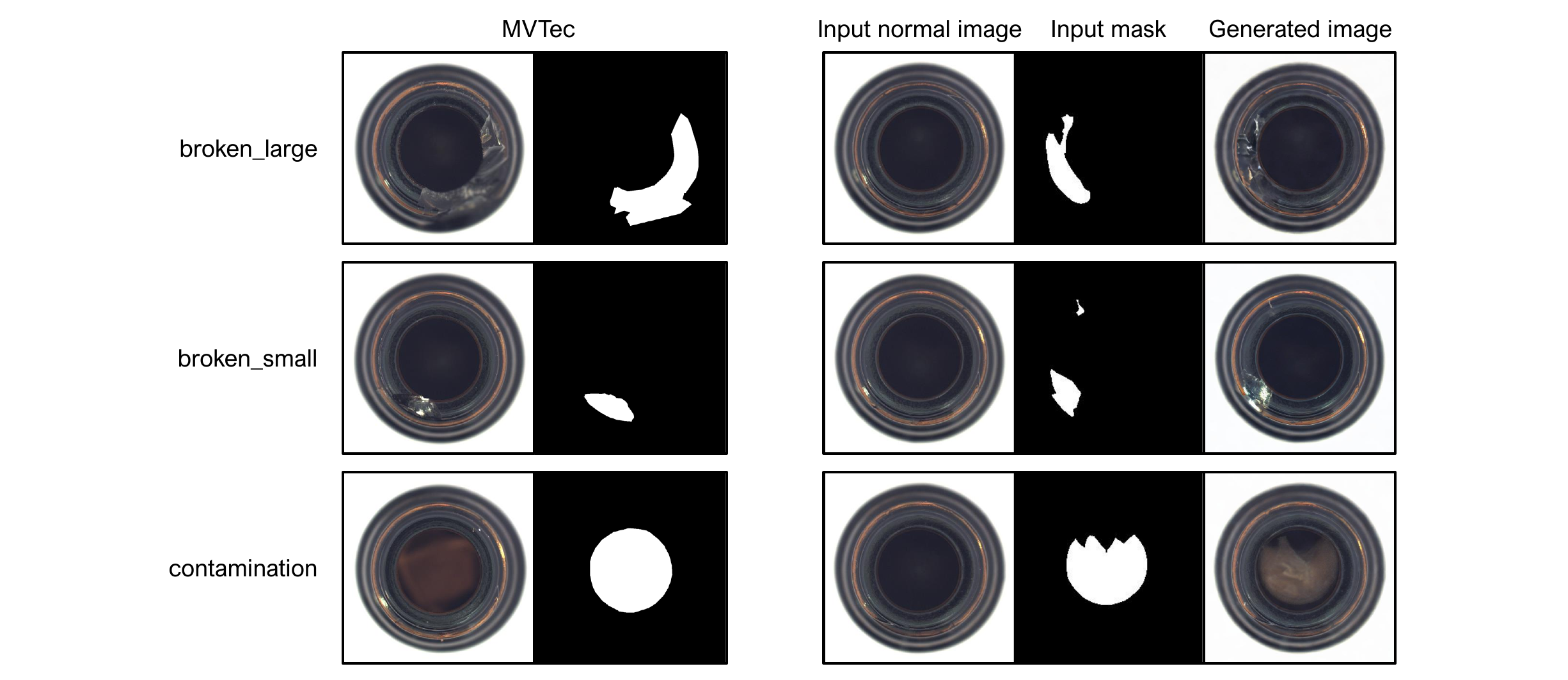}
    \vspace{-4mm}
    \caption{Generated images on \textit{bottle}
    }
    \vspace{-1mm}
    \label{fig:bottle_fig}
\end{figure*}

\begin{figure*}[t]
  \centering
  \includegraphics[width=0.9\textwidth]{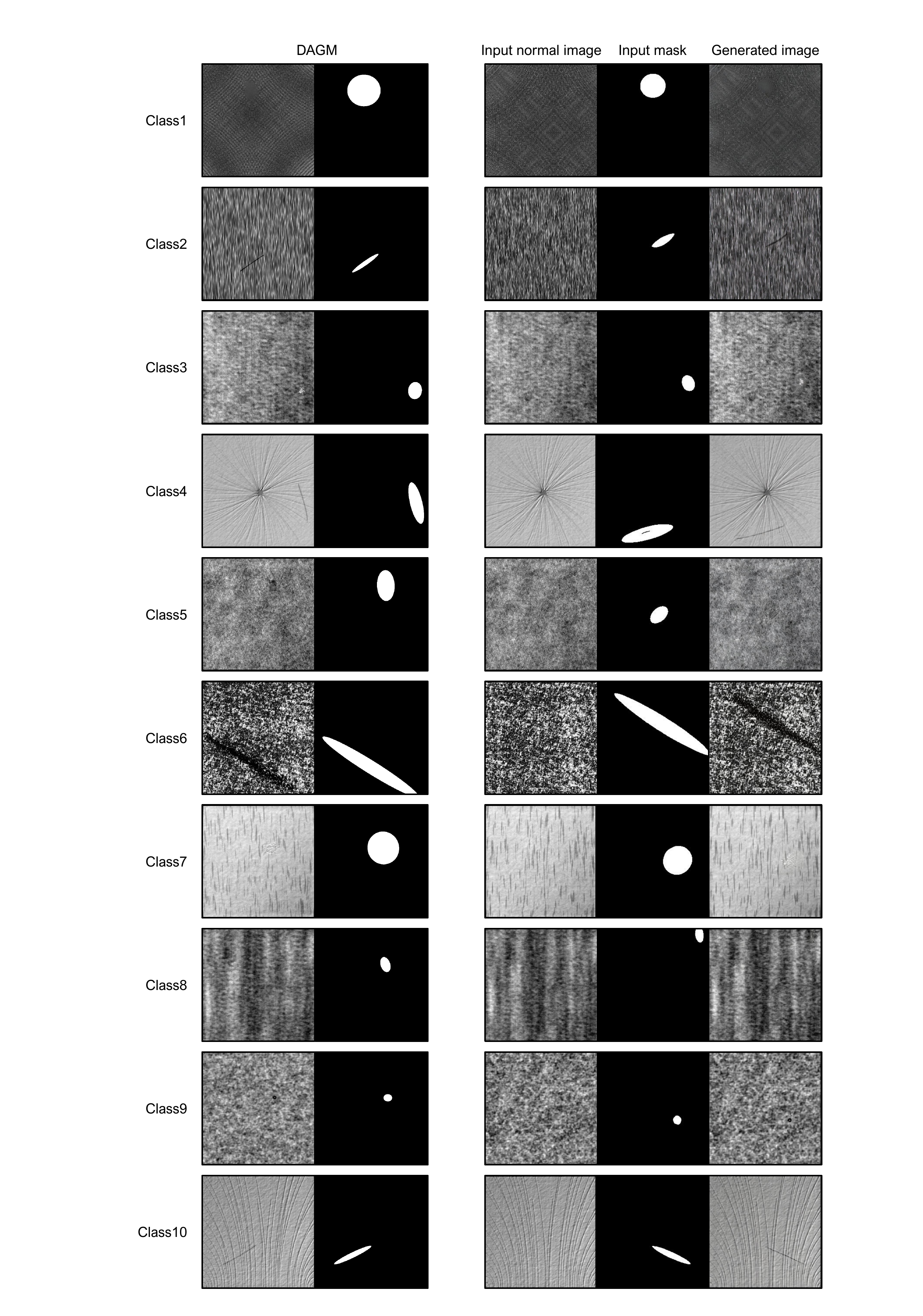}
  \vspace{-4mm}
    \caption{Generated images on every classes of \textit{DAGM}
    }
  \vspace{-10mm}
  \label{fig:DAGM_fig}
\end{figure*}

Fig.~\ref{fig:bottle_fig} to \ref{fig:DAGM_fig} present qualitative results of the anomalies generated by our method.

\clearpage

\begin{table*}[t]
    \centering
    \small
    \setlength{\tabcolsep}{4pt}
    \resizebox{\textwidth}{!}{%
    \begin{tabular}{c|ccc|ccc|ccc|ccc|ccc}
    \toprule
    \multirow{2}{*}{Category}
    & \multicolumn{3}{c|}{AnomalyDiffusion}
    & \multicolumn{3}{c|}{AnoGen}
    & \multicolumn{3}{c|}{DualAnoDiff}
    & \multicolumn{3}{c|}{SeaS}
    & \multicolumn{3}{c}{MAGIC (ours)} \\
    & AUROC & AP & $F_1$-max
    & AUROC & AP & $F_1$-max
    & AUROC & AP & $F_1$-max
    & AUROC & AP & $F_1$-max
    & AUROC & AP & $F_1$-max \\
    \midrule
    bottle     & \underline{99.7} & \underline{99.8} & \underline{98.9} & 99.3 & 99.7 & 97.6 & \textbf{100} & \textbf{100} & \textbf{100} & \textbf{100} & \textbf{100} & \textbf{100} & \textbf{100} & \textbf{100} & \textbf{100} \\
    cable      & \textbf{99.8} & \textbf{99.9} & \textbf{98.4} & 97.9 & 98.6 & 94.4 & 98.2 & 99 & \underline{96.8} & 97.2 & 97.9 & 94.4 & \underline{99.1} & \underline{99.3} & 96.1 \\
    capsule    & 96.5 & 98.9 & 95.2 & \underline{97.5} & \underline{99.3} & 95.5 & 96.5 & 99 & 94.6 & 96.9 & 99.2 & \underline{95.9} & \textbf{98.4} & \textbf{99.5} & \textbf{96.6} \\
    carpet     & 92.9 & 97.5 & 93.2 & 95.2 & 98.1 & 93.4 & \underline{99} & \underline{99.6} & \textbf{98.4} & \textbf{99.4} & \textbf{99.7} & \underline{96.9} & 96.8 & 98.7 & 94.4 \\
    grid       & 98.1 & 99.3 & \underline{98.7} & 97.6 & 99.2 & \underline{98.7} & 98.3 & 99.3 & 97.4 & \underline{99.3} & \underline{99.7} & 97.5 & \textbf{99.9} & \textbf{99.9} & \textbf{98.8} \\
    hazelnut   & \underline{99.9} & \textbf{100} & \underline{99} & 99.6 & 99.7 & 97.9 & \underline{99.9} & \underline{99.9} & 98.9 & 99.8 & \underline{99.9} & 98 & \textbf{100} & \textbf{100} & \textbf{100} \\
    leather    & \textbf{100} & \textbf{100} & \textbf{100} & \textbf{100} & \textbf{100} & \textbf{100} & \textbf{100} & \textbf{100} & \textbf{100} & \textbf{100} & \textbf{100} & \underline{99.2} & \textbf{100} & \textbf{100} & \textbf{100} \\
    metal\_nut & \textbf{100} & \textbf{100} & \textbf{100} & 99.7 & \underline{99.9} & 98.4 & 99.3 & \textbf{100} & 97.7 & \underline{99.9} & \textbf{100} & \underline{99.2} & \textbf{100} & \textbf{100} & \textbf{100} \\
    pill       & 97.9 & 99.5 & 96.8 & 96.4 & 99.1 & 95.3 & 89.7 & 97.4 & 90.2 & \textbf{98.7} & \textbf{99.7} & \underline{96.9} & \underline{98.3} & \underline{99.6} & \textbf{97.4} \\
    screw      & 93.5 & 96.9 & \underline{91.7} & 92.4 & 96.1 & 90.7 & 81.9 & 91.3 & 82.9 & \underline{94.6} & \underline{97.6} & 90.8 & \textbf{98.8} & \textbf{99.4} & \textbf{96.4} \\
    tile       & \textbf{100} & \textbf{100} & \textbf{100} & \textbf{100} & \textbf{100} & \textbf{100} & \textbf{100} & \textbf{100} & \textbf{100} & \underline{99.8} & \underline{99.9} & \underline{99.1} & \textbf{100} & \textbf{100} & \textbf{100} \\
    toothbrush & \underline{98.8} & \underline{99.3} & \underline{97.4} & \textbf{100} & \textbf{100} & \textbf{100} & 92.9 & 96.3 & 88.9 & \textbf{100} & \textbf{100} & \textbf{100} & \textbf{100} & \textbf{100} & \textbf{100} \\
    transistor & \textbf{100} & \textbf{100} & \textbf{100} & \underline{99.4} & \underline{98.8} & 94.7 & 99.1 & 98.1 & \underline{94.9} & 97.2 & 96.4 & 94.3 & \textbf{100} & \textbf{100} & \textbf{100} \\
    wood       & \textbf{99.9} & \textbf{99.9} & \textbf{98.8} & 99 & 99.5 & \textbf{98.8} & \textbf{99.9} & \textbf{99.9} & \textbf{98.8} & 98.1 & 99.3 & \underline{97.6} & \underline{99.1} & \underline{99.6} & \underline{97.6} \\
    zipper     & \textbf{100} & \textbf{100} & \underline{99.4} & \textbf{100} & \textbf{100} & \textbf{100} & \textbf{100} & \textbf{100} & \textbf{100} & \textbf{100} & \textbf{100} & \textbf{100} & \textbf{100} & \textbf{100} & \textbf{100} \\
    \midrule
    
    Average    
    & 98.47 & \underline{99.40} & 97.25 
    & 98.27 & 99.20 & 96.89
    & 97.09 & 98.64 & 95.29 & 
    \underline{98.73} & 99.29 & \underline{97.32} & 
    \textbf{99.36} & \textbf{99.73} & \textbf{98.49} \\
    \bottomrule
    \end{tabular}
    }
    \vspace{-2mm}

    \caption{Category-wise quantitative comparison of image-level anomaly localization on the MVTec-AD dataset.}
    \label{tab:image_detection_mvtec}
\end{table*}

\begin{table*}[t]
    \centering
    \small
    \setlength{\tabcolsep}{4pt}
    \resizebox{\textwidth}{!}{%
    \begin{tabular}{c|cccc|cccc|cccc|cccc|cccc}
    \toprule
    \multirow{2}{*}{Category}
    & \multicolumn{4}{c|}{AnomalyDiffusion}
    & \multicolumn{4}{c|}{AnoGen}
    & \multicolumn{4}{c|}{DualAnoDiff}
    & \multicolumn{4}{c|}{SeaS}
    & \multicolumn{4}{c}{MAGIC (ours)} \\
    & AUROC & AP & $F_1$-max & PRO
    & AUROC & AP & $F_1$-max & PRO
    & AUROC & AP & $F_1$-max & PRO
    & AUROC & AP & $F_1$-max & PRO
    & AUROC & AP & $F_1$-max & PRO \\
    \midrule
    bottle     & 99.3 & \underline{92.6} & 84.6 & 95.8 & 98 & 83 & 74.5 & 89.9 & 98.7 & 88.4 & 81.4 & 91.8 & \underline{99.5} & \underline{92.6} & \underline{86.8} & \underline{97} & \textbf{99.7} & \textbf{95.4} & \textbf{88.5} & \textbf{97.6} \\
    cable      & \textbf{98.4} & \textbf{84.3} & \textbf{76.3} & \underline{93.6} & 94.8 & 68.3 & 64.7 & 88.6 & 91.3 & 72.9 & 69.9 & 80.9 & 96 & 76.7 & 73 & 90.1 & \underline{96.5} & \underline{81.5} & \underline{75.4} & \textbf{93.9} \\
    capsule    & 97.9 & 41.6 & 43.7 & 89.7 & 96.3 & 34 & 37.5 & 92.4 & \textbf{98.8} & \underline{47.8} & \underline{52.8} & \textbf{95.5} & 96.8 & 44 & 46 & 94.7 & \underline{98.3} & \textbf{61.7} & \textbf{60.8} & \underline{95.2} \\
    carpet     & 96.4 & 74.8 & 71.8 & 83.6 & 98.8 & 81.4 & 74 & 89.6 & \underline{99.1} & \textbf{87.1} & \textbf{78.5} & 94.5 & \textbf{99.3} & 85.3 & 76.4 & \textbf{95.1} & \underline{99.1} & \underline{86.5} & \underline{78.4} & \underline{94.6} \\
    grid       & 97.3 & 43.6 & 46 & 93.6 & 96.1 & 33.1 & 41.8 & 93.9 & 95.4 & 54.9 & 55 & 93.2 & \underline{98.9} & \textbf{63.9} & \textbf{61} & \underline{97.2} & \textbf{99.5} & \underline{59.3} & \underline{59} & \textbf{97.8} \\
    hazelnut   & 99.3 & 89.4 & 81.9 & 94.5 & 97.2 & 58.3 & 56.7 & 93.1 & 99.5 & 88.7 & 81 & 95.4 & \underline{99.6} & \underline{90.1} & \underline{82.7} & \textbf{98.1} & \textbf{99.8} & \textbf{95.9} & \textbf{90.2} & \underline{97.3} \\
    leather    & \textbf{99.8} & 78.4 & 71 & \underline{98.3} & 99.4 & 77.7 & 70.4 & 98 & \textbf{99.8} & \textbf{82.8} & \underline{73.6} & \textbf{98.4} & 99.4 & 77.1 & 69 & 98.1 & \underline{99.6} & \underline{82.2} & \textbf{74.4} & 98 \\
    metal\_nut & 99.6 & 97.8 & 92.5 & 95.2 & 95.2 & 77 & 72.5 & 89.5 & 99.5 & 96.7 & 91.8 & 92.5 & \underline{99.7} & \underline{98.5} & \underline{94.1} & \underline{95.6} & \textbf{99.8} & \textbf{99} & \textbf{95.3} & \textbf{96.4} \\
    pill       & \textbf{99.6} & \textbf{95} & \textbf{88.5} & 94.4 & 99.2 & 89.1 & 80.7 & 93.2 & 97.1 & 83 & 77.9 & 78.4 & \underline{99.5} & \underline{90.5} & \underline{82.9} & \underline{97.1} & \textbf{99.6} & 90.2 & 82.4 & \textbf{97.5} \\
    screw      & 95.6 & 11.8 & 21.6 & 83.8 & 92.7 & 21.9 & 27.2 & 77.5 & \underline{98.2} & \underline{50.4} & \underline{51.2} & \underline{91.5} & 96.3 & 48.3 & 49.8 & 90.5 & \textbf{99.2} & \textbf{51.4} & \textbf{52.1} & \textbf{94.9} \\
    tile       & 99.5 & 96 & 88.9 & 96.9 & 99.2 & 93.1 & 85 & 97 & 99.6 & 95.4 & 89.2 & \underline{97.5} & \underline{99.7} & \underline{96.8} & \underline{91.4} & \underline{97.5} & \textbf{99.8} & \textbf{97.9} & \textbf{92.5} & \textbf{98.3} \\
    toothbrush & \underline{97.9} & 56.6 & 58.5 & 84.9 & \textbf{99.2} & \textbf{93.1} & \textbf{85} & 72.2 & 93.8 & 56.2 & 60.9 & 71.1 & 97.1 & 66 & 65.8 & \underline{86} & \textbf{99.2} & \underline{71.2} & \underline{74.5} & \textbf{93.5} \\
    transistor & \textbf{99.2} & \textbf{89.1} & \textbf{81.2} & \textbf{96.4} & 95 & 63.2 & 61.2 & 88.8 & 92.5 & 73.9 & 71.1 & 89.8 & 97.8 & 83.7 & 77 & \underline{92.6} & \underline{98.9} & \underline{84.9} & \underline{78.3} & \textbf{96.4} \\
    wood       & 96.7 & 74.6 & 67.4 & 91.3 & \underline{98.5} & 78.2 & 71.5 & 94 & 98.3 & \underline{84.6} & \underline{77.2} & \underline{94.5} & 96.7 & 80.4 & 74.8 & 90.8 & \textbf{98.8} & \textbf{85.1} & \textbf{77.4} & \textbf{94.7} \\
    zipper     & \underline{99.4} & 84.7 & 78.2 & \underline{96.6} & 99.1 & 78.9 & 71.8 & 95.6 & \textbf{99.5} & \textbf{89} & \textbf{82} & \underline{96.6} & 98.8 & 81.1 & 74.9 & 95.9 & \textbf{99.5} & \underline{87.6} & \underline{80.8} & \textbf{97.2} \\
    \midrule
    Average 
    & \underline{98.39} & 74.02 & 70.14 & 92.57 
    & 96.25 & 64.20 & 61.60 & 90.65 
    & 97.41 & 76.79 & 72.90 & 91.32
    & 98.34 & \underline{78.33} & \underline{73.71} & \underline{94.42}
    & \textbf{99.15} & \textbf{81.99} & \textbf{77.33} & \textbf{96.22} \\
    \bottomrule
    \end{tabular}
    }
    \vspace{-2mm}
    \caption{Category-wise quantitative comparison of pixel-level anomaly localization on the MVTec-AD dataset.}
    \label{tab:pixel_detection_mvtec}
\end{table*}

\begin{table*}[t]
    \centering
    \small
    \setlength{\tabcolsep}{4pt}
    \resizebox{\textwidth}{!}{%
    \begin{tabular}{c|ccc|ccc|ccc|ccc|ccc}
    \toprule
    \multirow{2}{*}{Category}
    & \multicolumn{3}{c|}{AnomalyDiffusion}
    & \multicolumn{3}{c|}{AnoGen}
    & \multicolumn{3}{c|}{DualAnoDiff}
    & \multicolumn{3}{c|}{SeaS}
    & \multicolumn{3}{c}{MAGIC (ours)} \\
    & AUROC & AP & $F_1$-max
    & AUROC & AP & $F_1$-max
    & AUROC & AP & $F_1$-max
    & AUROC & AP & $F_1$-max
    & AUROC & AP & $F_1$-max \\
    \midrule
    candle
    & \underline{93.6} & \textbf{93.1} & \textbf{87}
    & 87.1 & 84.4 & 76
    & 86.6 & 80.9 & 74.5
    & 78.4 & 76.6 & 66
    & \textbf{94.5} & \underline{91.3} & \underline{84} \\
    
    capsules
    & 63.9 & 44.8 & 46.3
    & 63.6 & 37.5 & 52.2
    & 89.8 & 75.8 & 71.1
    & \textbf{97.3} & \textbf{96.3} & \textbf{92.7}
    & \underline{93.5} & \underline{86.1} & \underline{78.3} \\
    
    cashew
    & 96.4 & \underline{97.1} & 93
    & \textbf{98.7} & \textbf{99} & \textbf{95.3}
    & 94.4 & 95.7 & 87.9
    & 85.6 & 88.6 & 82.2
    & 96 & 96.4 & \underline{93.8} \\
    
    chewinggum
    & 94.6 & 95.3 & 89.7
    & 95.1 & 95.6 & 90.1
    & 93.3 & 91.5 & 84.6
    & \underline{96.1} & \underline{96.5} & \underline{93.2}
    & \textbf{99.3} & \textbf{99.1} & \textbf{96.3} \\
    
    fryum
    & \textbf{91.5} & \textbf{93.7} & \textbf{86.8}
    & \underline{88} & \underline{92.4} & 83.3
    & 87.6 & 90.7 & 82.7
    & 86.5 & 89.5 & \underline{84.8}
    & 82.8 & 86.8 & 81.2 \\
    
    macaroni1
    & 82.6 & 72.2 & 69.4
    & \underline{98.9} & 98.1 & 93.6
    & 87.4 & 82.8 & 71.1
    & 94.5 & 89.2 & 86
    & \textbf{99.5} & \textbf{99.3} & \textbf{96.4} \\
    
    macaroni2
    & 45.8 & 30.9 & 52.8
    & \underline{78.4} & \underline{71.5} & \underline{66.7}
    & 74.3 & 55.8 & 63
    & 75.1 & 70.7 & 62.3
    & \textbf{91.8} & \textbf{90.4} & \textbf{82.4} \\
    
    pcb1
    & \textbf{95.7} & \textbf{94.3} & 84.2
    & \underline{94.9} & \underline{93.7} & \textbf{85.7}
    & 78.6 & 80.3 & 71.3
    & 92.7 & 90.5 & 79.7
    & 91.3 & 86.6 & \underline{82.2} \\
    
    pcb2
    & 94.6 & 94.1 & 88.3
    & 97 & 96.2 & 89.8
    & 88.3 & 88.6 & 81.1
    & \underline{93.1} & \underline{92.8} & \underline{84.4}
    & \textbf{97.3} & \textbf{97} & \textbf{94.2} \\
    
    pcb3
    & 85.3 & 85.7 & 78.9
    & \underline{91.7} & \underline{88.3} & 82.7
    & 77.5 & 74.9 & 65.4
    & 85.4 & 84.8 & 76.5
    & \textbf{94.3} & \textbf{92.4} & \textbf{87.3} \\
    
    pcb4
    & \underline{97.6} & 89.2 & 84.7
    & 94.7 & 79.8 & 77.6
    & 88.6 & 73.8 & 66.7
    & 95.9 & 88.8 & \underline{87.5}
    & \textbf{98.8} & \textbf{93.3} & \textbf{94.3} \\
    
    pipe\_fryum
    & 78.7 & 84.5 & 75.5
    & 87.6 & 91.2 & 82.3
    & 91 & 93.8 & \underline{86.2}
    & 82.7 & 85 & 82.8
    & \textbf{92.3} & \textbf{94.9} & \textbf{87} \\
    
    \midrule
    
    Average
    & 85.03 & 81.24 & 78.05
    & \underline{89.64} & 85.64 & 81.28
    & 86.45 & 82.05 & 75.47
    & 88.61 & \underline{87.44} & \underline{81.51}
    & \textbf{94.28} & \textbf{92.8} & \textbf{88.12} \\
    \bottomrule
    \end{tabular}
    }
    \vspace{-2mm}
    \caption{Category-wise quantitative comparison of image-level anomaly localization on the VisA dataset.}
    \label{tab:image_detection_visa}
\end{table*}

\begin{table*}[t]
    \centering
    \small
    \setlength{\tabcolsep}{4pt}
    \resizebox{\textwidth}{!}{%
    \begin{tabular}{c|cccc|cccc|cccc|cccc|cccc}
    \toprule
    \multirow{2}{*}{Category}
    & \multicolumn{4}{c|}{AnomalyDiffusion}
    & \multicolumn{4}{c|}{AnoGen}
    & \multicolumn{4}{c|}{DualAnoDiff}
    & \multicolumn{4}{c|}{SeaS}
    & \multicolumn{4}{c}{MAGIC (ours)} \\
    & AUROC & AP & $F_1$-max & PRO
    & AUROC & AP & $F_1$-max & PRO
    & AUROC & AP & $F_1$-max & PRO
    & AUROC & AP & $F_1$-max & PRO
    & AUROC & AP & $F_1$-max & PRO \\
    \midrule
    candle
    & 98.4 & 33.8 & 39.4 & \underline{90.3}
    & 96.8 & 40.2 & 43.6 & 88.1
    & 92.1 & 18.3 & 28.3 & 85
    & 93.7 & \textbf{55.8} & \textbf{59.9} & 82.9
    & \textbf{99.3} & \underline{47.1} & \underline{46.7} & \textbf{96.4} \\
    
    capsules
    & 87.6 & 8 & 12.6 & 56.4
    & 92.3 & 3.8 & 11.4 & 68.8
    & 97.5 & 51.2 & 50.6 & 80.7
    & \textbf{99.9} & \underline{72.9} & \underline{65.1} & \underline{93.3}
    & \textbf{99.9} & \textbf{81.5} & \textbf{74.1} & \textbf{96} \\
    
    cashew
    & 99 & 83 & 78.1 & 81.5
    & 97.8 & 68.4 & 66 & \textbf{94.9}
    & \textbf{99.8} & \textbf{95} & \textbf{92.9} & \underline{92.4}
    & 98.1 & 85 & 81.5 & 83.9
    & \underline{99.5} & \underline{92} & \underline{88.7} & 86.6 \\
    
    chewinggum
    & 99.4 & 69.2 & 64.8 & 92
    & 99.4 & 80.8 & 73.9 & 90.1
    & 99.4 & 65.5 & 59.5 & 90.4
    & \textbf{99.8} & \textbf{88.8} & \textbf{80} & \textbf{96.7}
    & \underline{99.5} & 30 & 55.9 & \underline{96.5} \\
    
    fryum
    & 96.8 & 53.3 & 51.8 & \textbf{94}
    & 97.1 & 55.5 & 53.5 & 89.6
    & \underline{97.6} & 55.3 & 55.9 & 91.1
    & \textbf{98.8} & \textbf{83.9} & \textbf{77.7} & 77.7
    & 97.2 & 54.3 & 52.4 & \underline{92.9} \\
    
    macaroni1
    & 92.6 & 3.4 & 9.6 & 83.8
    & 97.8 & 35.6 & 41.1 & 95.7
    & 94.5 & 5.2 & 11.1 & 90.2
    & \textbf{99} & 54.8 & \textbf{56.3} & 89.3
    & \underline{98.8} & \textbf{57} & \underline{54.4} & \textbf{96.3} \\
    
    macaroni2
    & 93 & 0.2 & 0.6 & 78
    & 97.8 & 8.5 & 13.3 & 92.4
    & 96.1 & 9.5 & 12.2 & 87.7
    & \underline{98.2} & \underline{14.2} & \underline{23.6} & 90.3
    & \textbf{99.3} & \textbf{40.4} & \textbf{44} & \textbf{95.7} \\
    
    pcb1
    & 94.9 & 71.4 & 70 & \textbf{88.9}
    & 97 & 69.9 & 66.8 & 87.3
    & 96.1 & 70.9 & 68.5 & 76.8
    & \textbf{98.4} & \textbf{86} & \textbf{84.5} & 84.2
    & \underline{97.9} & \underline{78.7} & \underline{76.7} & \underline{87.7} \\
    
    pcb2
    & 94.9 & 25.6 & 33.5 & 82.5
    & \textbf{97.3} & 31.3 & 39.4 & \textbf{90}
    & \underline{96.4} & \underline{34.4} & \underline{43.3} & \underline{85.1}
    & 93.3 & \textbf{54.7} & \textbf{55.9} & 79.8
    & 93.4 & 24.9 & 32.1 & 82.1 \\
    
    pcb3
    & \underline{97.5} & 28.9 & 31.9 & 81.5
    & \textbf{97.9} & 33.1 & 34.9 & \textbf{86.3}
    & 96.5 & 25.9 & 35 & 80.6
    & 96.8 & \textbf{68.2} & \textbf{65.8} & 74.1
    & 96.1 & \underline{35.4} & \underline{42.2} & \underline{81.9} \\
    
    pcb4
    & 98.3 & 66.4 & 62.1 & 92.3
    & 96.4 & 37.9 & 46.2 & 79.5
    & 98 & 68.5 & 68.9 & 85.4
    & \underline{99.4} & \textbf{82.9} & \textbf{78.4} & 91.9
    & \textbf{99.8} & \underline{71.3} & \underline{66.4} & \textbf{96.9} \\
    
    pipe\_fryum
    & \underline{99.7} & 90.6 & 84.4 & 88.3
    & 99.4 & 79.9 & 72.1 & 86.6
    & \underline{99.7} & \underline{93.8} & \underline{86.2} & \underline{95.4}
    & 99.2 & 88 & 82 & 86.7
    & \textbf{99.9} & \textbf{96.3} & \textbf{90} & \textbf{95.7} \\
    
    \midrule
    Average
    & 96.01 & 44.48 & 44.9 & 84.13
    & 97.25 & 45.41 & 46.85 & 87.44
    & 96.98 & 49.46 & 51.03 & 86.73
    & \underline{97.88} & \textbf{69.60} & \textbf{67.56} & \underline{85.90}
    & \textbf{98.38} & \underline{59.08} & \underline{60.30} & \textbf{92.06} \\
    \bottomrule
    \end{tabular}
    }
    \vspace{-2mm}
    \caption{Category-wise quantitative comparison of pixel-level anomaly localization on the VisA dataset.}
    \label{tab:pixel_detection_visa}
\end{table*}

\begin{table*}[t]
    \centering
    \small
    \setlength{\tabcolsep}{4pt}
    \resizebox{\textwidth}{!}{%
    \begin{tabular}{c|ccc|ccc|ccc|ccc|ccc}
    \toprule
    \multirow{2}{*}{Category}
    & \multicolumn{3}{c|}{AnomalyDiffusion}
    & \multicolumn{3}{c|}{AnoGen}
    & \multicolumn{3}{c|}{DualAnoDiff}
    & \multicolumn{3}{c|}{SeaS}
    & \multicolumn{3}{c}{MAGIC (ours)} \\
    & AUROC & AP & $F_1$-max
    & AUROC & AP & $F_1$-max
    & AUROC & AP & $F_1$-max
    & AUROC & AP & $F_1$-max
    & AUROC & AP & $F_1$-max \\
    \midrule
    Class1
    & \textbf{100} & \textbf{100} & \textbf{100}
    & \textbf{100} & \textbf{100} & \underline{99.7}
    & \textbf{100} & \textbf{100} & \underline{99.7}
    & \textbf{100} & \textbf{100} & \textbf{100}
    & \underline{99.9} & \underline{99.6} & 99.3 \\
    
    Class2
    & \underline{99.9} & \underline{99.8} & 97.9
    & \underline{99.9} & 99.7 & \underline{99.3}
    & \underline{99.9} & 99.7 & 98.6
    & \textbf{100} & \textbf{100} & \textbf{100}
    & \textbf{100} & \textbf{100} & \textbf{100} \\
    
    Class3
    & \textbf{100} & \textbf{100} & \underline{99.7}
    & \textbf{100} & \textbf{100} & \textbf{100}
    & \textbf{100} & \textbf{100} & \textbf{100}
    & \textbf{100} & \textbf{100} & \textbf{100}
    & \textbf{100} & \textbf{100} & \textbf{100} \\
    
    Class4
    & 99 & 98.2 & 94.7
    & 99.4 & 98.8 & 96.9
    & 95 & 94.1 & 91.4
    & \textbf{100} & \textbf{100} & \textbf{99.7}
    & \underline{99.8} & \underline{99.5} & \underline{99} \\
    
    Class5
    & 98.6 & 97.3 & 94.6
    & 86.5 & 80.3 & 73.6
    & \textbf{100} & \textbf{100} & \textbf{100}
    & \textbf{100} & \textbf{100} & \textbf{100}
    & \underline{99.5} & \underline{99} & \underline{96.9} \\
    
    Class6
    & \textbf{100} & \textbf{100} & \textbf{100}
    & \textbf{100} & \textbf{100} & \textbf{100}
    & \textbf{100} & \textbf{100} & \textbf{100}
    & \textbf{100} & \textbf{100} & \textbf{100}
    & \textbf{100} & \textbf{100} & \textbf{100} \\
    
    Class7
    & \underline{99.9} & \underline{99.9} & \underline{99.8}
    & \textbf{100} & \textbf{100} & \textbf{100}
    & \textbf{100} & \textbf{100} & \textbf{100}
    & \textbf{100} & \textbf{100} & \underline{99.8}
    & \textbf{100} & \textbf{100} & \textbf{100} \\
    
    Class8
    & 72.2 & 51.9 & 49.9
    & 82.7 & 71.8 & 63.7
    & \underline{93.4} & \underline{89.5} & \underline{83.1}
    & 53.1 & 25.6 & 37.2
    & \textbf{99.4} & \textbf{98.7} & \textbf{95.9} \\
    
    Class9
    & \textbf{100} & \underline{99.9} & \underline{99}
    & \underline{99.9} & 99.7 & 97.5
    & \textbf{100} & \textbf{100} & \textbf{100}
    & \textbf{100} & \textbf{100} & \textbf{100}
    & \textbf{100} & \textbf{100} & \textbf{100} \\
    
    Class10
    & \underline{99.9} & 99.8 & 99.2
    & \textbf{100} & \underline{99.9} & 99
    & \textbf{100} & \underline{99.9} & 99.7
    & \textbf{100} & \textbf{100} & \textbf{100}
    & \textbf{100} & \textbf{100} & \underline{99.8} \\
    \midrule
    
    Average
    & 96.95 & 94.68 & 93.48
    & 96.84 & 95.02 & 92.97
    & \underline{98.83} & \underline{98.32} & \underline{97.25}
    & 95.31 & 92.56 & 93.67
    & \textbf{99.86} & \textbf{99.68} & \textbf{99.09} \\
    \bottomrule
    \end{tabular}
    }
    \vspace{-2mm}
    \caption{Category-wise quantitative comparison of image-level anomaly localization on the DAGM dataset.}
    \label{tab:image_detection_dagm}
\end{table*}

\begin{table*}[t]
    \centering
    \small
    \setlength{\tabcolsep}{4pt}
    \resizebox{\textwidth}{!}{%
    \begin{tabular}{c|cccc|cccc|cccc|cccc|cccc}
    \toprule
    \multirow{2}{*}{Category}
    & \multicolumn{4}{c|}{AnomalyDiffusion}
    & \multicolumn{4}{c|}{AnoGen}
    & \multicolumn{4}{c|}{DualAnoDiff}
    & \multicolumn{4}{c|}{SeaS}
    & \multicolumn{4}{c}{MAGIC (ours)} \\
    & AUROC & AP & $F_1$-max & PRO
    & AUROC & AP & $F_1$-max & PRO
    & AUROC & AP & $F_1$-max & PRO
    & AUROC & AP & $F_1$-max & PRO
    & AUROC & AP & $F_1$-max & PRO \\
    \midrule
    Class1
    & \textbf{99.8} & 83.1 & 72.7 & \textbf{98.4}
    & \underline{99.6} & 86 & 76.1 & \underline{98.2}
    & 99.2 & \textbf{88} & \textbf{80.5} & 97.5
    & 98.9 & 86.1 & \underline{79.4} & 96.9
    & 99.5 & \underline{87.2} & 79.1 & 97.7 \\
    
    Class2
    & \textbf{99.6} & 70.9 & 66.2 & \textbf{97.9}
    & 98.4 & 64.4 & 61.9 & 97.5
    & \underline{99.3} & 69 & 64.7 & \underline{97.8}
    & \underline{99.3} & \underline{71.2} & \underline{66.4} & \textbf{97.9}
    & 99 & \textbf{77.7} & \textbf{71.7} & 97.1 \\
    
    Class3
    & \underline{99} & 74.4 & 68.3 & 96.7
    & 98.6 & 77.2 & 70.9 & 96.4
    & 98.9 & \textbf{83.2} & \textbf{75.7} & \underline{97.1}
    & 98.6 & 76 & 71.5 & 96.6
    & \textbf{99.5} & \underline{80.3} & \underline{73.2} & \textbf{97.7} \\
    
    Class4
    & 98.1 & 69.3 & 65.3 & 94
    & 97.3 & 69.7 & 65.7 & 93.9
    & 95.4 & 71.9 & 69.7 & 86.8
    & \textbf{99.6} & \textbf{84.5} & \textbf{77.4} & \textbf{97.6}
    & \underline{99.3} & \underline{79.4} & \underline{73.8} & \underline{96.8} \\
    
    Class5
    & 97.8 & 71 & 68.9 & 92.5
    & 92.7 & 47.7 & 51.8 & 80.4
    & 99.1 & \textbf{87.3} & \textbf{81.3} & \underline{96.7}
    & \textbf{99.5} & \underline{85} & \underline{77.1} & \textbf{97.4}
    & \underline{99.2} & 76.5 & 70.4 & \underline{96.7} \\
    
    Class6
    & 99.5 & 90.8 & 82.4 & 97.4
    & 99.6 & 90.7 & 81.8 & 97.4
    & \underline{99.8} & 95.1 & 88.1 & 97.9
    & \textbf{99.9} & \textbf{97.5} & \textbf{90.9} & \textbf{98.4}
    & \textbf{99.9} & \underline{96.5} & \underline{89.5} & \underline{98.2} \\
    
    Class7
    & 98.8 & 85.3 & 78.9 & 96.5
    & 96.8 & 80.3 & 77 & 93.5
    & \textbf{99.7} & \textbf{91.3} & \textbf{82.2} & \textbf{98.6}
    & \underline{99.4} & 85.9 & 77.9 & 97.7
    & \textbf{99.7} & \underline{89.3} & \underline{80.6} & \underline{98} \\
    
    Class8
    & 91 & 19.5 & 26.7 & 72.1
    & \underline{91.2} & 32.1 & 40.7 & 77.8
    & 89.2 & \underline{42.2} & \underline{48.4} & \underline{82.5}
    & 50.5 & 0.1 & 0.2 & 16.3
    & \textbf{98.7} & \textbf{59.1} & \textbf{58.8} & \textbf{97.1} \\
    
    Class9
    & \textbf{100} & 92.2 & 83.6 & 98.5
    & \textbf{100} & 85.6 & 75.7 & 98.2
    & \textbf{100} & \underline{97.5} & \textbf{90.8} & \underline{99}
    & \textbf{100} & 96 & \underline{88.1} & 98.8
    & \textbf{100} & \textbf{97.6} & \textbf{90.8} & \textbf{99.1} \\
    
    Class10
    & 96.1 & 66.8 & 63.7 & 94.3
    & \textbf{98.2} & 65.9 & 62.8 & \textbf{96.5}
    & \underline{97.1} & \textbf{77.4} & \textbf{71.7} & \underline{95.8}
    & 97 & 66.5 & 65.2 & 95.2
    & 96.7 & \underline{75.8} & \underline{70.5} & 95.4 \\
    \midrule

    Average
    & \underline{97.97} & 72.33 & 67.67 & 93.83
    & 97.24 & 69.96 & 66.44 & 92.98
    & 97.77 & \underline{80.29} & \underline{75.31} & \underline{94.97}
    & 94.27 & 74.88 & 69.41 & 89.28
    & \textbf{99.15} & \textbf{81.94} & \textbf{75.84} & \textbf{97.38} \\
    \bottomrule
    \end{tabular}
    }
    \vspace{-2mm}
    \caption{Category-wise quantitative comparison of pixel-level anomaly localization on the DAGM dataset.}
    \label{tab:pixel_detection_dagm}
\end{table*}

\begin{table*}[t]
    \centering
    \small
    \setlength{\tabcolsep}{4pt}
    \resizebox{\textwidth}{!}{%
    \begin{tabular}{c|ccc|ccc|ccc|ccc|ccc}
    \toprule
    \multirow{2}{*}{Category}
    & \multicolumn{3}{c|}{AnomalyDiffusion}
    & \multicolumn{3}{c|}{AnoGen}
    & \multicolumn{3}{c|}{DualAnoDiff}
    & \multicolumn{3}{c|}{SeaS}
    & \multicolumn{3}{c}{MAGIC (ours)} \\
    & AUROC & AP & $F_1$-max
    & AUROC & AP & $F_1$-max
    & AUROC & AP & $F_1$-max
    & AUROC & AP & $F_1$-max
    & AUROC & AP & $F_1$-max \\
    \midrule
    bagel
    & 94.5 & 98.1 & 93.4
    & 93.5 & 97.9 & 94.1
    & \textbf{98.7} & \textbf{99.6} & \textbf{97.5}
    & \underline{97.9} & \underline{99.3} & 96.7
    & 97.0 & 98.7 & \underline{96.8} \\
    
    cable\_gland
    & \textbf{98.9} & \textbf{99.6} & \textbf{97.5}
    & 88.5 & 94.7 & 91.8
    & 61.8 & 84.8 & 84.9
    & 91.2 & 96.8 & 92.1
    & \underline{98.2} & \underline{99.4} & \underline{96.6} \\
    
    carrot
    & \textbf{97.2} & \underline{99.2} & \textbf{96.7}
    & 94.1 & 98 & 94.1
    & 90.7 & 97.1 & 91.3
    & 92.6 & 97.2 & 94.4
    & \underline{97.0} & \textbf{99.4} & \underline{95.2} \\
    
    cookie
    & 88.9 & 94.7 & 91.8
    & 89.9 & 95.6 & \underline{91.9}
    & 84.3 & 93.7 & 85.7
    & 92.0 & \underline{97.2} & 90.2
    & \textbf{95.8} & \textbf{98.4} & \textbf{92.7} \\
    
    dowel
    & 64.4 & 86.4 & 84.8
    & \underline{83.7} & 94.3 & 86.3
    & 82.7 & 93.5 & \underline{88}
    & 79.8 & 92.8 & 85.2
    & \textbf{99.0} & \textbf{99.7} & \textbf{97.8} \\
    
    foam
    & \textbf{93.5} & \textbf{97.9} & \underline{91.6}
    & \underline{92.9} & \underline{97.5} & \textbf{92.7}
    & 82.1 & 93.7 & 86.6
    & 71.9 & 90.1 & 85.9
    & 91.2 & 97.0 & 90.4 \\
    
    peach
    & 79.7 & 90.3 & 85.7
    & 90.2 & 96.7 & 88.7
    & 88 & 95.3 & 89.7
    & \textbf{95.7} & \textbf{98.4} & \textbf{93.3}
    & \underline{91.8} & \underline{97.3} & \underline{90.2} \\
    
    potato
    & 71.4 & 87.8 & 87
    & \underline{74.1} & \underline{88.3} & \underline{87.1}
    & 62.9 & 83.1 & 85.3
    & 61.6 & 83.5 & 85.9
    & \textbf{86.9} & \textbf{95.8} & \textbf{87.8} \\
    
    rope
    & \underline{99.5} & \underline{99.7} & \underline{97.9}
    & 98.5 & 99.2 & 96.8
    & 98.1 & 99 & 96.7
    & \textbf{100.0} & \textbf{100.0} & \textbf{100.0}
    & 91.2 & 95.1 & 86.5 \\
    
    tire
    & 72.7 & 88.3 & 83.2
    & 75.6 & 88.6 & 85.5
    & 58.3 & 75.9 & 83.1
    & \textbf{89.7} & \textbf{96.0} & \textbf{88.7}
    & \underline{85.6} & \underline{94.7} & \underline{88.1} \\
    \midrule
    
    Average
    & 86.07 & 94.2 & 90.96
    & \underline{88.1} & 95.08 & 90.9
    & 80.76 & 91.57 & 88.88
    & 87.24 & \underline{95.13} & \underline{91.24}
    & \textbf{93.37} & \textbf{97.52} & \textbf{92.21} \\
    \bottomrule
    \end{tabular}
    }
    \vspace{-2mm}
    \caption{Category-wise quantitative comparison of image-level anomaly localization on the MVTec 3D-AD dataset.}
    \label{tab:image_detection_3d}
\end{table*}

\begin{table*}[t]
    \centering
    \small
    \setlength{\tabcolsep}{4pt}
    \resizebox{\textwidth}{!}{%
    \begin{tabular}{c|cccc|cccc|cccc|cccc|cccc}
    \toprule
    \multirow{2}{*}{Category}
    & \multicolumn{4}{c|}{AnomalyDiffusion}
    & \multicolumn{4}{c|}{AnoGen}
    & \multicolumn{4}{c|}{DualAnoDiff}
    & \multicolumn{4}{c|}{SeaS}
    & \multicolumn{4}{c}{MAGIC (ours)} \\
    & AUROC & AP & $F_1$-max & PRO
    & AUROC & AP & $F_1$-max & PRO
    & AUROC & AP & $F_1$-max & PRO
    & AUROC & AP & $F_1$-max & PRO
    & AUROC & AP & $F_1$-max & PRO \\
    \midrule
    bagel
    & 98.4 & 3.4 & 8.8 & 94.7
    & 98.5 & 4.7 & 11.7 & 95.6
    & \underline{99.3} & 7.8 & 15.2 & 95.5
    & \underline{99.3} & 5.3 & 12.2 & 97.1
    & \textbf{99.7} & \textbf{11.0} & \textbf{19.4} & \textbf{98.2} \\
    
    cable\_gland
    & 95.7 & 1.5 & 4.4 & 86.2
    & \textbf{99.3} & \textbf{9.5} & \textbf{18.9} & \textbf{97}
    & 95.3 & 1.6 & 5.7 & 83.6
    & 93.4 & 4.3 & \underline{12.2} & 89.4
    & \underline{98.7} & \underline{5.2} & 12.0 & \underline{95.4} \\
    
    carrot
    & 98.9 & 7.2 & 17.1 & 95.2
    & 99.4 & 6.3 & 14 & \underline{97.8}
    & 99.2 & 14.2 & 23.7 & 97.1
    & \underline{99.8} & \underline{24.1} & \underline{31.6} & \textbf{98.1}
    & \textbf{99.9} & \textbf{29.9} & \textbf{34.7} & \underline{97.8} \\
    
    cookie
    & 88.4 & 2.3 & 6.2 & 87.8
    & 91.2 & 6.8 & 17.4 & 85.1
    & \underline{94.9} & \textbf{18.9} & \textbf{33.5} & 93.3
    & 94.7 & 8.3 & 19.2 & 92.2
    & \textbf{97.5} & \underline{13.2} & \underline{23.4} & \textbf{95.9} \\
    
    dowel
    & 99.3 & 33.8 & 38.4 & 87.2
    & 99 & 25.7 & 36.8 & \underline{89.4}
    & 98.4 & 30.5 & 39.7 & 89.1
    & \textbf{99.8} & \textbf{57.3} & \textbf{57.8} & 86.9
    & \underline{99.7} & \underline{54.0} & \underline{51.7} & \textbf{97.9} \\
    
    foam
    & \underline{99.9} & 74.6 & 68.7 & 90.2
    & \underline{99.9} & 65.3 & 61.1 & 82.2
    & 96.4 & 58 & 63 & 73
    & 99.8 & 70.0 & 66.5 & 83.4
    & \textbf{100.0} & \textbf{78.5} & \textbf{71.0} & \textbf{96.5} \\
    
    peach
    & 97.2 & 3.4 & 9.3 & 91.8
    & \underline{99.3} & 15.5 & 24.3 & 96.1
    & 99.2 & \underline{21.3} & 29.6 & 96.1
    & \underline{99.3} & \textbf{37.0} & \textbf{40.3} & \underline{96.8}
    & \textbf{99.8} & 20.0 & \underline{31.1} & \textbf{97.3} \\
    
    potato
    & 97.5 & 1.9 & 5.1 & 90.7
    & \underline{99.5} & 14.8 & \underline{25.4} & 90.4
    & 96.9 & 7.5 & 14 & 90.8
    & 96.4 & 10.1 & 14.1 & \underline{95.1}
    & \textbf{99.6} & \textbf{17.7} & \textbf{25.9} & \textbf{97.1} \\
    
    rope
    & \textbf{99.2} & 7.3 & 12.9 & \textbf{97.1}
    & 98.5 & \textbf{11} & 11.5 & 96.5
    & \underline{98.8} & 5.2 & 11.7 & \underline{96.6}
    & 94.7 & \underline{10.3} & \textbf{17.0} & 93.5
    & 98.3 & 8.6 & 10.0 & 95.2 \\
    
    tire
    & \underline{98.6} & 16.2 & 27.6 & 92.3
    & 98.4 & \textbf{40} & \textbf{43.5} & 83.2
    & 95 & 2.9 & 10.3 & 81.7
    & \textbf{99.5} & \underline{31.9} & \underline{35.8} & \textbf{93.1}
    & \underline{98.6} & 22.0 & 28.0 & \underline{92.5} \\
    
    \midrule
    Average
    & 97.31 & 15.16 & 19.85 & 91.32
    & \underline{98.3} & 19.96 & 26.46 & 91.33
    & 97.34 & 16.79 & 24.64 & 89.68
    & 97.67 & \underline{25.86} & \underline{30.67} & \underline{92.56}
    & \textbf{99.18} & \textbf{26.01} & \textbf{30.72} & \textbf{96.38} \\
    \bottomrule
    \end{tabular}
    }
    \vspace{-2mm}
    \caption{Category-wise quantitative comparison of pixel-level anomaly localization on the MVTec 3D-AD dataset.}
    \label{tab:pixel_detection_3d}
\end{table*}

\clearpage

\begin{figure*}[t]
  \centering
  \includegraphics[width=1.0\textwidth]{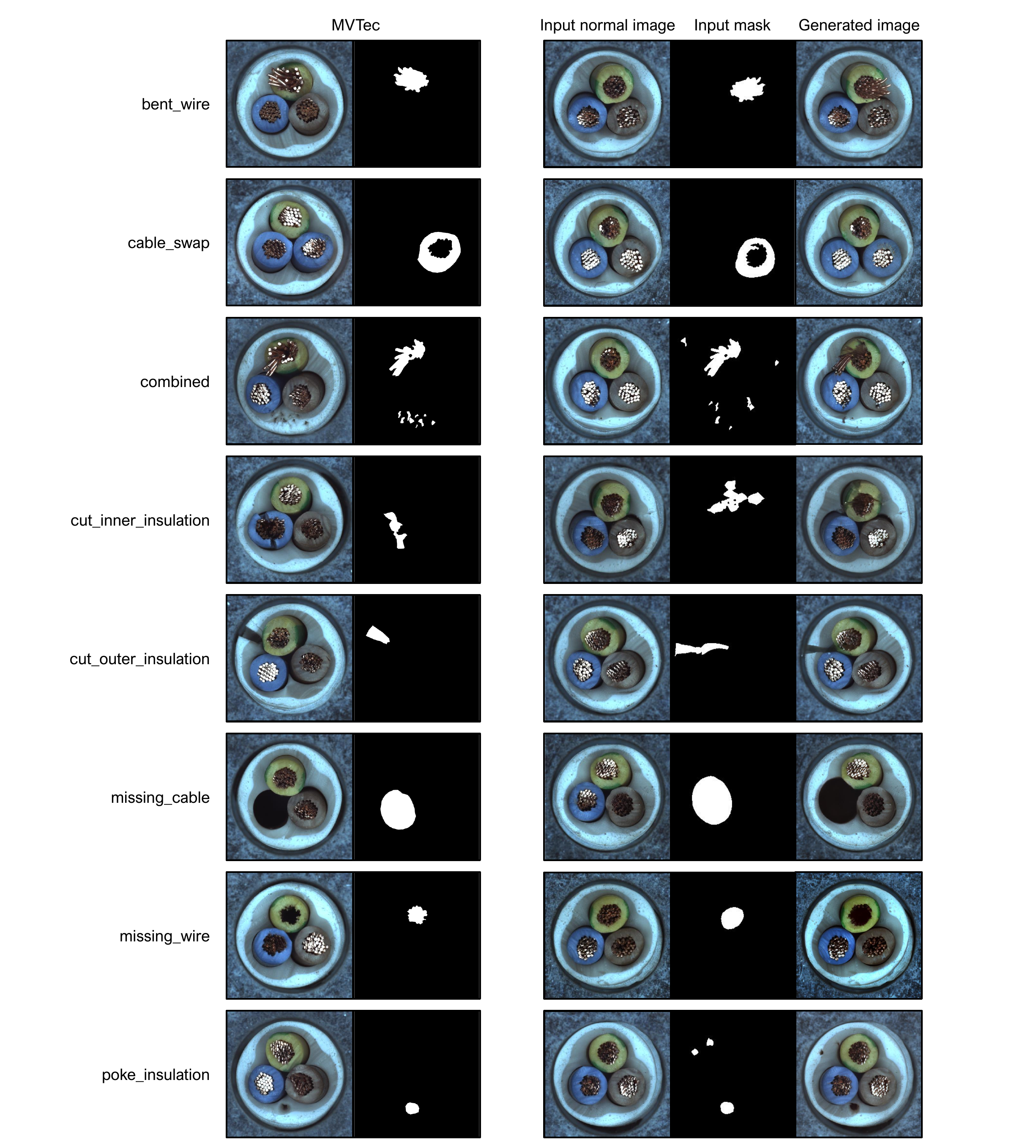}
  \vspace{-3mm}
    \caption{Generated images on \textit{cable}
    }
  \vspace{-5mm}
\end{figure*}

\begin{figure*}[t]
  \centering
  \includegraphics[width=1.0\textwidth]{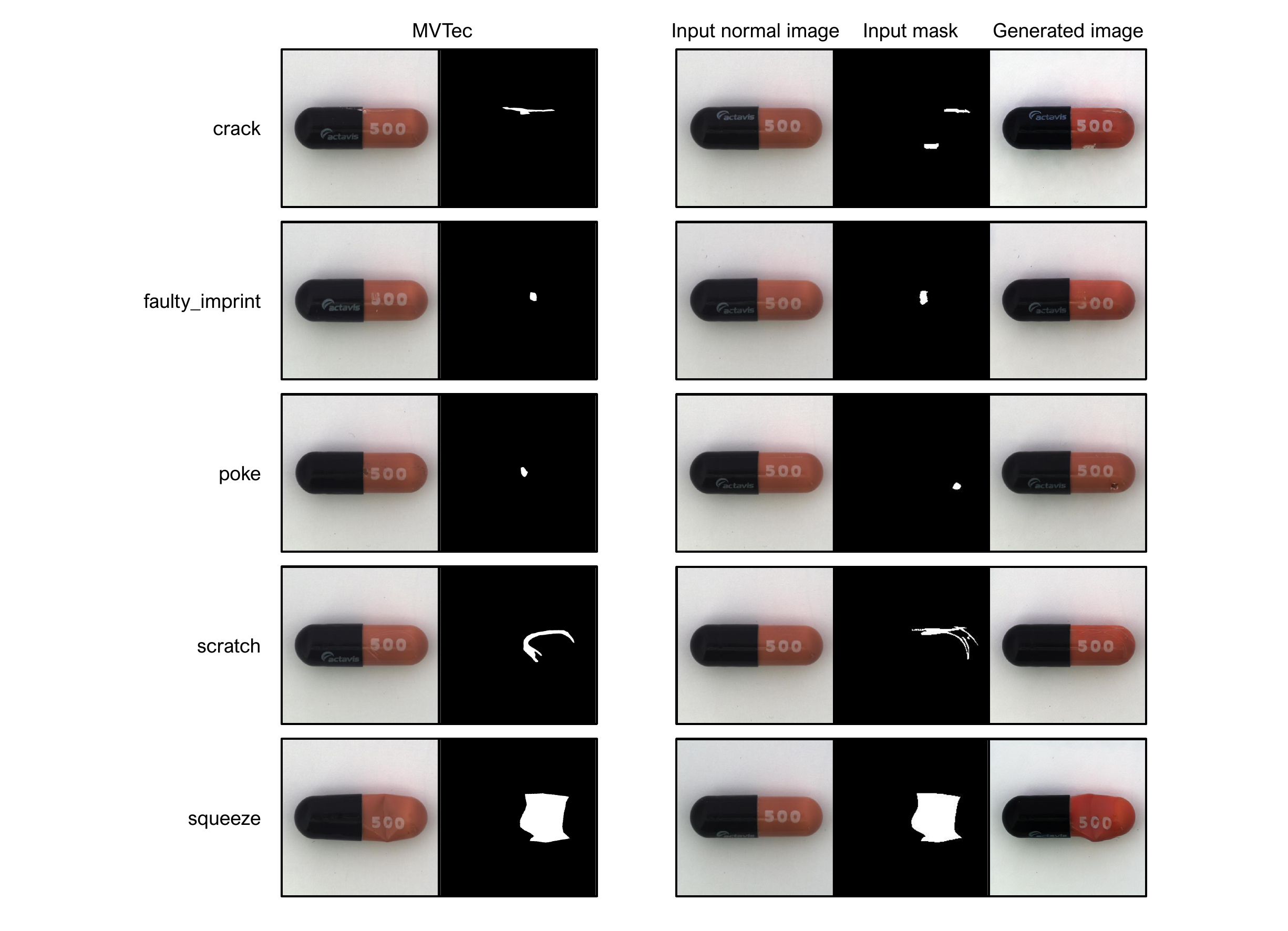}
  \vspace{-3mm}
    \caption{Generated images on \textit{capsule}
    }
  \vspace{-5mm}
\end{figure*}

\begin{figure*}[t]
  \centering
  \includegraphics[width=1.0\textwidth]{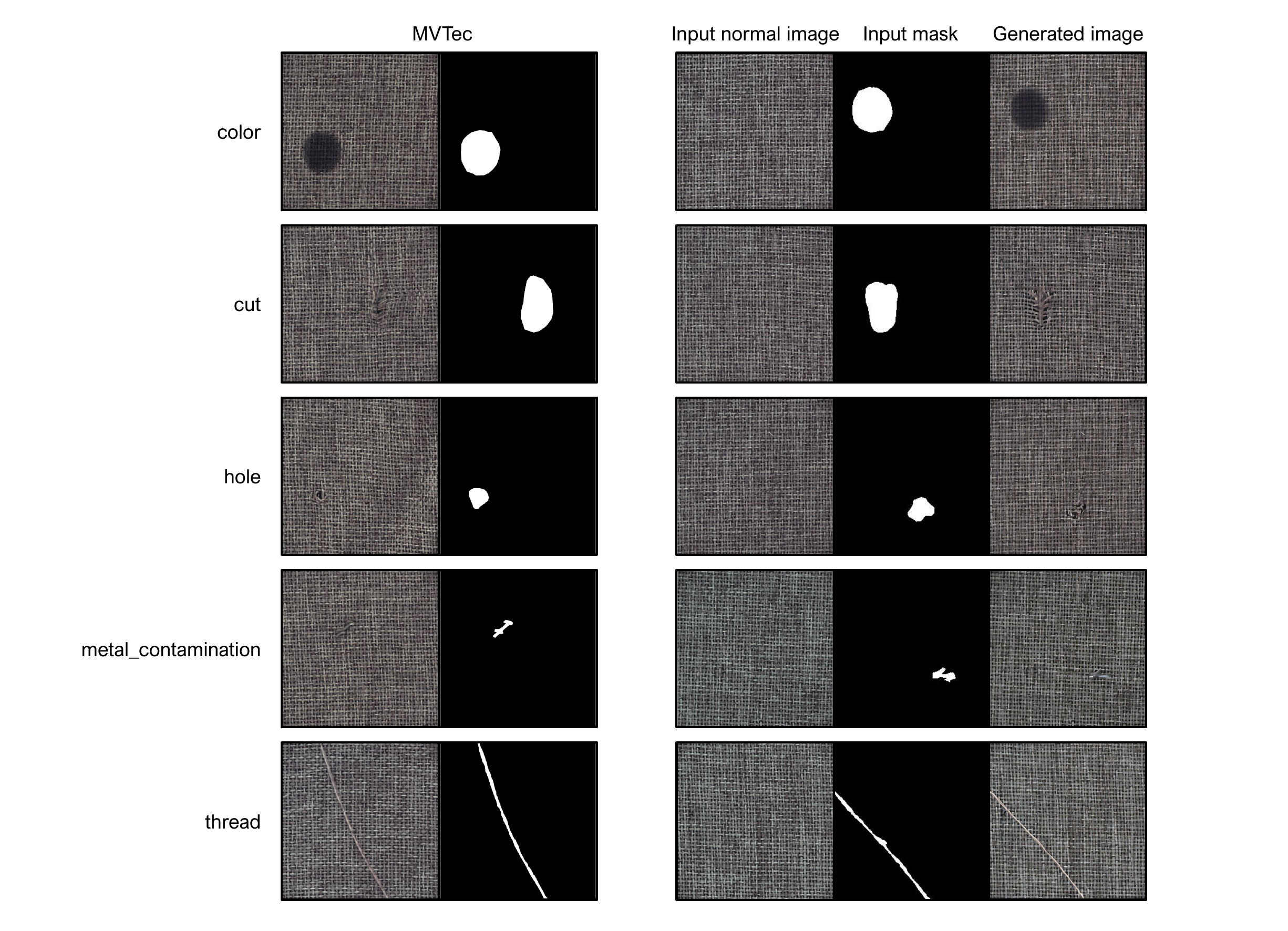}
  \vspace{-3mm}
    \caption{Generated images on \textit{carpet}
    }
  \vspace{-5mm}
\end{figure*}

\begin{figure*}[t]
  \centering
  \includegraphics[width=1.0\textwidth]{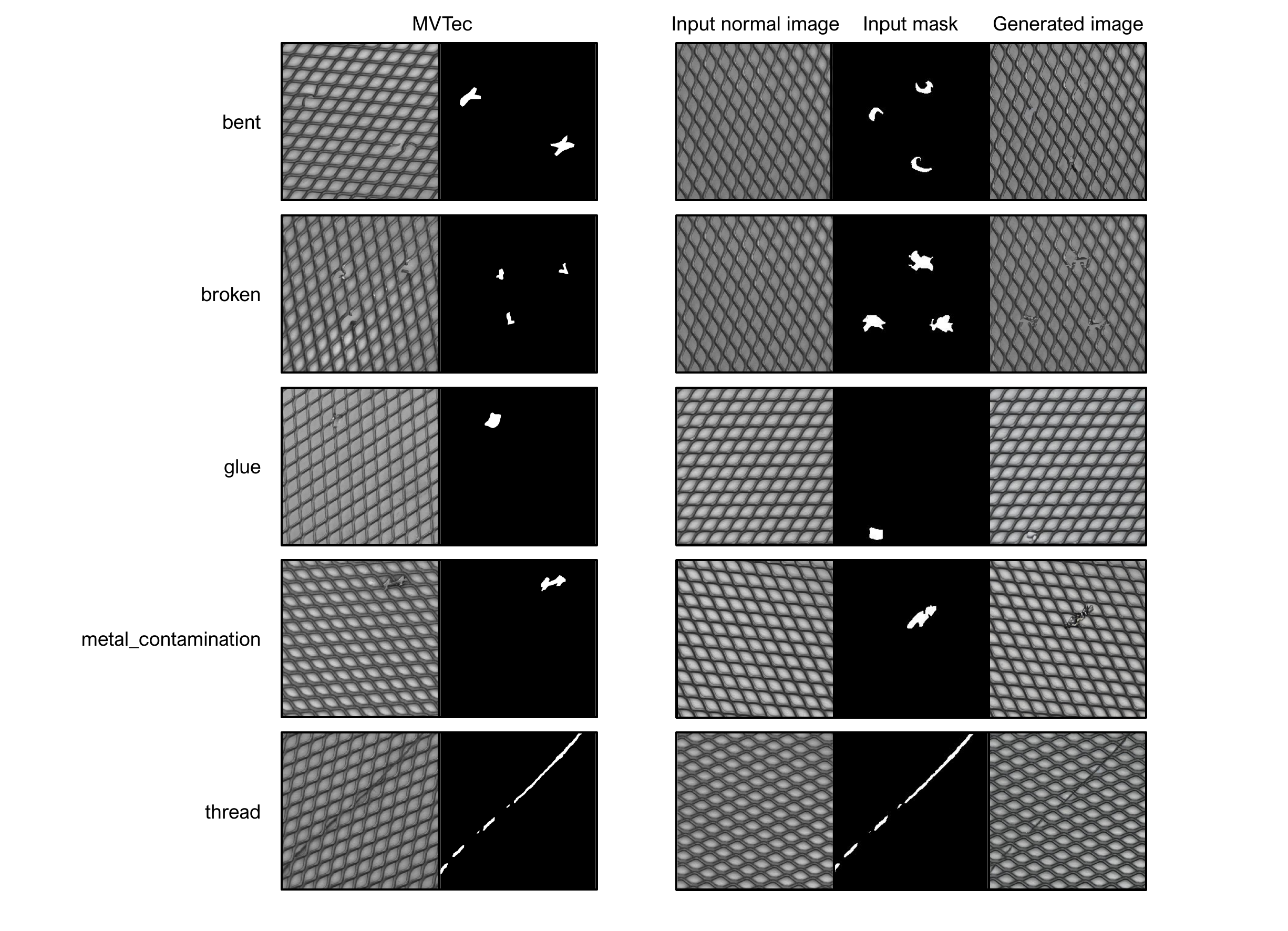}
  \vspace{-3mm}
    \caption{Generated images on \textit{grid}
    }
  \vspace{-5mm}
\end{figure*}

\begin{figure*}[t]
  \centering
  \includegraphics[width=1.0\textwidth]{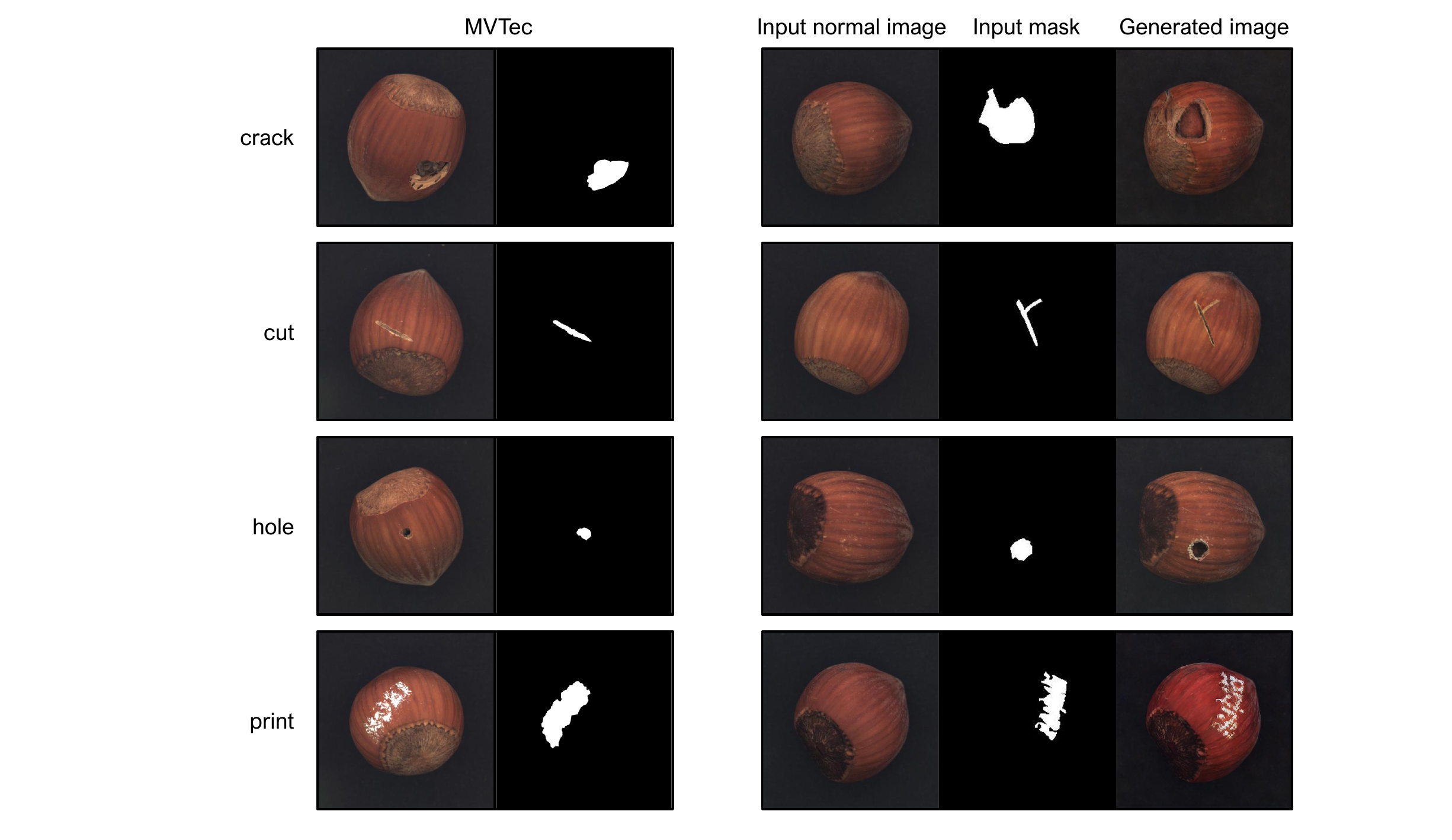}
  \vspace{-3mm}
    \caption{Generated images on \textit{hazelnut}
    }
  \vspace{-5mm}
\end{figure*}

\begin{figure*}[t]
  \centering
  \includegraphics[width=1.0\textwidth]{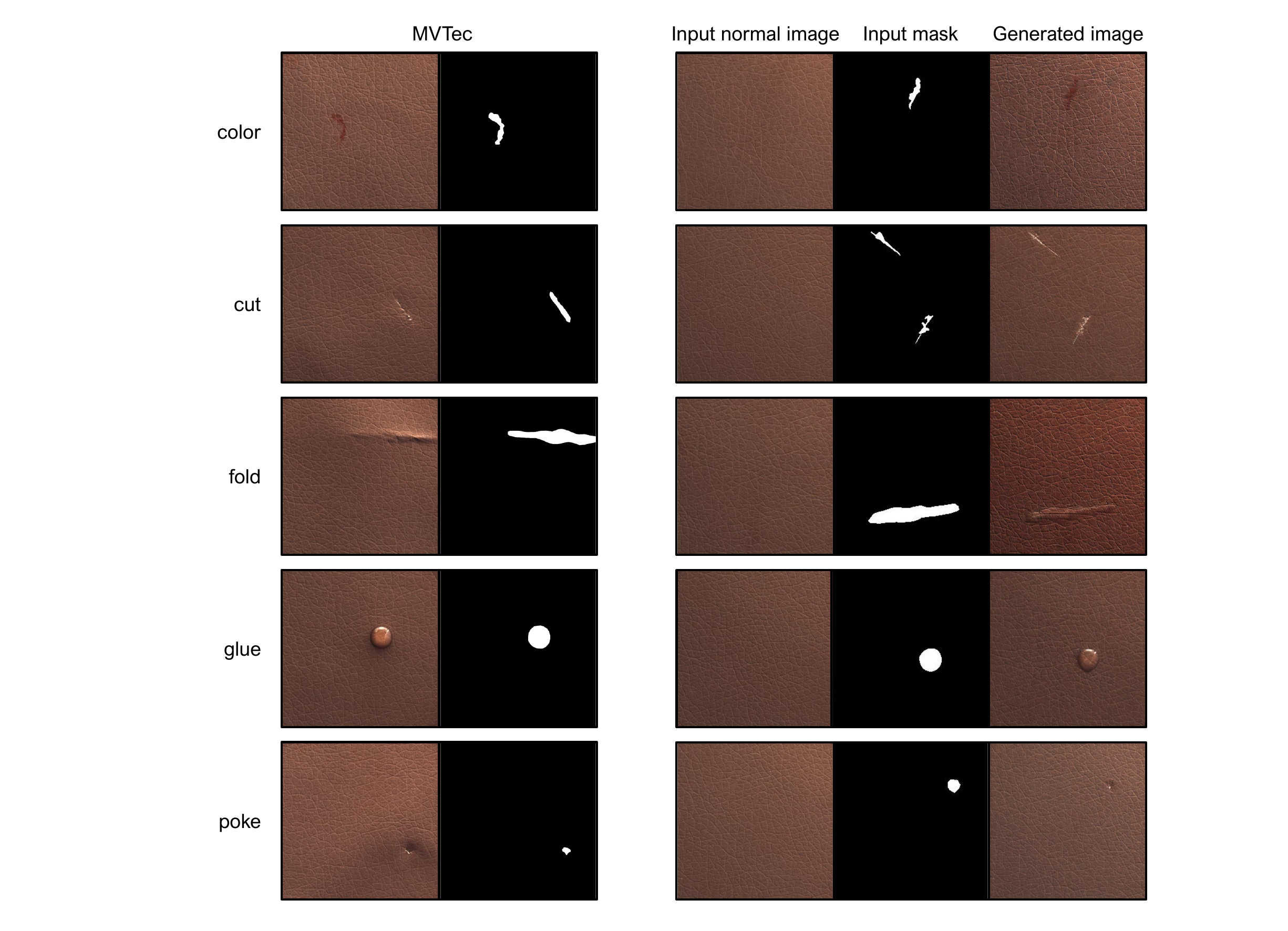}
  \vspace{-3mm}
    \caption{Generated images on \textit{leather}
    }
  \vspace{-5mm}
\end{figure*}

\begin{figure*}[t]
  \centering
  \includegraphics[width=1.0\textwidth]{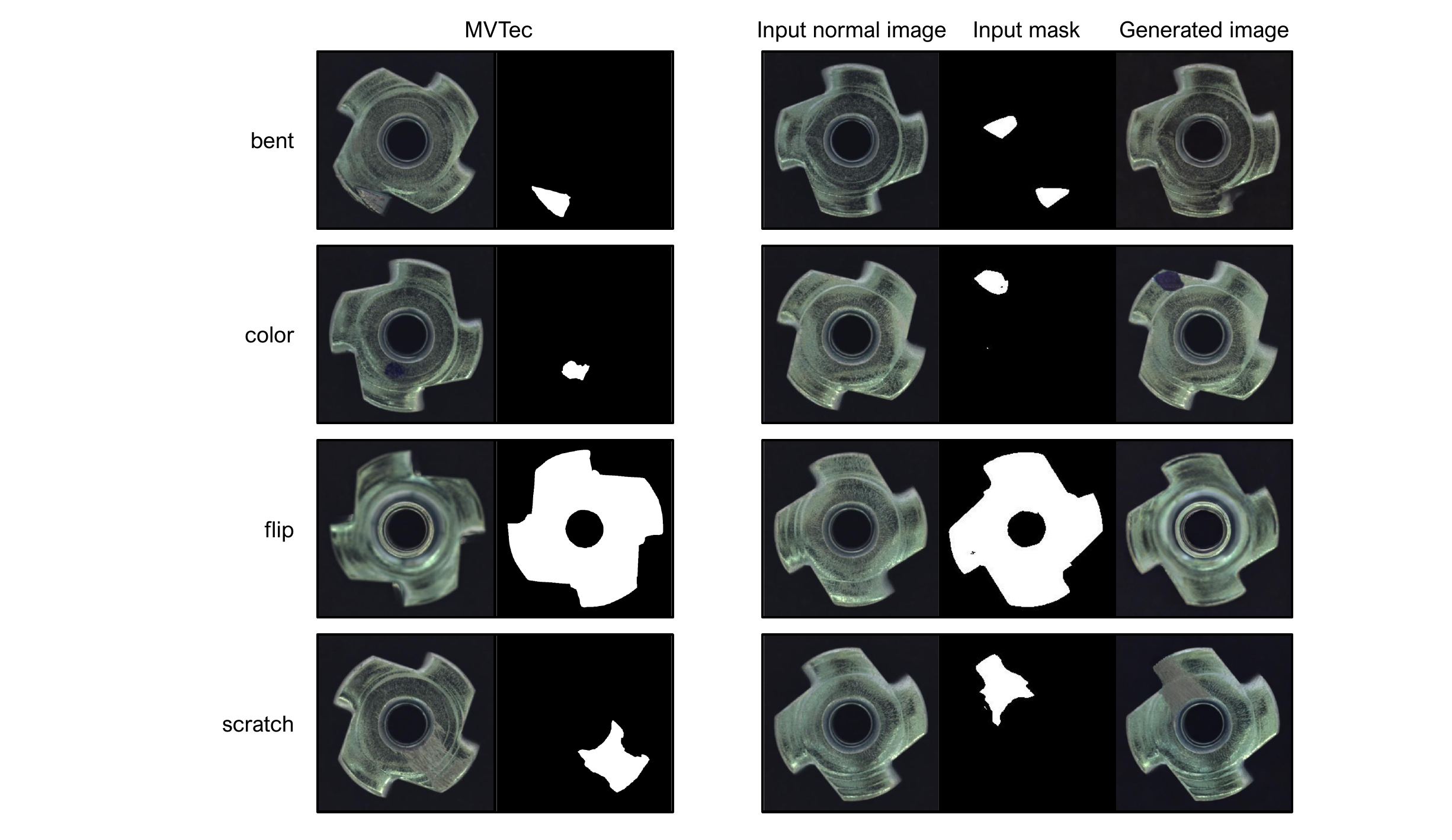}
  \vspace{-3mm}
    \caption{Generated images on \textit{metal\_nut}
    }
  \vspace{-5mm}
\end{figure*}

\begin{figure*}[t]
  \centering
  \includegraphics[width=1.0\textwidth]{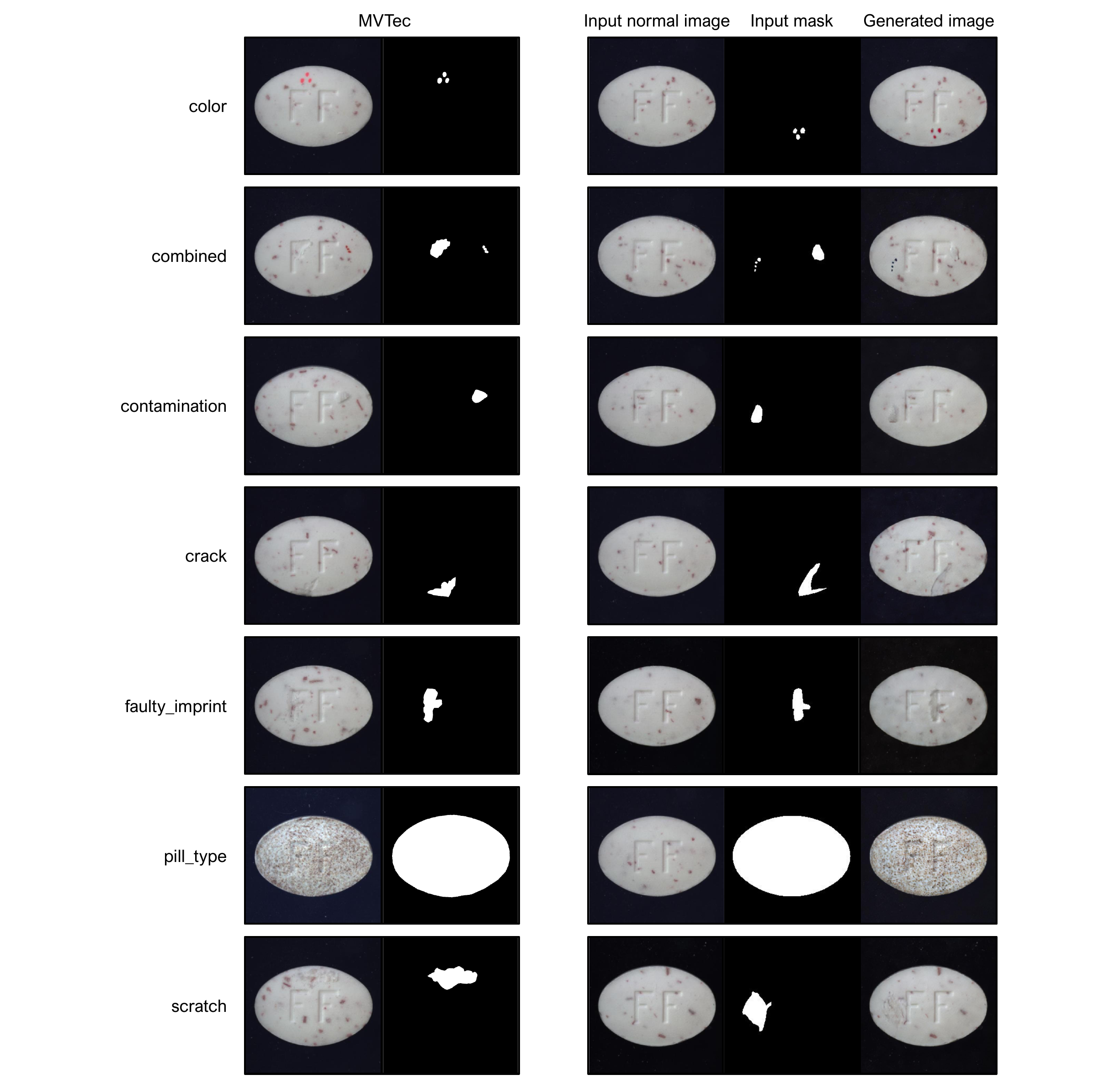}
  \vspace{-3mm}
    \caption{Generated images on \textit{pill}
    }
  \vspace{-5mm}
\end{figure*}

\begin{figure*}[t]
  \centering
  \includegraphics[width=1.0\textwidth]{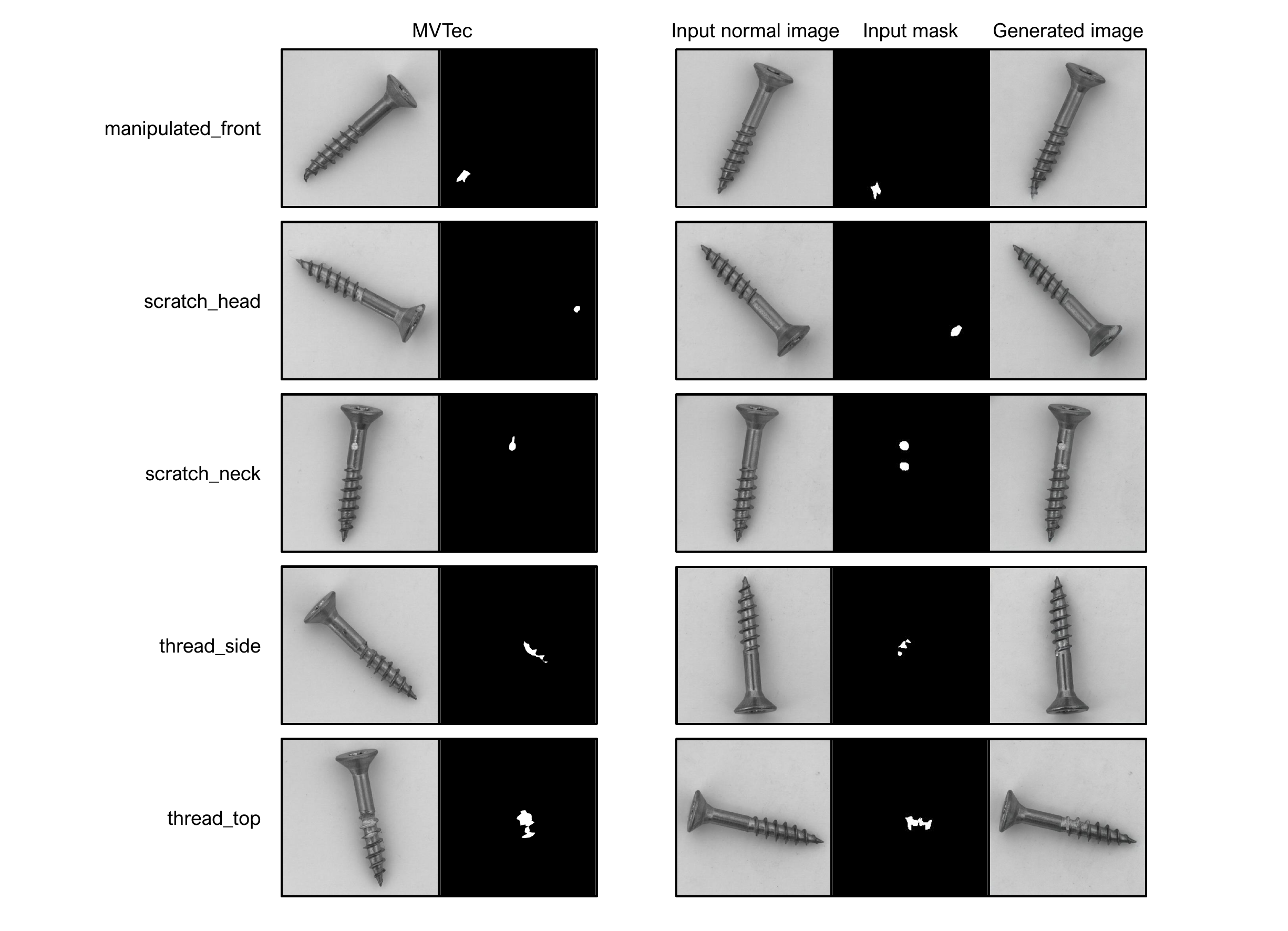}
  \vspace{-3mm}
    \caption{Generated images on \textit{screw}
    }
  \vspace{-5mm}
\end{figure*}

\begin{figure*}[t]
  \centering
  \includegraphics[width=1.0\textwidth]{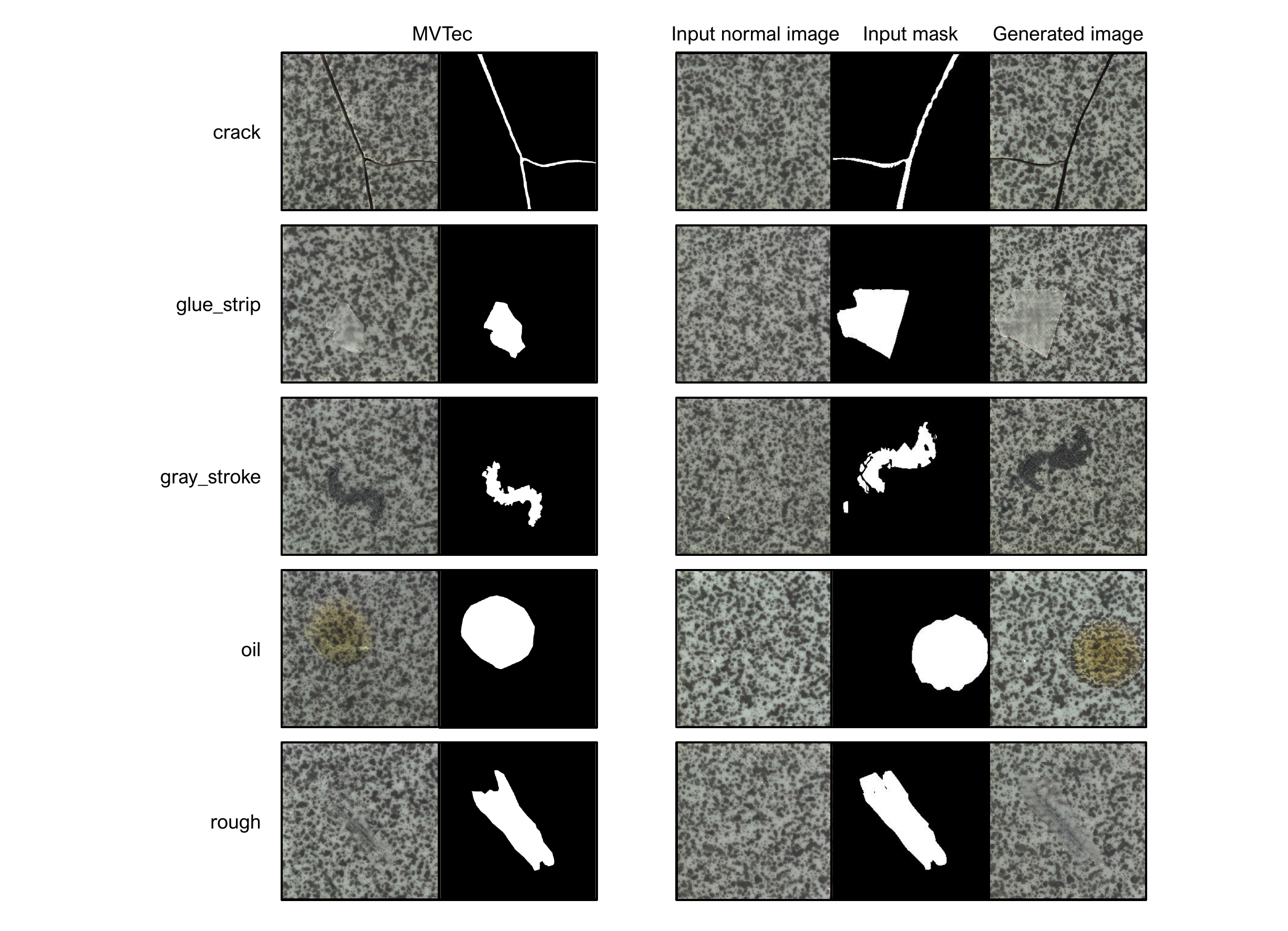}
  \vspace{-3mm}
    \caption{Generated images on \textit{tile}
    }
  \vspace{-5mm}
\end{figure*}

\begin{figure*}[t]
  \centering
  \includegraphics[width=1.0\textwidth]{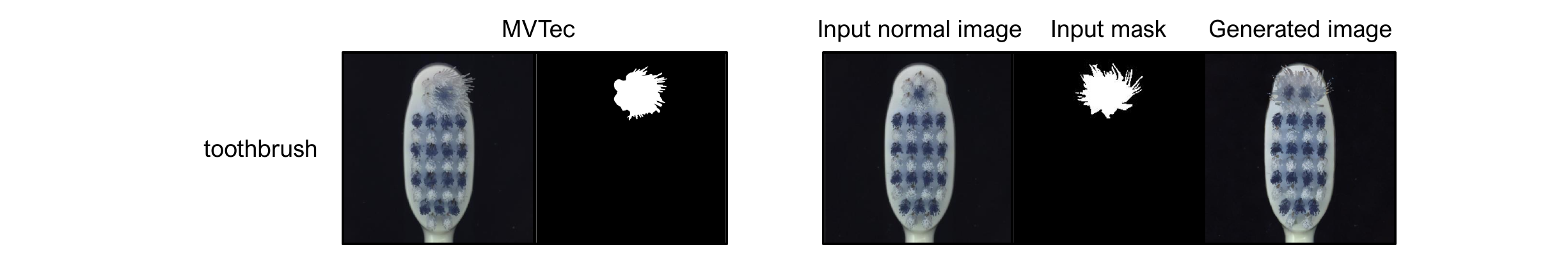}
  \vspace{-3mm}
    \caption{Generated images on \textit{toothbrush}
    }
  \vspace{-5mm}
\end{figure*}

\begin{figure*}[t]
  \centering
  \includegraphics[width=1.0\textwidth]{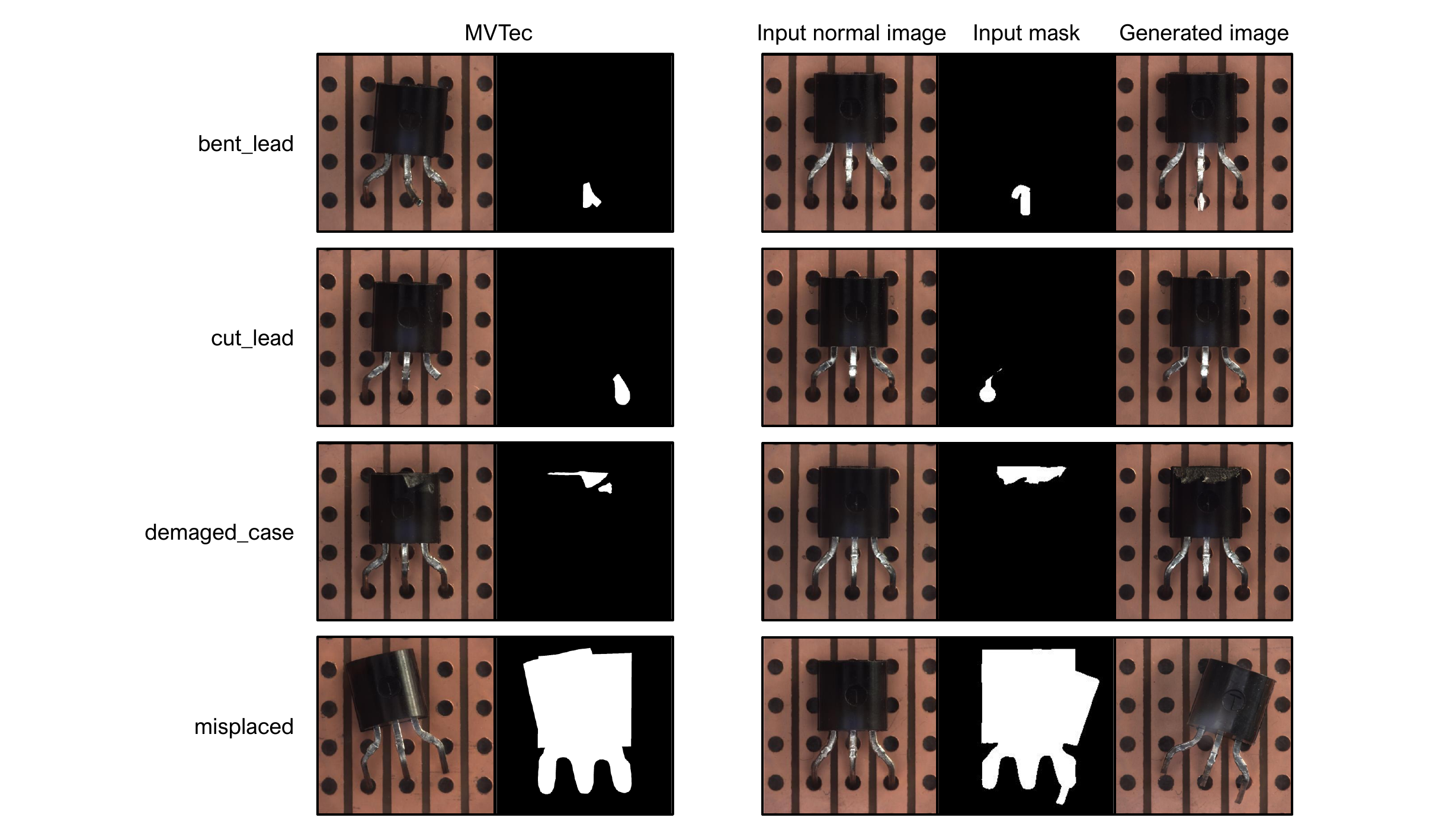}
  \vspace{-3mm}
    \caption{Generated images on \textit{transistor}
    }
  \vspace{-5mm}
\end{figure*}

\begin{figure*}[t]
  \centering
  \includegraphics[width=1.0\textwidth]{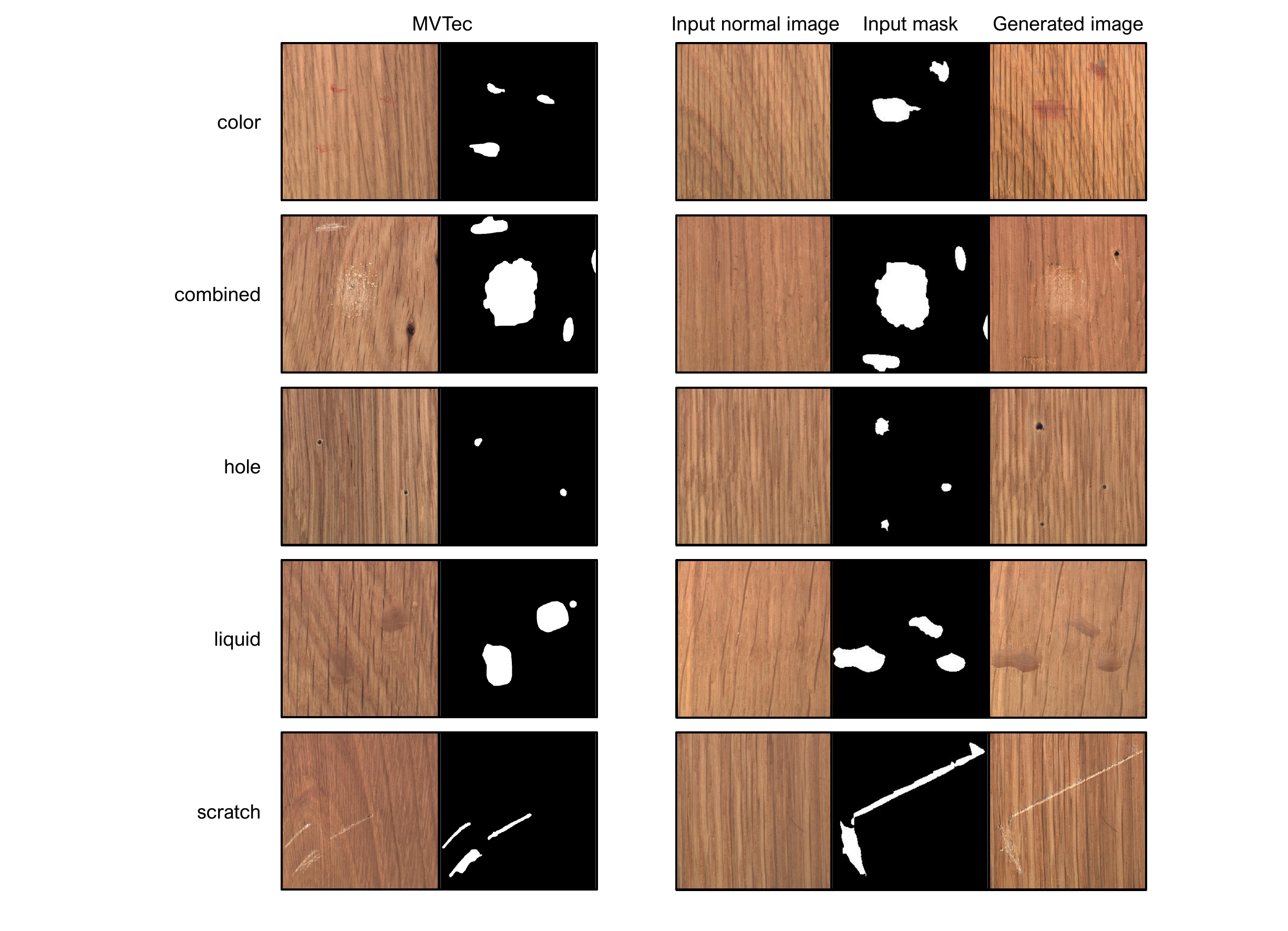}
  \vspace{-3mm}
    \caption{Generated images on \textit{wood}
    }
  \vspace{-5mm}
\end{figure*}

\begin{figure*}[t]
  \centering
  \includegraphics[width=1.0\textwidth]{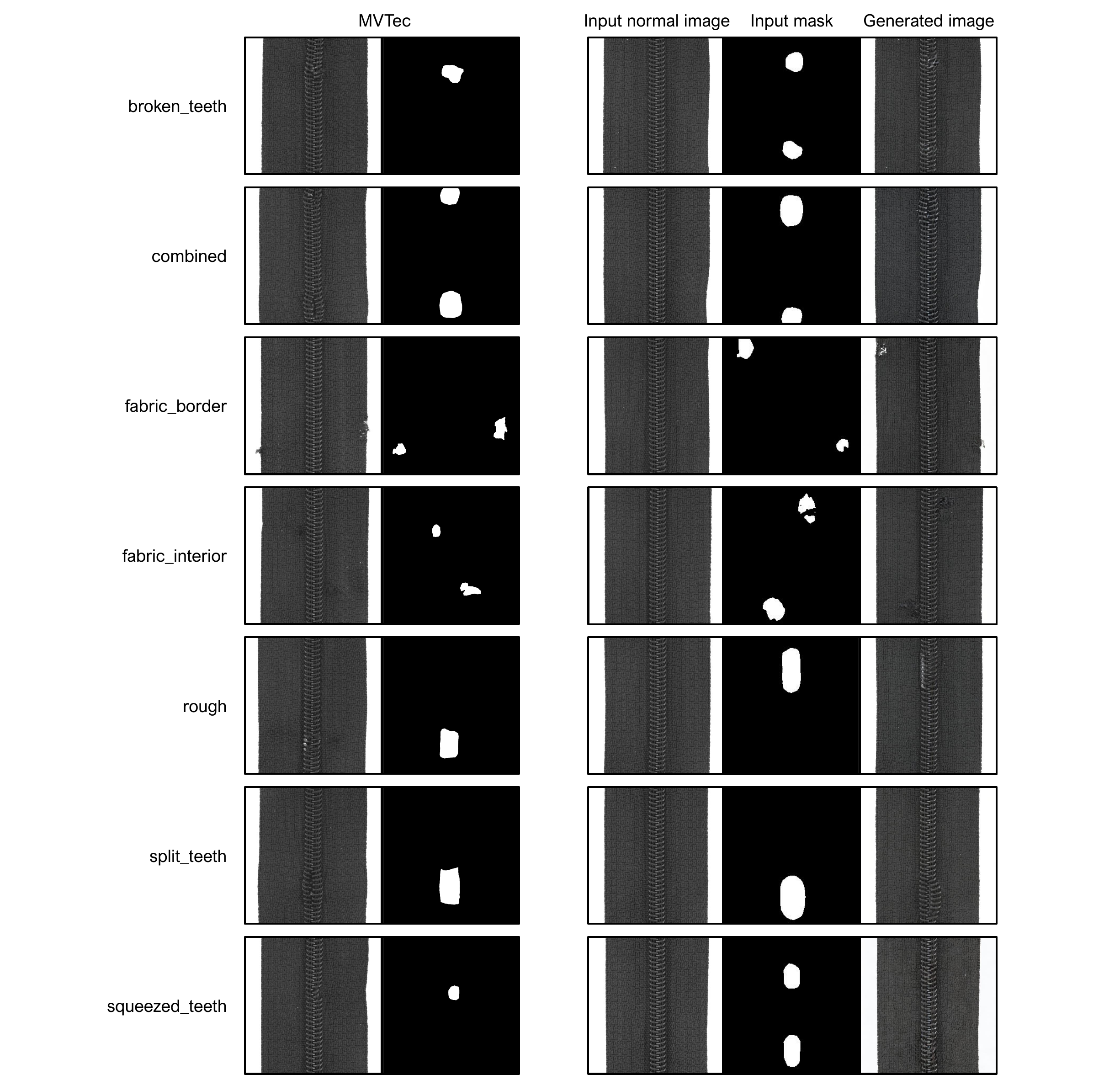}
  \vspace{-3mm}
    \caption{Generated images on \textit{zipper}
    }
  \vspace{-5mm}
\end{figure*}

\begin{figure*}[t]
  \centering
  \includegraphics[width=1.0\textwidth]{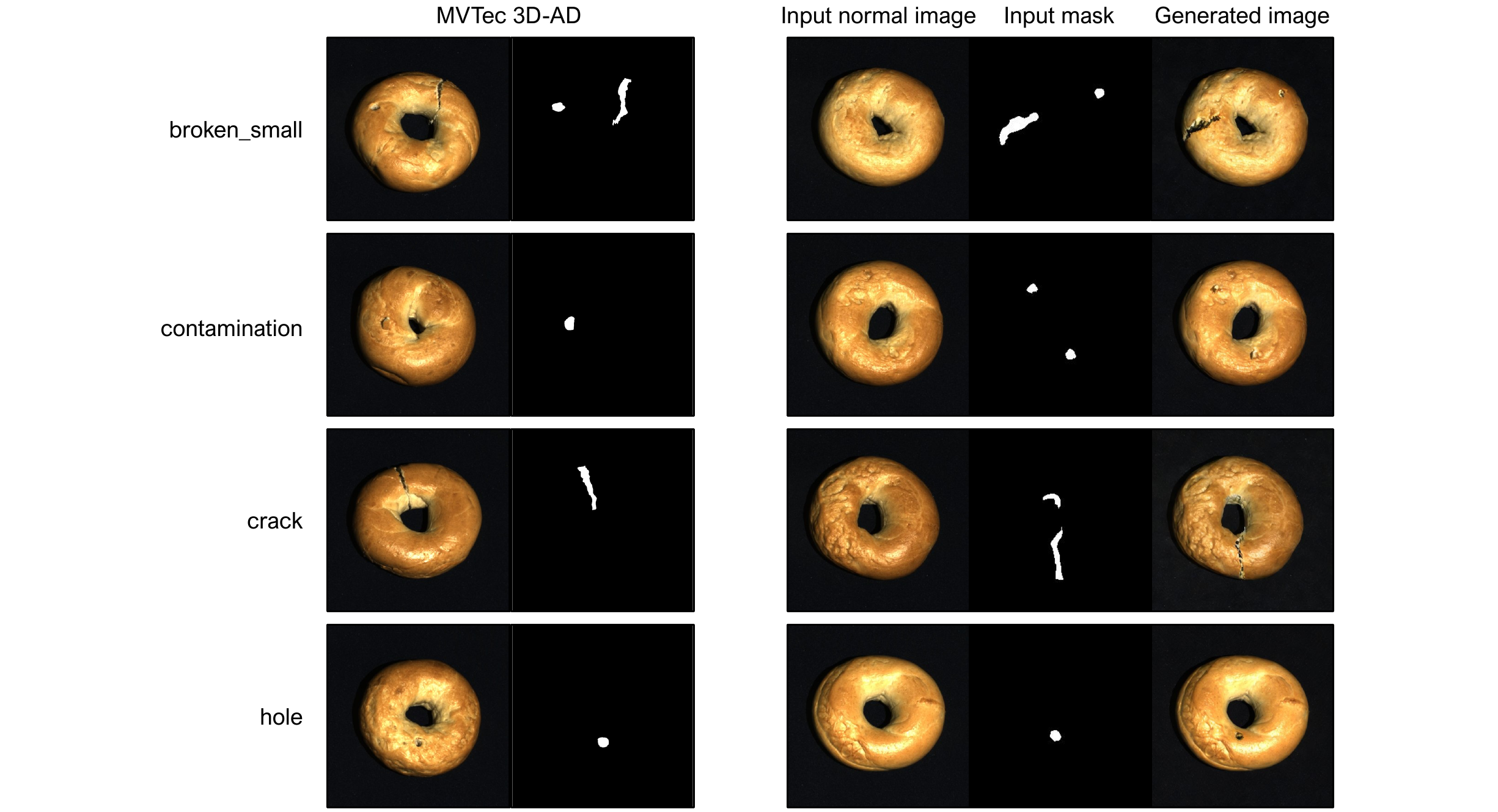}
  \vspace{-3mm}
    \caption{Generated images on \textit{bagel}
    }
  \vspace{-5mm}
\end{figure*}

\begin{figure*}[t]
  \centering
  \includegraphics[width=1.0\textwidth]{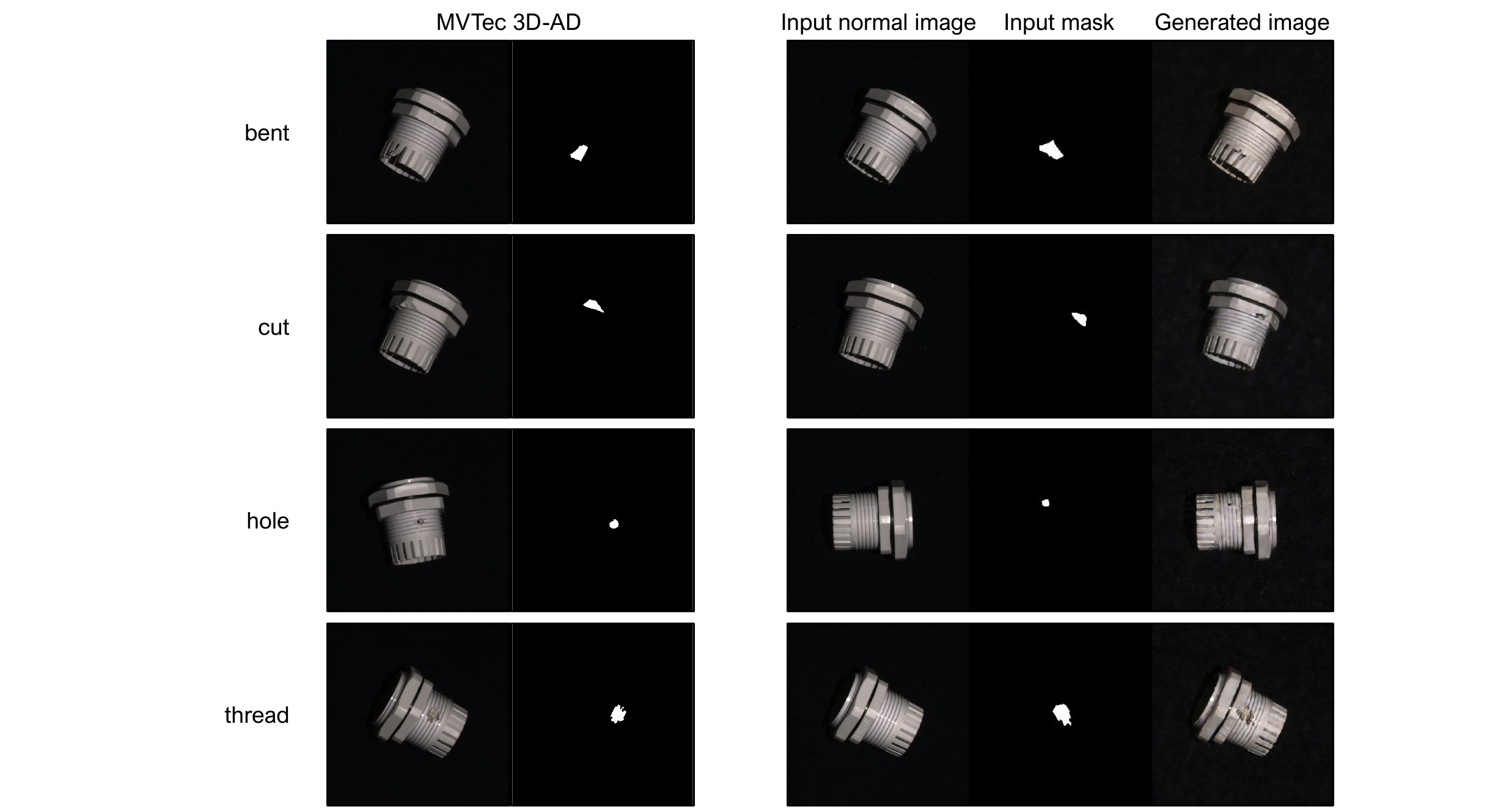}
  \vspace{-4mm}
    \caption{Generated images on \textit{cable\_gland}}
  \vspace{-7mm}
\end{figure*}

\begin{figure*}[t]
  \centering
  \includegraphics[width=1.0\textwidth]{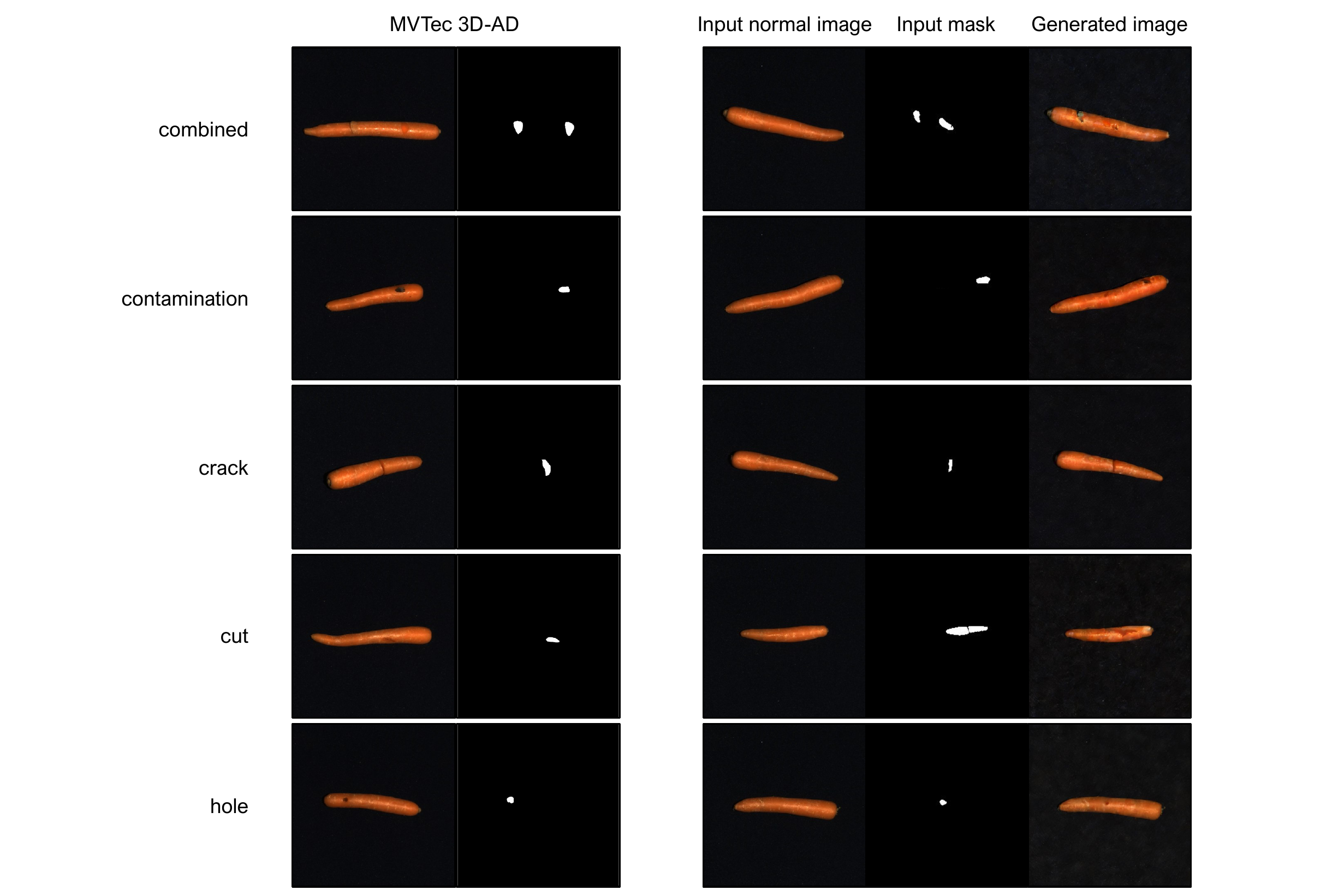}
  \vspace{-4mm}
    \caption{Generated images on \textit{carrot}
    }
  \vspace{-7mm}
\end{figure*}

\begin{figure*}[t]
  \centering
  \includegraphics[width=1.0\textwidth]{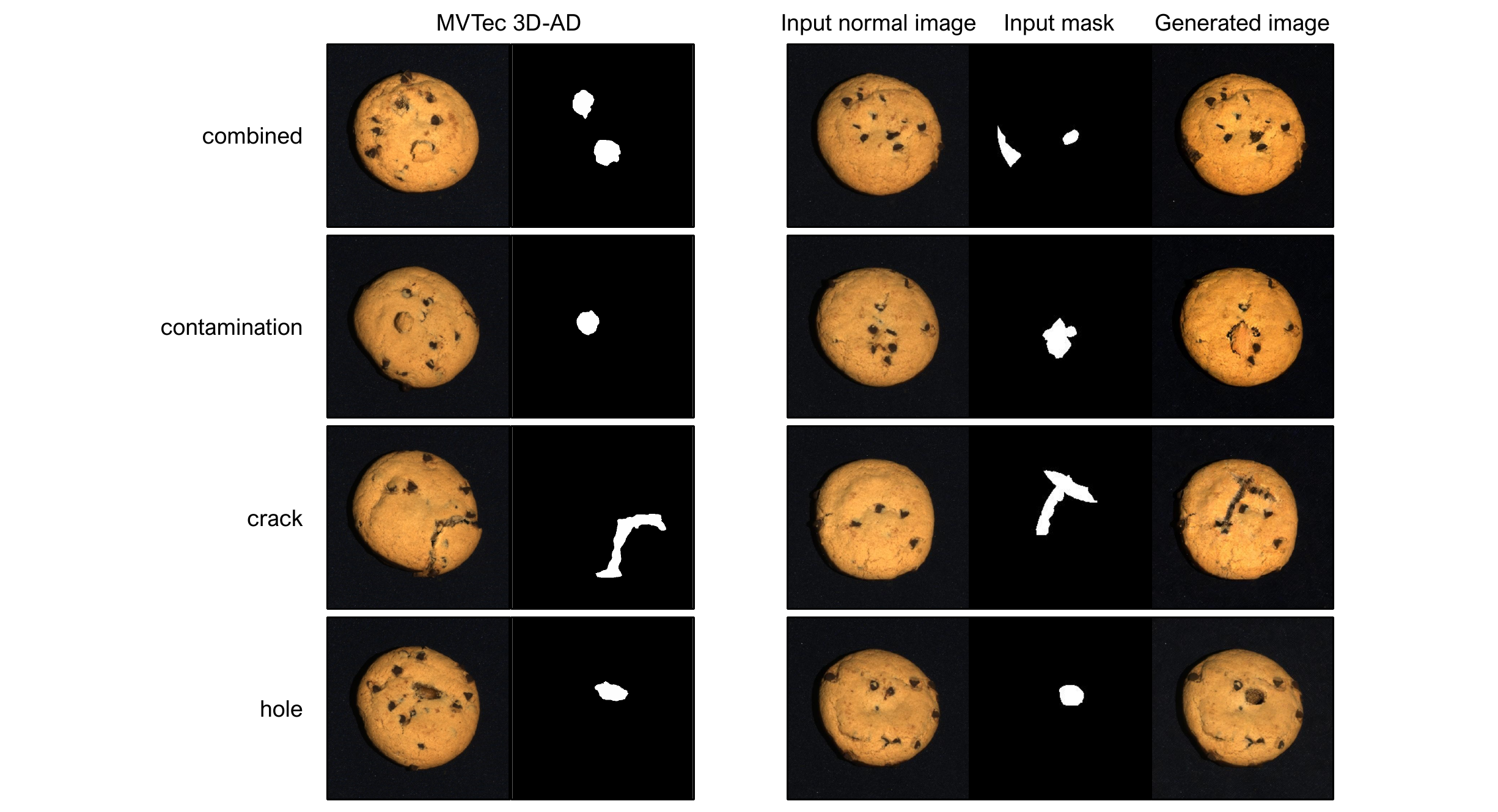}
  \vspace{-4mm}
    \caption{Generated images on \textit{cookie}
    }
  \vspace{-7mm}
\end{figure*}

\begin{figure*}[t]
  \centering
  \includegraphics[width=1.0\textwidth]{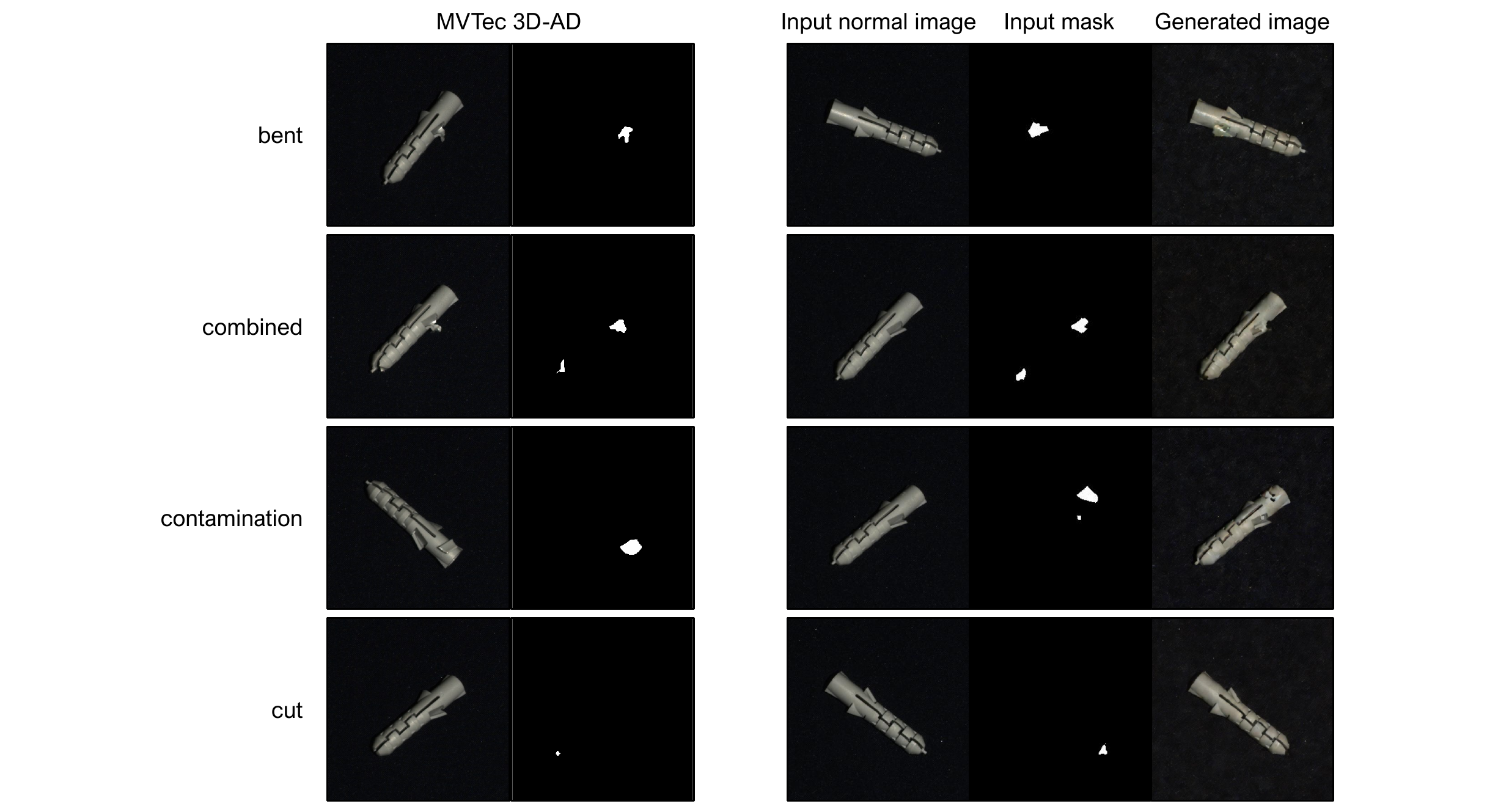}
  \vspace{-4mm}
    \caption{Generated images on \textit{dowel}
    }
  \vspace{-7mm}
\end{figure*}

\begin{figure*}[t]
  \centering
  \includegraphics[width=1.0\textwidth]{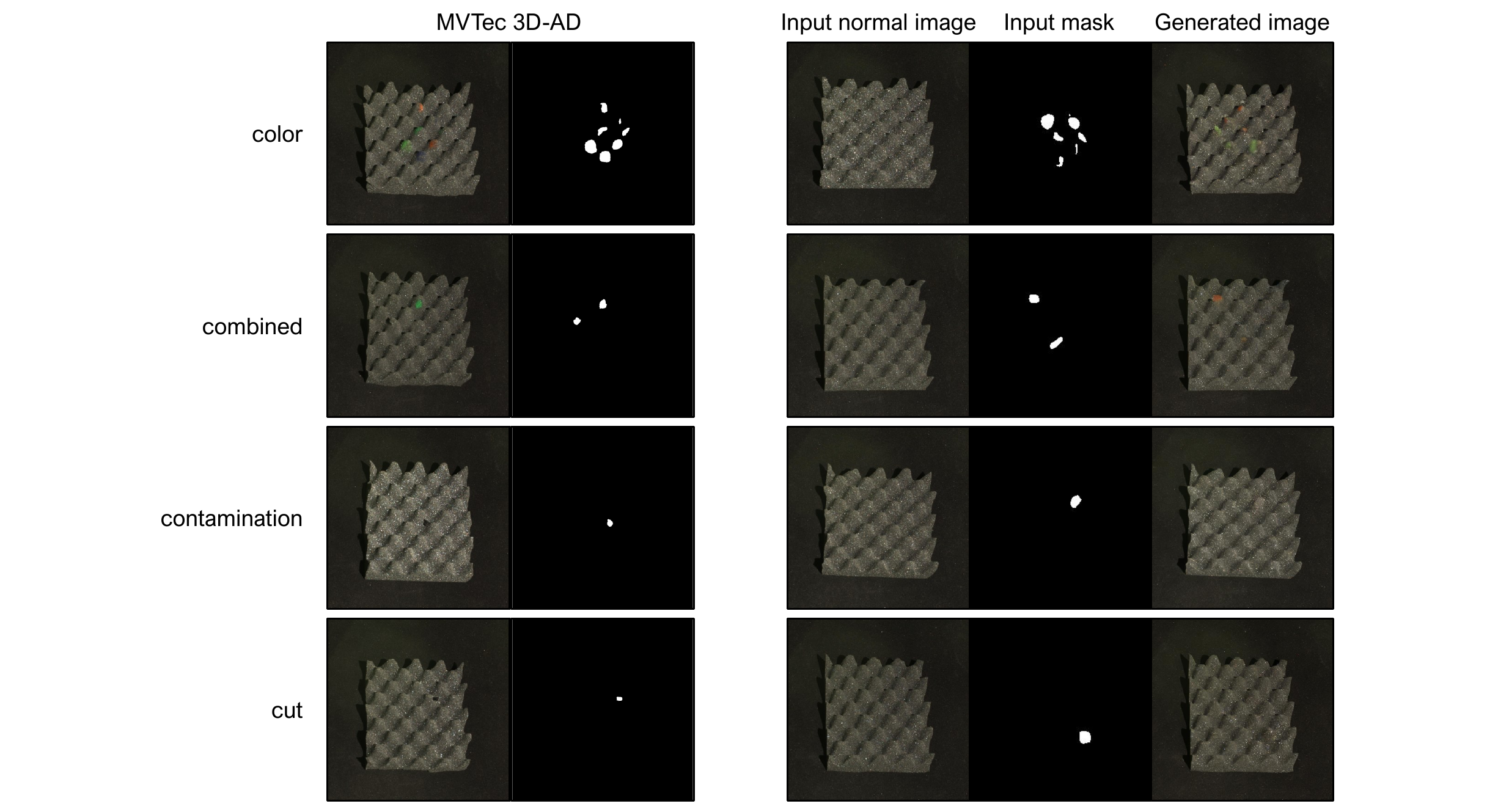}
  \vspace{-4mm}
    \caption{Generated images on \textit{foam}
    }
  \vspace{-7mm}
\end{figure*}

\begin{figure*}[t]
  \centering
  \includegraphics[width=1.0\textwidth]{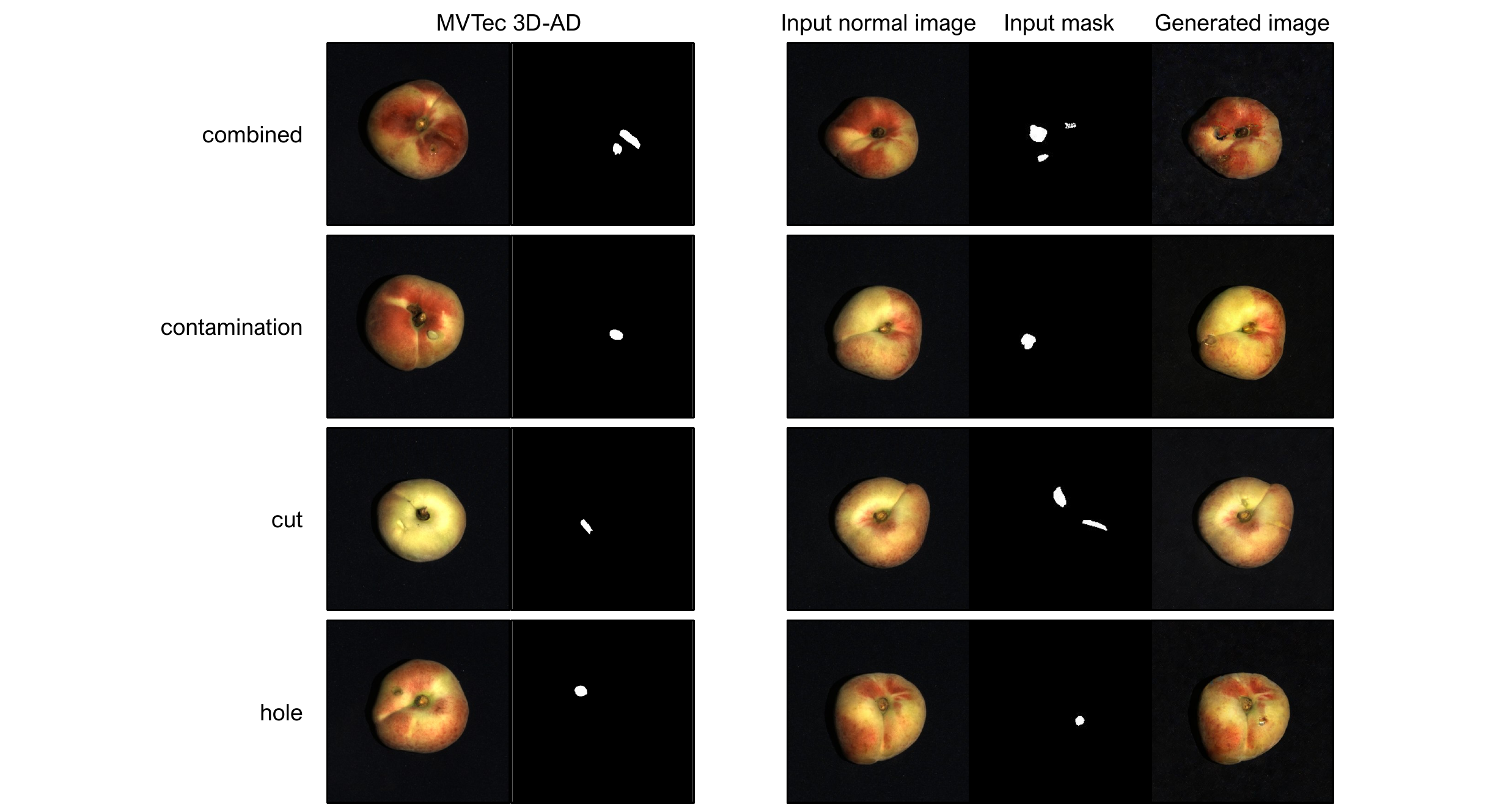}
  \vspace{-4mm}
    \caption{Generated images on \textit{peach}
    }
  \vspace{-7mm}
\end{figure*}

\begin{figure*}[t]
  \centering
  \includegraphics[width=1.0\textwidth]{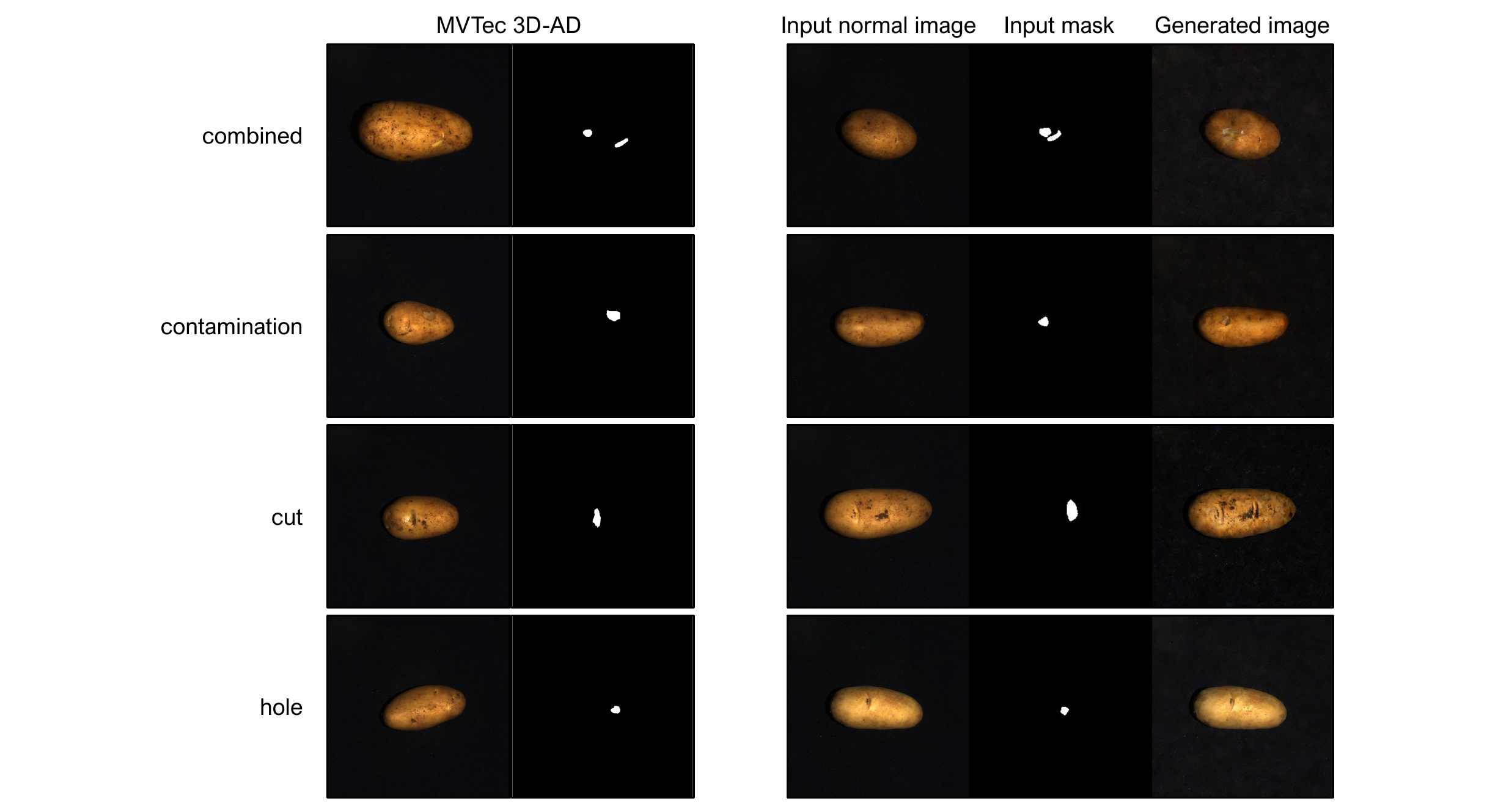}
  \vspace{-4mm}
    \caption{Generated images on \textit{potato}
    }
  \vspace{-7mm}
\end{figure*}

\begin{figure*}[t]
  \centering
  \includegraphics[width=1.0\textwidth]{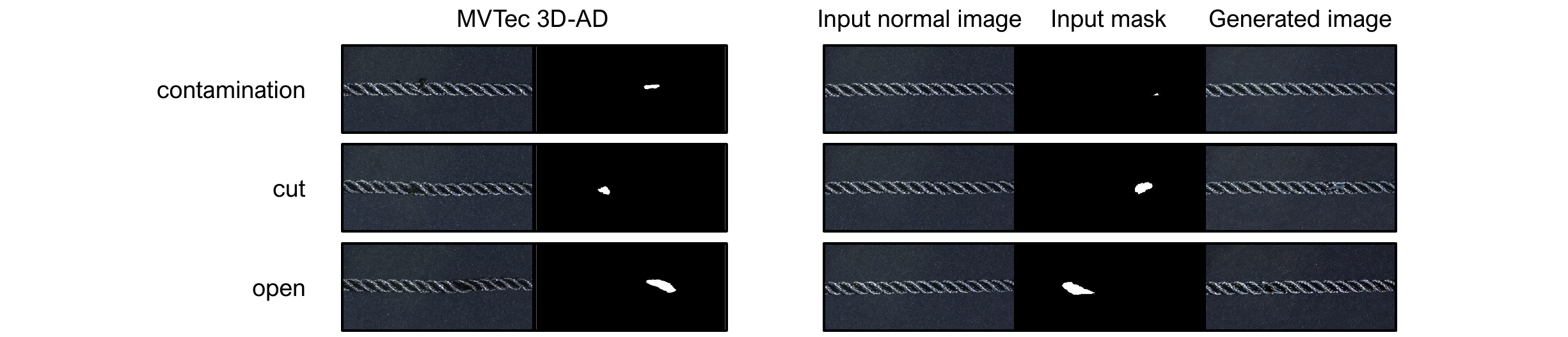}
  \vspace{-4mm}
    \caption{Generated images on \textit{rope}
    }
  \vspace{-7mm}
\end{figure*}

\begin{figure*}[t]
  \centering
  \includegraphics[width=1.0\textwidth]{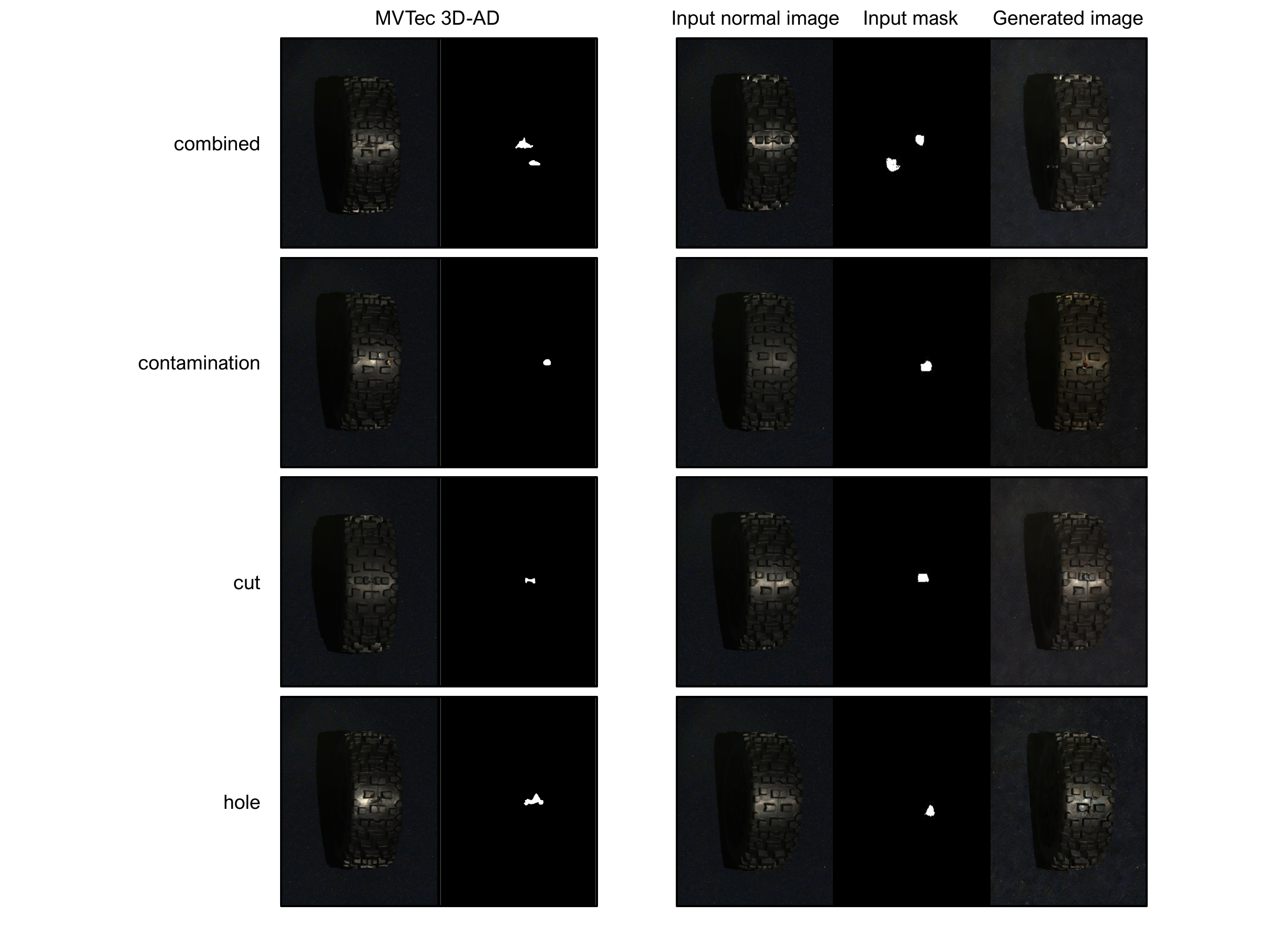}
  \vspace{-4mm}
    \caption{Generated images on \textit{tire}
    }
  \vspace{-7mm}
\end{figure*}


\begin{figure*}[t]
  \centering
  \includegraphics[width=1.0\textwidth]{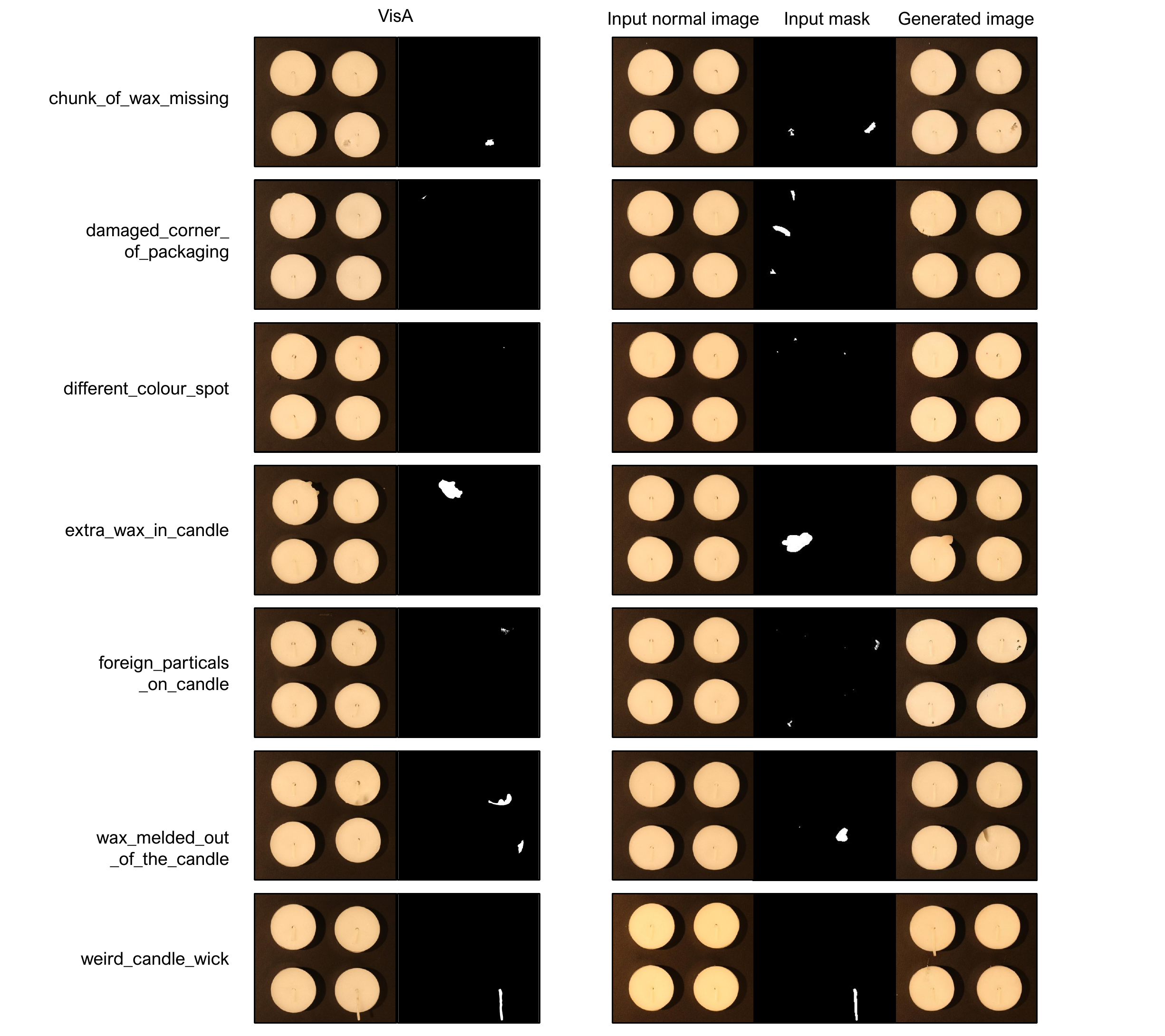}
  \vspace{-4mm}
    \caption{Generated images on \textit{candle}
    }
  \vspace{-7mm}
\end{figure*}

\begin{figure*}[t]
  \centering
  \includegraphics[width=1.0\textwidth]{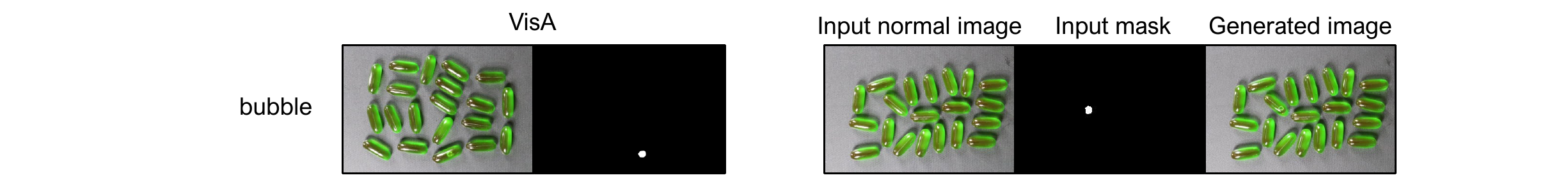}
  \vspace{-4mm}
    \caption{Generated images on \textit{capsule}
    }
  \vspace{-7mm}
\end{figure*}

\begin{figure*}[t]
  \centering
  \includegraphics[width=1.0\textwidth]{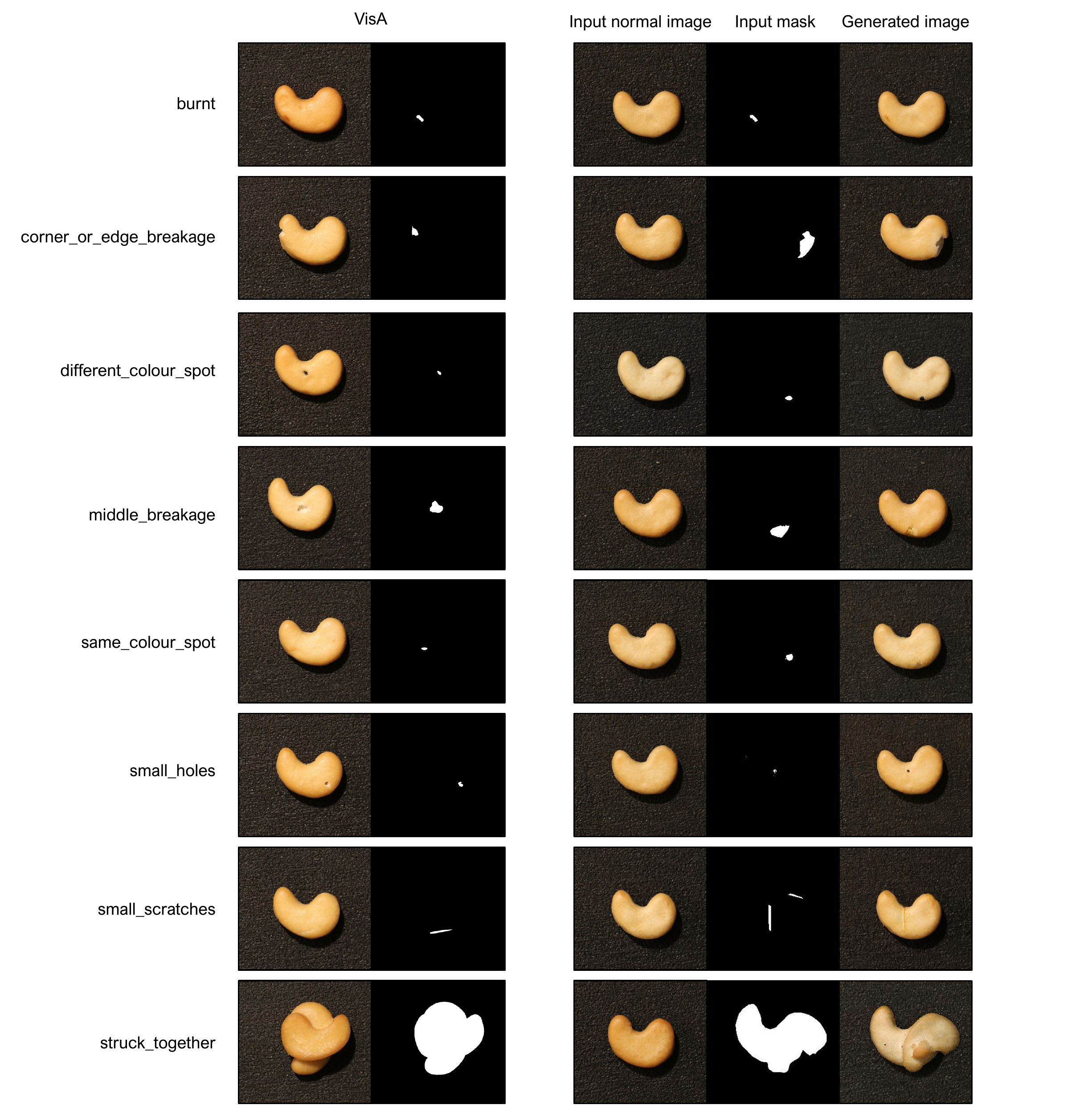}
  \vspace{-4mm}
    \caption{Generated images on \textit{cashew}
    }
  \vspace{-7mm}
\end{figure*}

\begin{figure*}[t]
  \centering
  \includegraphics[width=1.0\textwidth]{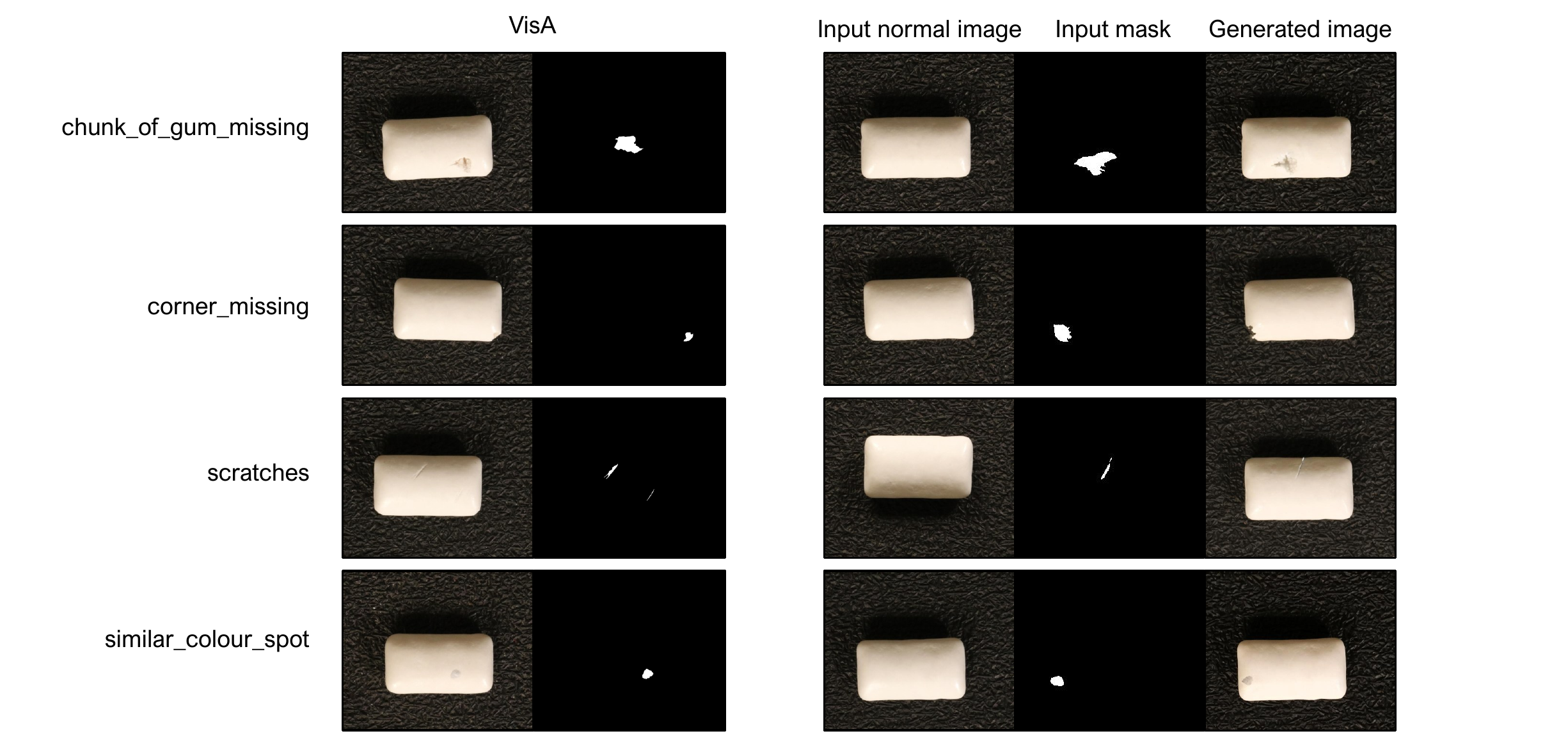}
  \vspace{-4mm}
    \caption{Generated images on \textit{chewinggum}
    }
  \vspace{-7mm}
\end{figure*}

\begin{figure*}[t]
  \centering
  \includegraphics[width=1.0\textwidth]{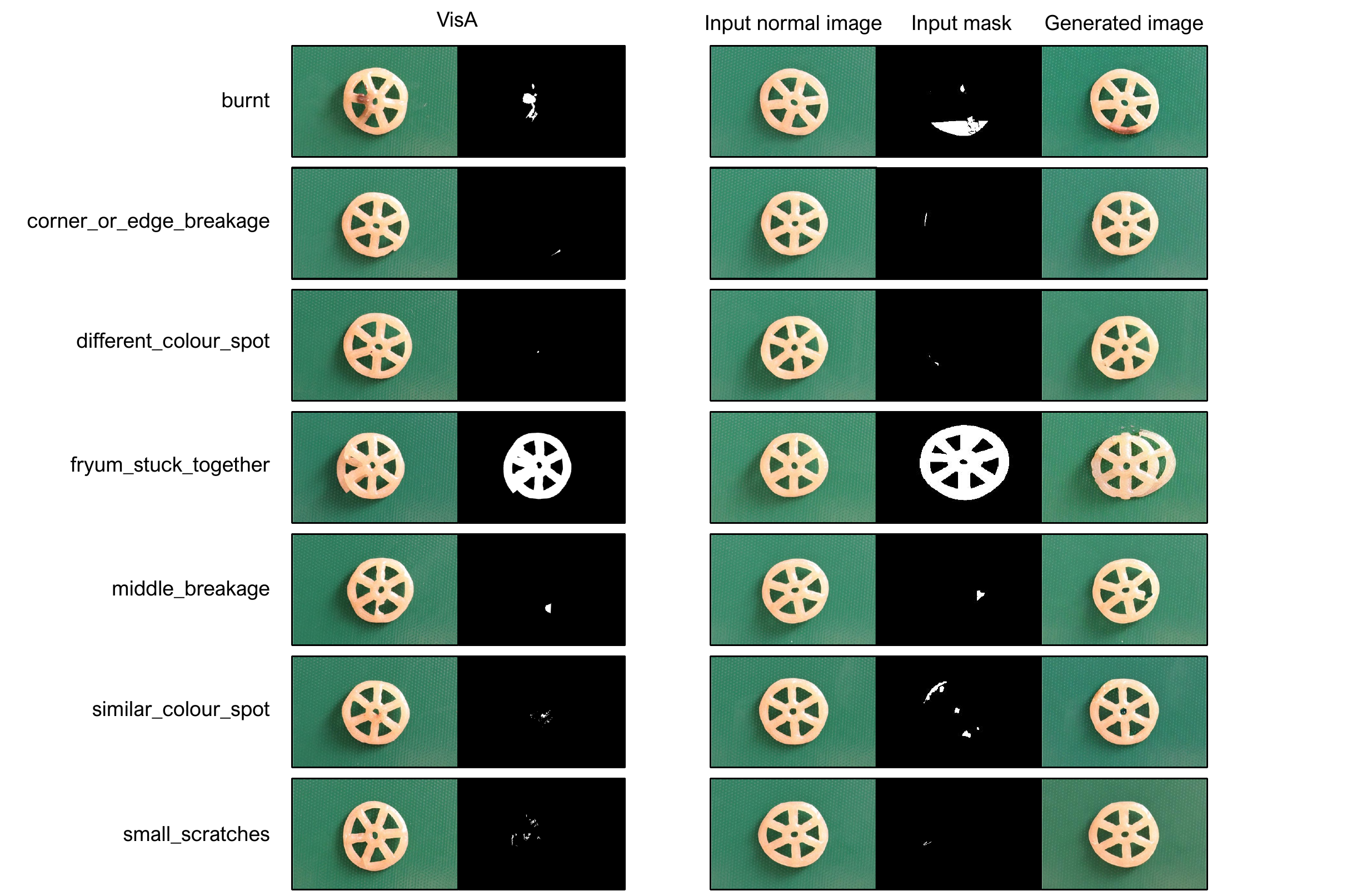}
  \vspace{-4mm}
    \caption{Generated images on \textit{fryum}
    }
  \vspace{-7mm}
\end{figure*}

\begin{figure*}[t]
  \centering
  \includegraphics[width=1.0\textwidth]{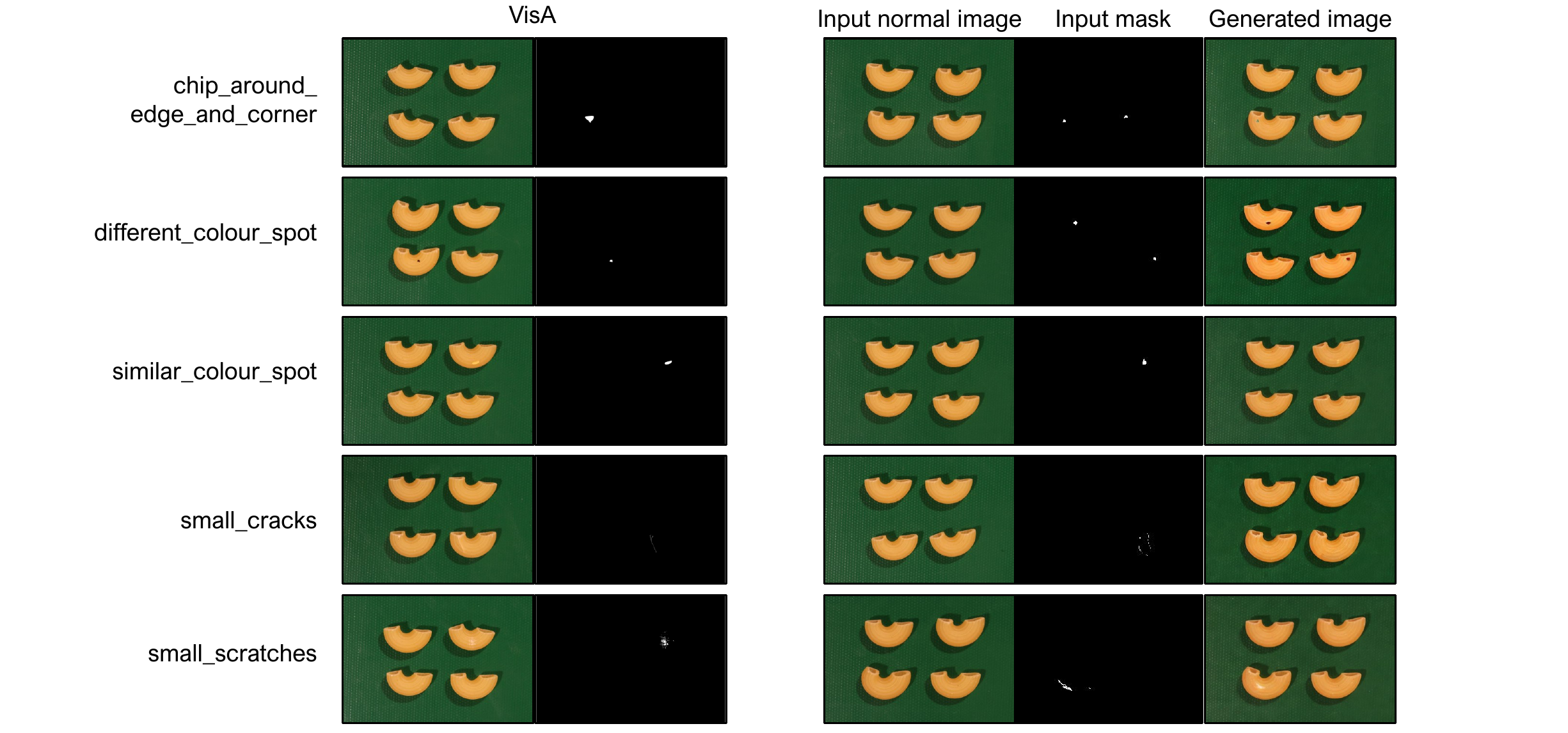}
  \vspace{-4mm}
    \caption{Generated images on \textit{macaroni1}
    }
  \vspace{-7mm}
\end{figure*}

\begin{figure*}[t]
  \centering
  \includegraphics[width=1.0\textwidth]{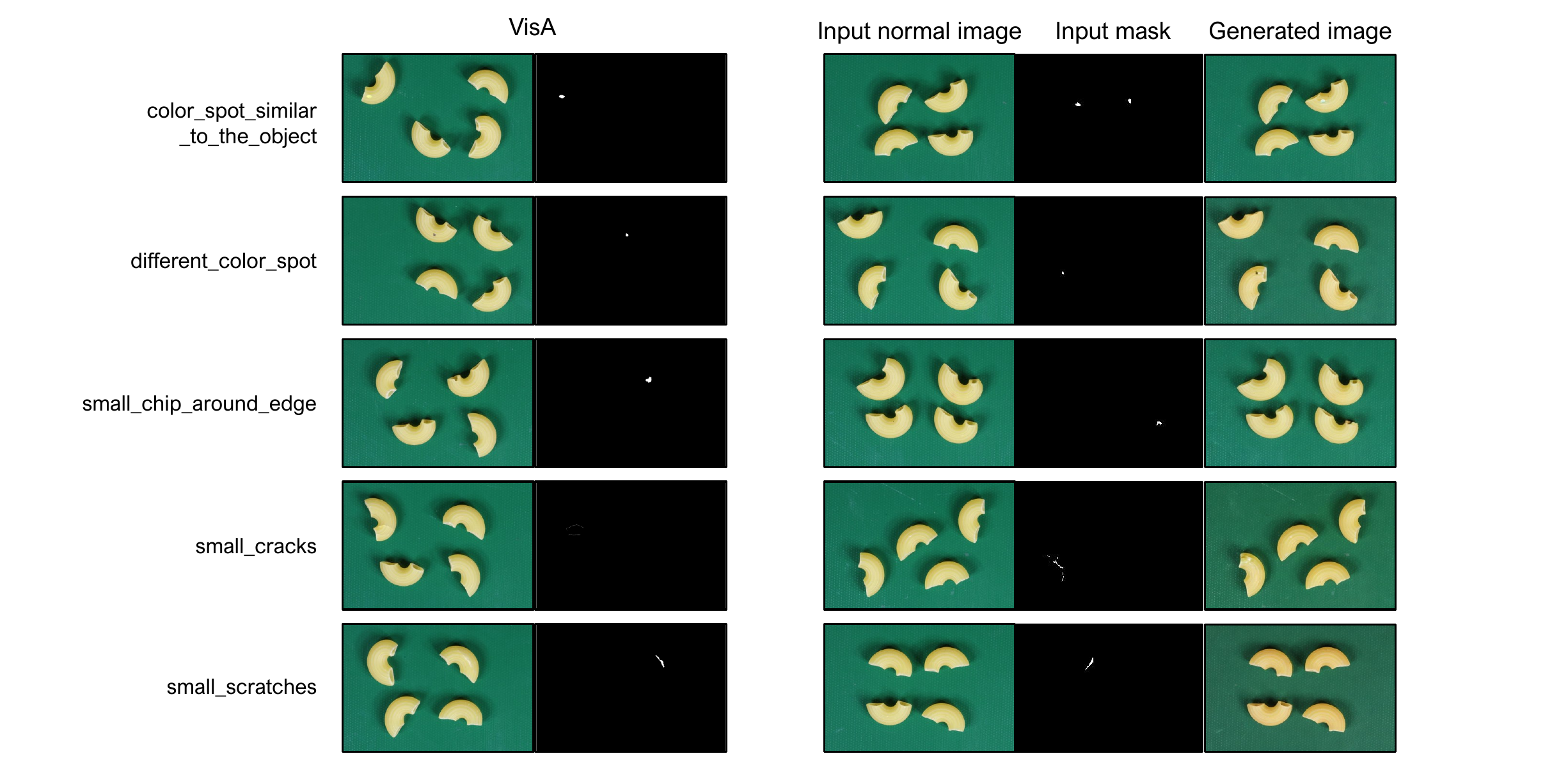}
  \vspace{-4mm}
    \caption{Generated images on \textit{macaroni2}
    }
  \vspace{-7mm}
\end{figure*}

\begin{figure*}[t]
  \centering
  \includegraphics[width=1.0\textwidth]{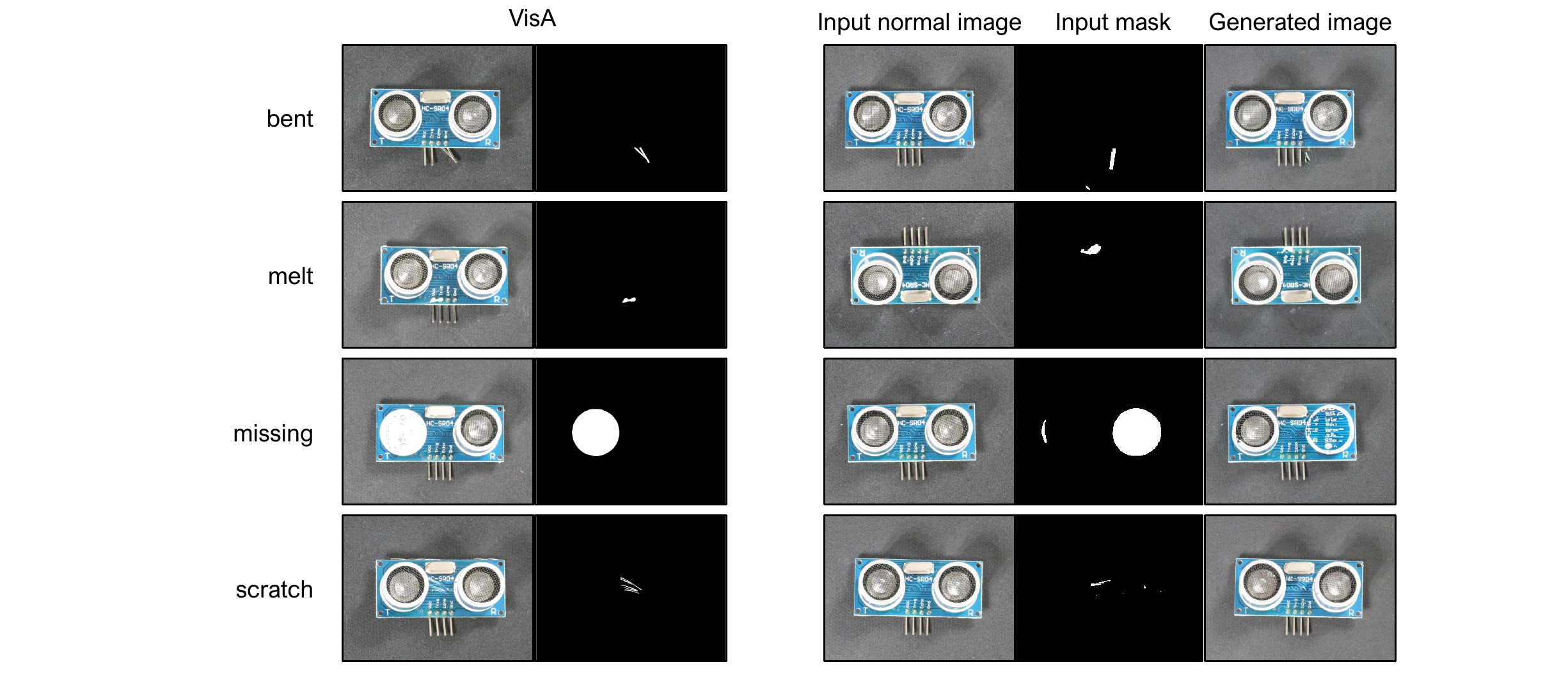}
  \vspace{-4mm}
    \caption{Generated images on \textit{pcb1}
    }
  \vspace{-7mm}
\end{figure*}

\begin{figure*}[t]
  \centering
  \includegraphics[width=1.0\textwidth]{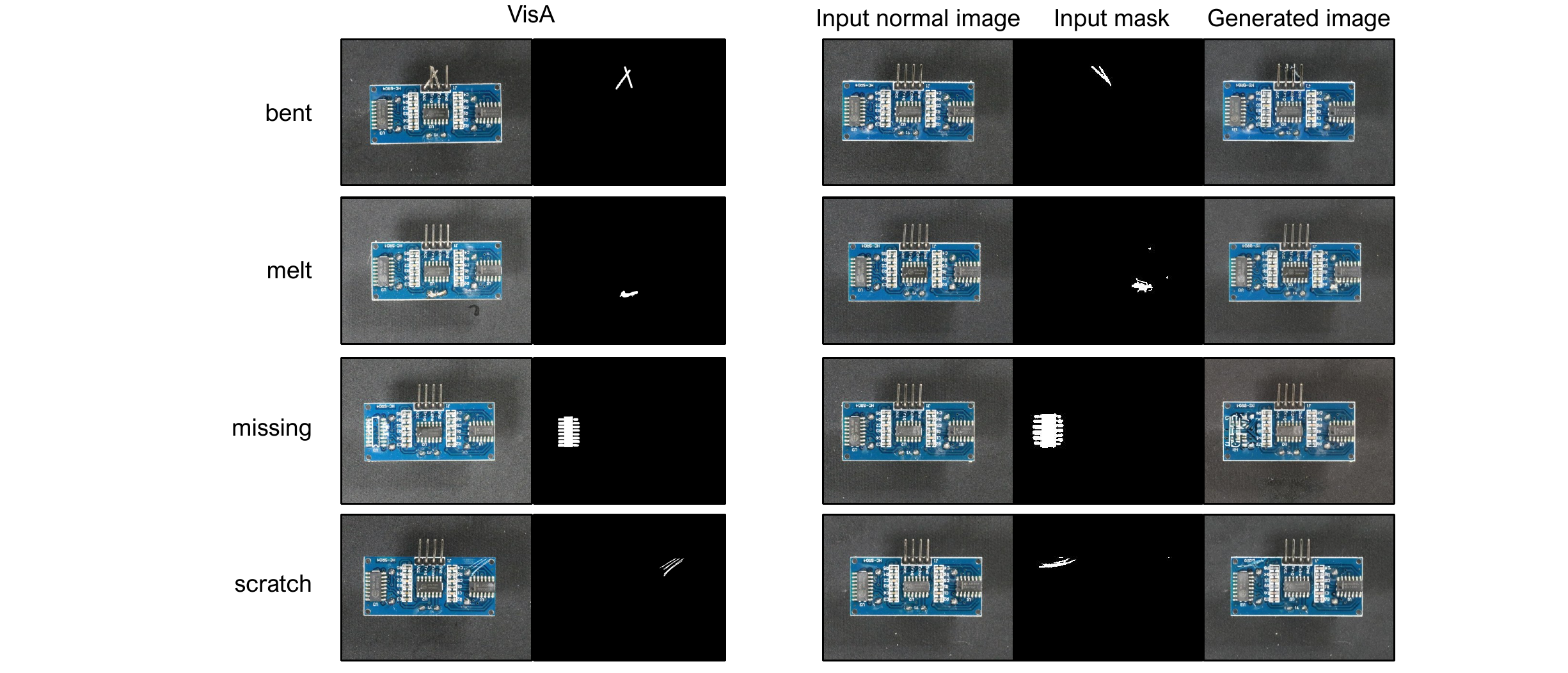}
  \vspace{-4mm}
    \caption{Generated images on \textit{pcb2}
    }
  \vspace{-7mm}
\end{figure*}

\begin{figure*}[t]
  \centering
  \includegraphics[width=1.0\textwidth]{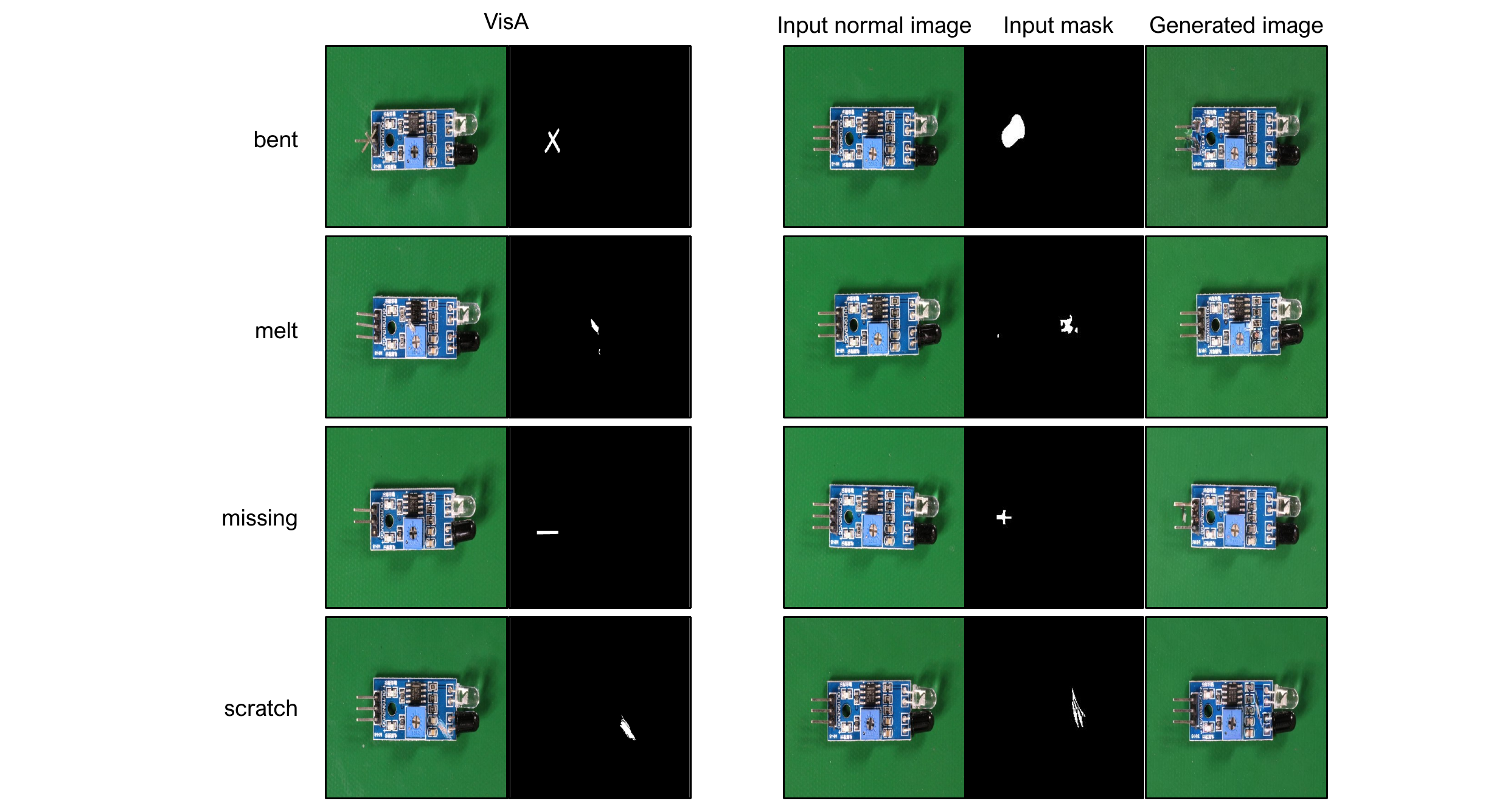}
  \vspace{-4mm}
    \caption{Generated images on \textit{pcb3}
    }
  \vspace{-7mm}
\end{figure*}

\begin{figure*}[t]
  \centering
  \includegraphics[width=1.0\textwidth]{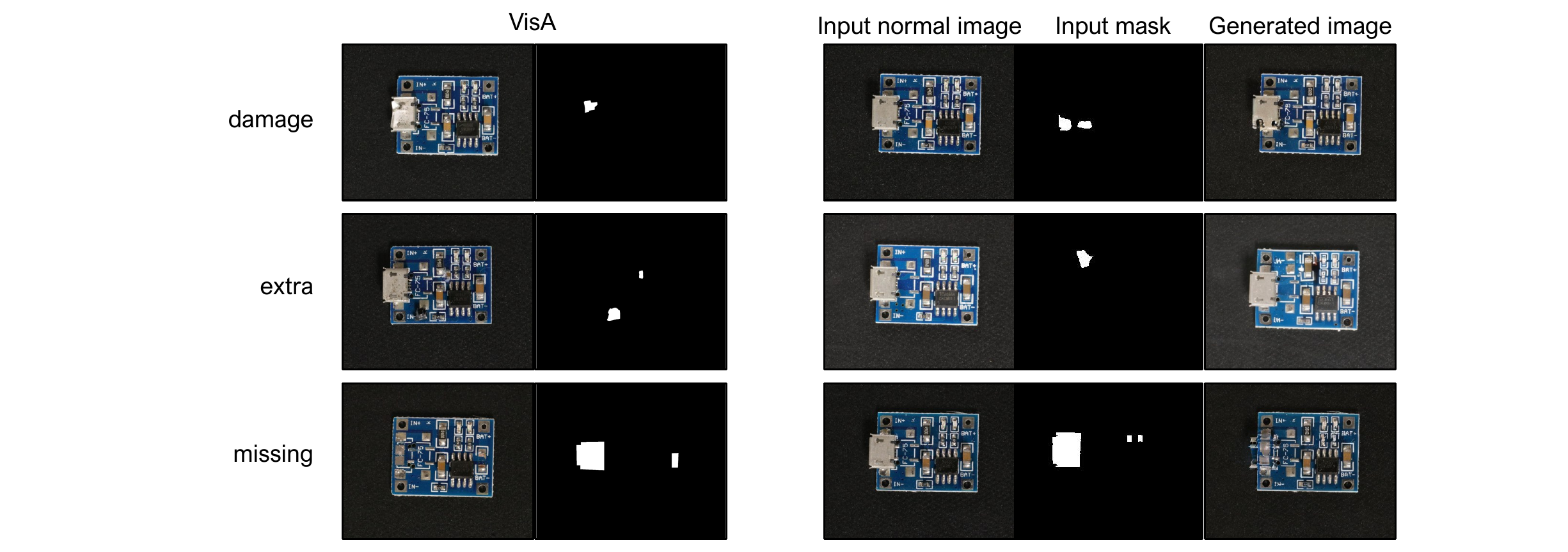}
  \vspace{-4mm}
    \caption{Generated images on \textit{pcb4}
    }
  \vspace{-7mm}
\end{figure*}

\begin{figure*}[t]
  \centering
  \includegraphics[width=1.0\textwidth]{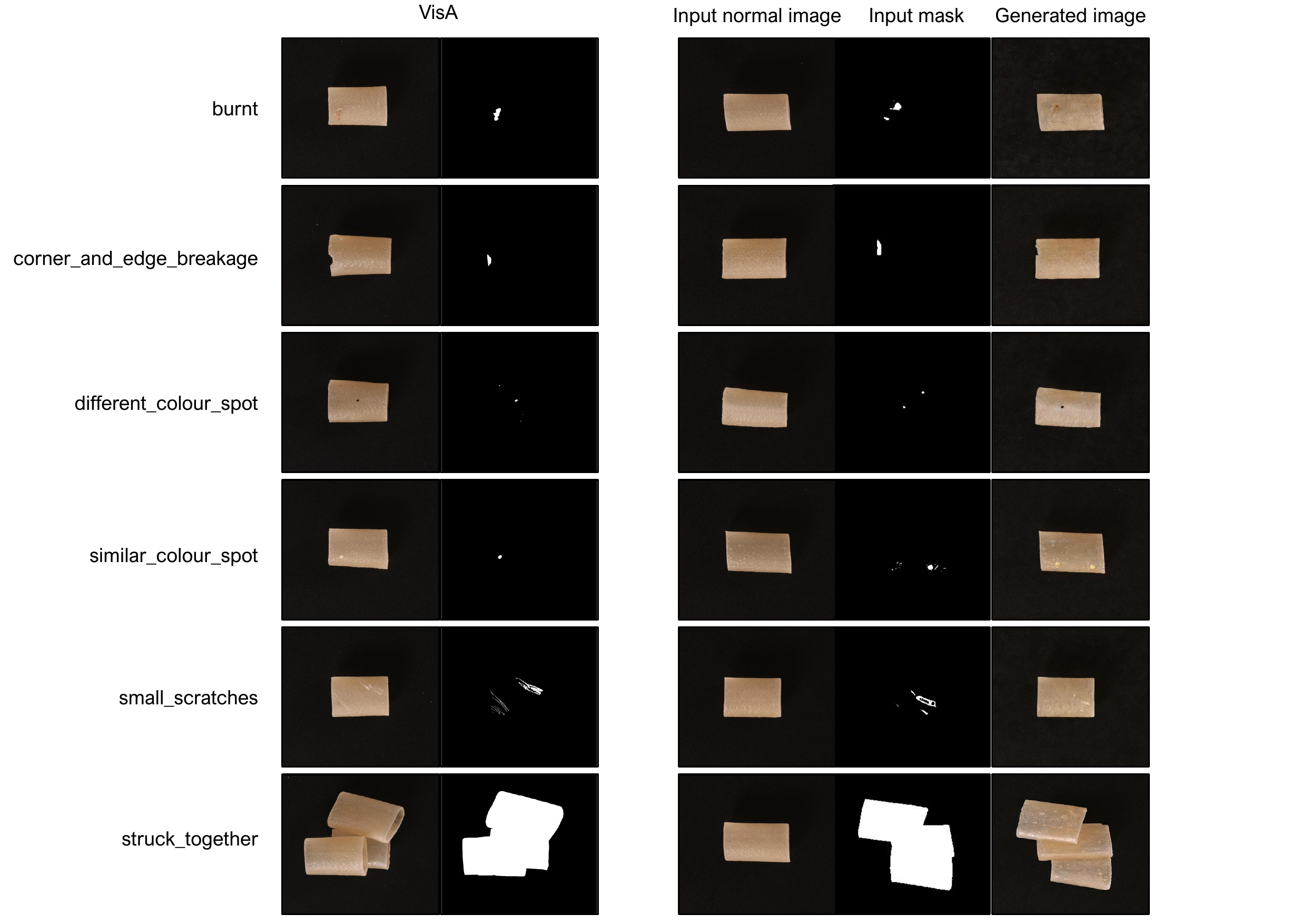}
  \vspace{-4mm}
    \caption{Generated images on \textit{pipe\_fryum}
    }
  \vspace{-7mm}
\end{figure*}

\clearpage

{
    \small
}


\end{document}